\documentclass{article}

\PassOptionsToPackage{numbers, sort}{natbib}
\usepackage{listings}
\usepackage{comment}

\usepackage[toc,page,header]{appendix}
\usepackage{minitoc}

\usepackage[final]{neurips_2026}

\usepackage[utf8]{inputenc} 
\usepackage[T1]{fontenc}    
\usepackage{hyperref}       
\usepackage{url}            
\usepackage{booktabs}       
\usepackage{amsfonts}       
\usepackage{nicefrac}       
\usepackage{microtype}      
\usepackage[table]{xcolor}         
\usepackage{wrapfig}
\usepackage{booktabs}
\usepackage{array}
\usepackage{longtable}
\usepackage{subcaption}

\usepackage[most]{tcolorbox}
\definecolor{tocrulecolor}{RGB}{40,80,180}       %
\definecolor{toctitlecolor}{RGB}{30,60,160}

\usepackage{import}
\usepackage{latexsym}
\usepackage{amsmath}
\usepackage{threeparttable}
\usepackage{graphicx}
\usepackage{placeins}
\usepackage{flafter}
\usepackage{afterpage}
\usepackage{pdflscape}
\usepackage{cleveref}
\usepackage{longtable}
\usepackage{booktabs}
\usepackage{array}
\usepackage{ragged2e}
\usepackage{svg}
\usepackage{tikz}

\definecolor{codebg}{RGB}{250,252,255}          %
\definecolor{coderule}{RGB}{180,200,230}         %
\definecolor{jkey}{RGB}{0,70,180}               %
\definecolor{jstr}{RGB}{20,110,40}              %
\definecolor{jnum}{RGB}{160,60,0}               %
\definecolor{shellbg}{RGB}{245,248,245}          %
\definecolor{shellrule}{RGB}{160,185,160}        %
\definecolor{tocrulecolor}{RGB}{40,80,180}       %
\definecolor{toctitlecolor}{RGB}{30,60,160}      %

\lstdefinelanguage{json}{
  basicstyle      = \small\ttfamily,
  breaklines      = true,
  captionpos      = t,
  frame           = single,
  rulecolor       = \color{coderule},
  framesep        = 5pt,
  backgroundcolor = \color{codebg},
  showstringspaces= false,
  moredelim       = [is][\color{jkey}\bfseries]{|>}{<|},
  morestring      = [b][\color{jstr}]",
  literate        =
    *{0}{{{\color{jnum}0}}}{1}
     {1}{{{\color{jnum}1}}}{1}
     {2}{{{\color{jnum}2}}}{1}
     {3}{{{\color{jnum}3}}}{1}
     {4}{{{\color{jnum}4}}}{1}
     {5}{{{\color{jnum}5}}}{1}
     {6}{{{\color{jnum}6}}}{1}
     {7}{{{\color{jnum}7}}}{1}
     {8}{{{\color{jnum}8}}}{1}
     {9}{{{\color{jnum}9}}}{1}
     {true}{{{\color{jnum}true}}}{4}
     {false}{{{\color{jnum}false}}}{5}
     {null}{{{\color{jnum}null}}}{4},
}

\lstdefinestyle{shell}{
  language        = bash,
  basicstyle      = \small\ttfamily,
  breaklines      = true,
  captionpos      = t,
  frame           = single,
  rulecolor       = \color{shellrule},
  framesep        = 5pt,
  backgroundcolor = \color{shellbg},
  commentstyle    = \color{gray!70}\itshape,
  showstringspaces= false,
}

\usepackage{enumitem}

\usetikzlibrary{arrows.meta, positioning, calc, fit, backgrounds, shapes.geometric, shapes.misc}

\usepackage{fancyvrb}
\usepackage{multirow}
\usepackage{float}
\usepackage{calc}
\usepackage{caption}
\usepackage{newunicodechar}
\usepackage{makecell}
\usepackage{soul}
\usepackage{bibunits}

\definecolor{lightgray}{gray}{0.9}

\newunicodechar{●}{$\bullet$}
\newunicodechar{◖}{$\circ$}
\newunicodechar{◐}{$\circ$}
\newcolumntype{L}[1]{>{\RaggedRight\arraybackslash}p{#1}}
\newcommand{\TODO}[1]{\textcolor{red}{[\textbf{TODO:} #1]}}

\usepackage{xcolor}
\usepackage{listings}

\definecolor{jsonbg}{HTML}{F7F7F8}
\definecolor{jsonframe}{HTML}{D0D7DE}
\definecolor{jsonkey}{HTML}{0550AE}
\definecolor{jsonstring}{HTML}{0A3069}
\definecolor{jsonnumber}{HTML}{953800}
\definecolor{jsonpunct}{HTML}{57606A}
\lstdefinelanguage{json}{
  morestring=[b]",
  stringstyle=\color{jsonstring},
  literate=
   *{0}{{{\color{jsonnumber}0}}}{1}
    {1}{{{\color{jsonnumber}1}}}{1}
    {2}{{{\color{jsonnumber}2}}}{1}
    {3}{{{\color{jsonnumber}3}}}{1}
    {4}{{{\color{jsonnumber}4}}}{1}
    {5}{{{\color{jsonnumber}5}}}{1}
    {6}{{{\color{jsonnumber}6}}}{1}
    {7}{{{\color{jsonnumber}7}}}{1}
    {8}{{{\color{jsonnumber}8}}}{1}
    {9}{{{\color{jsonnumber}9}}}{1}
    {:}{{{\color{jsonpunct}:}}}{1}
    {,}{{{\color{jsonpunct},}}}{1}
    {\{}{{{\color{jsonpunct}\{}}}{1}
    {\}}{{{\color{jsonpunct}\}}}}{1}
    {[}{{{\color{jsonpunct}[}}}{1}
    {]}{{{\color{jsonpunct}]}}}{1},
}

\lstdefinestyle{jsonstyle}{
  language=json,
  basicstyle=\ttfamily\small,
  backgroundcolor=\color{jsonbg},
  frame=single,
  rulecolor=\color{jsonframe},
  framerule=0.4pt,
  framesep=6pt,
  breaklines=true,
  showstringspaces=false,
  columns=fullflexible,
  keepspaces=true,
  upquote=true
}

\definecolor{ecViolet}{HTML}{6A5ACD}
\definecolor{ecTeal}{HTML}{1AA6A6}
\definecolor{ecOrange}{HTML}{F28E2B}
\definecolor{ecGreen}{HTML}{59A14F}
\definecolor{ecRed}{HTML}{E15759}
\definecolor{ecBlue}{HTML}{4E79A7}
\definecolor{ecGray}{HTML}{6B7280}
\definecolor{ecLight}{HTML}{F8FAFC}

\definecolor{lightestblue}{HTML}{F0F8FF}
\definecolor{mediumblue}{HTML}{6BAED6}
\definecolor{darkestblue}{HTML}{08519C}

\newcommand{\bbullet}{\textcolor{darkestblue}{\ensuremath{\bullet}}}
\newcommand{\ocirc}{\textcolor{mediumblue}{\ensuremath{\circ}}}
\renewcommand{\arraystretch}{0.95} 

\newcommand\EvalCards{\textsc{Evaluation Cards}}
\newcommand\EvalCard{\textsc{Evaluation Card}}

\title{\EvalCards{}: An Interpretive Layer for AI Evaluation Reporting}

\author{%
Avijit Ghosh$^{1,*}$ \quad Anka Reuel$^{2,*}$ \quad Jenny Chim$^{3,*}$ \quad Wm. Matthew Kennedy$^{19,*}$\\[0.4em]
Srishti Yadav$^{20,\diamond}$ \quad Jennifer Mickel$^{6,\diamond}$ \quad Yanan Long$^{13,\diamond}$ \quad Andrew Tran$^{14,\diamond}$\\[0.4em]
Anastassia Kornilova$^{5}$ \quad Damian Stachura$^{20}$ \quad Kevin Klyman$^{2}$ \quad Felix Friedrich$^{7}$ \quad Jeba Sania$^{9}$\\
Jan Batzner$^{8}$ \quad Anoop Mishra$^{22}$ \quad Eliya Habba$^{10}$ \quad Yixiong Hao$^{13}$ \quad Shalaleh Rismani$^{24}$\\
Nathan Heath$^{15,23}$ \quad Jessica Ji$^{31}$ \quad Usman Gohar$^{11}$ \quad David Manheim$^{26}$ \quad Aarush Sinha$^{4}$\\
Andrea Loehr$^{25}$ \quad Sree Harsha Nelaturu$^{27}$ \quad Leshem Choshen$^{12,33}$ \quad Asaf Yehudai$^{10}$\\
Drishti Sharma$^{20}$ \quad Ishan Khire$^{28}$ \quad Amit Saha$^{28}$ \quad Subramanyam Sahoo$^{20}$ \quad Michael Hardy$^{2}$\\
Michael Alexander Riegler$^{16}$ \quad Kabir Manghnani$^{2}$ \quad Michelle Lin$^{29}$ \quad Yanan Jiang$^{2}$\\
Yilin Huang$^{21}$ \quad Aris Hofmann$^{12,32}$ \quad Mubashara Akhtar$^{18}$ \quad Ruchira Dhar$^{4}$\\[0.4em]
Max Lamparth$^{2,\dagger}$ \quad Nuno Moniz$^{30}$ \quad Yacine Jernite$^{1,\dagger}$ \quad Stella Biderman$^{6,\dagger}$ \quad Zeerak Talat$^{17,\dagger}$\\
Sanmi Koyejo$^{2,\dagger}$ \quad Mykel Kochenderfer$^{2,\dagger}$ \quad Irene Solaiman$^{1,\dagger}$\\[0.4em]
\footnotesize$^{1}$Hugging Face \quad $^{2}$Stanford University \quad $^{3}$Queen Mary University of London \quad $^{4}$University of Copenhagen\\
\footnotesize$^{5}$Trustible \quad $^{6}$EleutherAI \quad $^{7}$TU Darmstadt \quad $^{8}$Weizenbaum Institute \& Technical University of Munich\\
\footnotesize$^{9}$Harvard University \quad $^{10}$The Hebrew University of Jerusalem \quad $^{11}$Iowa State University\\
\footnotesize$^{12}$IBM Research \quad $^{13}$StickFlux Labs \quad $^{14}$Independent \quad $^{15}$Oxford Martin AIGI\\ 
\footnotesize$^{16}$Simula \quad $^{17}$University of Edinburgh \quad $^{18}$ETH Zurich \& ETH AI Center \quad $^{19}$Oxford Internet Institute\\
\footnotesize$^{20}$Independent \quad $^{21}$Amherst College \quad $^{22}$University of Nebraska \quad $^{23}$Syntony Research\\
\footnotesize$^{24}$McGill University \quad $^{25}$Evals Consensus \quad $^{26}$Israel Institute of Technology\\
\footnotesize$^{27}$IOL.Learn \& Zuse Institute Berlin \quad $^{28}$Georgia Institute of Technology\\
\footnotesize$^{29}$Quebec AI Institute, Universit\'e de Montr\'eal \quad $^{30}$University of Notre Dame\\
\footnotesize$^{31}$Georgetown University \quad $^{32}$DHBW Stuttgart \quad $^{33}$Massachusetts Institute of Technology\\[0.1em]
\footnotesize$^{*}$First authors \quad $^{\diamond}$Top contributors \quad $^{\dagger}$Senior authors\\[0.1em]
\footnotesize Correspondence to: \texttt{avijit@huggingface.co} \quad \texttt{anka@cs.stanford.edu}\\[0.2em]
\footnotesize This project was completed as part of the EvalEval Coalition: \raisebox{-0.2ex}{\includegraphics[height=1em]{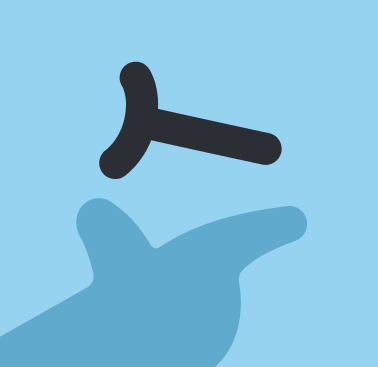}}\hspace{0.4em}\url{https://evalevalai.com/}
}

\begin{document}

\doparttoc 
\faketableofcontents 


\maketitle

\begin{abstract}
AI evaluation results are produced at scale but reported inconsistently across leaderboards, model cards, benchmark papers, and company blogs. The cost is interpretive: readers cannot reliably compare results across sources, identify what a report omits, or trace an aggregate claim to its underlying evidence. Recent efforts address isolated components but leave three gaps: they cover only narrow slices of the evaluation lifecycle and do not compose into a single interpretable record; they specify static representations that do not differentiate the questions different stakeholders bring to the same evidence; and they remain proposals on paper, lacking the extraction infrastructure required for adoption at scale. We present \EvalCards{}, an operational reporting layer that composes benchmark metadata, evaluation run data, and model metadata into a unified record. We (1) derive a reporting schema from a structured review of 52 papers and 10 stakeholder interviews, (2) implement four interpretive signals (reproducibility, documentation completeness, provenance and risk, and score comparability), rendered through reader modes calibrated to research and non-research audiences, and (3) deploy a monitoring tool that applies \EvalCards{} across 5,816 models, 635 benchmarks, and 101,843 results, surfacing systematic gaps in current reporting practice.

\end{abstract}

\section{Introduction}


The emerging AI evaluation ecosystem lacks shared infrastructure and conventions for reporting evaluation results. Leaderboards, model cards, benchmark papers, and company blogs use incompatible formats, omit fields required for interpretation, and provide no standard basis for cross-source comparison.  In addition, model developers report an inconsistent set of evaluations, creating gaps that are extremely difficult to define. Even when results are accessible, they frequently arrive without the context needed to assess their reliability, completeness, or comparability, leaving critical interpretive work to those least well positioned to perform it.


The cost of such fragmentation is not borne by evaluation developers, but by those who look to evaluation results to inform deployment decisions, regulatory assessments, and scientific claims about model capability \citep{reuel2024betterbench, weidinger2025evaluation, 10.1145/3708359.3712152}. Such claims depend on what is measured, under what assumptions, and how results are interpreted; when those three are not legible to the reader, a reported score does not translate into an actionable claim. The maturity of AI evaluations therefore depends on a vibrant ecosystem of actors, and on shared methods for reporting results, surfacing assumptions, and conveying their interpretations \citep{blind2016impact}.

Prior work addresses isolated components of this problem, focusing on two of its most important aspects: (i) reporting artifacts, which specify what should be documented about a model, dataset, benchmark, or evaluation; and (ii) evaluation infrastructure, which provides the schemas, repositories, and extraction pipelines that store and standardize evaluation evidence at scale. For instance, BenchmarkCards \citep{sokol2025benchmarkcards} and Auto-BenchmarkCards \citep{hofmann2026auto} standardize benchmark metadata, while Every Eval Ever (EEE) \citep{evalevalcoalition2024} specifies a schema for evaluation run data and maintains a community repository of standardized results. On the other hand, another line of work \citep{staufer2025audit, dhar2025evalcards, bordes2025evalfactsheetsstructuredframework, zhao-etal-2025-sphere} proposes reporting artifacts but covers only specific portions of the evaluation lifecycle (\Cref{tab:lifecycle-categories-table}). This has led to a need for a standardized format that supports comparative evaluation reporting and a shared interface for rendering evaluation evidence to readers with different information needs. Moreover, most reporting artifacts remain proposals on paper, without the extraction pipelines, hosted interfaces, or infrastructure required for adoption at scale. In practice, evaluation documentation has relied on risk management frameworks \citep{ai2023artificial} and checklists \citep{mitchell2019model, gebru-2021-datasheets}, approaches that impose a high cognitive burden, duplicate effort across teams, and produce information silos rather than shared, comparable records. Requiring evaluators to manually populate new forms and checklists duplicates effort for model and system cards. This kind of documentation has historically not been widely taken up \citep{rao2025aimodelriskcatalog}. 
\begin{figure}[t]
\centering
\resizebox{\textwidth}{!}{%
\begin{tikzpicture}[
    every node/.style={font=\Large},
    source/.style={
        rectangle, rounded corners=4pt,
        draw=blue!60, fill=blue!8,
        text width=4.5cm, align=center,
        minimum height=2.2cm, minimum width=4.7cm},
    resolver/.style={
        rectangle, rounded corners=4pt,
        draw=none, fill=none,
        text width=3.4cm, align=center,
        minimum height=1.6cm, minimum width=3.6cm},
    grouptwo/.style={
        rectangle, rounded corners=6pt,
        draw=#1!70, fill=#1!6,
        minimum height=9.0cm, minimum width=6.4cm},
    substage/.style={
        rectangle, rounded corners=4pt,
        draw=#1!80, fill=#1!16,
        text width=5.2cm, align=center,
        minimum height=3.2cm, minimum width=5.4cm,
        font=\Large\bfseries},
    substageJ/.style={
        rectangle, rounded corners=4pt,
        draw=orange!80, fill=orange!16,
        text width=5.2cm, align=center,
        minimum height=2.4cm, minimum width=5.4cm,
        font=\Large\bfseries},
    arrow/.style={-{Stealth[length=7pt]}, line width=1.2pt},
    dasharrow/.style={-{Stealth[length=6pt]}, line width=1.0pt,
        dashed, gray!60},
    thickarrow/.style={-{Stealth[length=9pt]}, line width=2.0pt},
]


\node[source] (EEE) at (0, 0) {%
    \textbf{EEE Datastore}\\
    {\large Evaluation run data}};

\node[source, right=0.8cm of EEE] (CARDS) {%
    \textbf{Auto-BenchmarkCards}\\
    {\large Benchmark metadata}};

\node[source, right=0.8cm of CARDS] (META) {%
    \textbf{Model Metadata}\\
    {\large Developer, size,\\release date}};

\node[source, right=0.8cm of META] (REG) {%
    \textbf{Entity Registry}\\
    {\large Canonical dims,\\aliases, promotions}};

\node[font=\LARGE\itshape, text=blue!70]
    at ($(EEE.north)!0.5!(REG.north) + (0, 0.65cm)$) {Sources};


\coordinate (rowstart) at ($(EEE.south west) + (0, -1.8cm)$);

\node[grouptwo=violet, anchor=north west] (GRP1) at (rowstart) {};

\node[substage=violet, anchor=north]
    (A) at ($(GRP1.north) + (0, -0.6cm)$) {%
    \textbf{A \textperiodcentered\ Load}\\[3pt]
    {\normalfont\large Validate \& ingest}\\
    {\normalfont\large all sources into}\\
    {\normalfont\large typed tables}};

\node[substage=violet, anchor=north]
    (B) at ($(A.south) + (0, -0.5cm)$) {%
    \textbf{B \textperiodcentered\ Explode}\\[3pt]
    {\normalfont\large One row per result}\\
    {\normalfont\large Mint stable fact IDs}};

\node[font=\LARGE\bfseries, text=violet!70, anchor=south]
    at ($(GRP1.south) + (0, 0.3cm)$) {Ingest};

\node[grouptwo=violet, anchor=north west]
    (GRP2) at ($(GRP1.north east) + (0.8cm, 0)$) {};

\node[substage=violet, anchor=north]
    (C) at ($(GRP2.north) + (0, -0.6cm)$) {%
    \textbf{C \textperiodcentered\ Resolve}\\[3pt]
    {\normalfont\large Map to canonical}\\
    {\normalfont\large model, benchmark}\\
    {\normalfont\large \& metric IDs}};

\node[substage=violet, anchor=north]
    (D) at ($(C.south) + (0, -0.5cm)$) {%
    \textbf{D \textperiodcentered\ Flatten}\\[3pt]
    {\normalfont\large Join benchmark}\\
    {\normalfont\large metadata to}\\
    {\normalfont\large run records}};

\node[font=\LARGE\bfseries, text=violet!70, anchor=south]
    at ($(GRP2.south) + (0, 0.3cm)$) {Standardize};

\node[grouptwo=teal, anchor=north west]
    (GRP3) at ($(GRP2.north east) + (0.8cm, 0)$) {};

\node[substage=teal, anchor=north]
    (E) at ($(GRP3.north) + (0, -0.6cm)$) {%
    \textbf{E \textperiodcentered\ Per-row}\\[3pt]
    {\normalfont\large Reproducibility \&}\\
    {\normalfont\large Completeness}\\
    {\normalfont\large flags per result}};

\node[substage=teal, anchor=north]
    (F) at ($(E.south) + (0, -0.5cm)$) {%
    \textbf{F \textperiodcentered\ Group}\\[3pt]
    {\normalfont\large Provenance \&}\\
    {\normalfont\large Comparability}\\
    {\normalfont\large across sources}};

\node[font=\LARGE\bfseries, text=teal!70, anchor=south]
    at ($(GRP3.south) + (0, 0.3cm)$) {Compute signals};

\node[grouptwo=orange, anchor=west, minimum height=12.0cm]
    (GRP4) at ($(GRP3.east) + (0.8cm, 0)$) {};

\node[substage=violet, anchor=north]
    (G) at ($(GRP4.north) + (0, -0.4cm)$) {%
    \textbf{G \textperiodcentered\ Dims}\\[3pt]
    {\normalfont\large Build rollout hierarchy}\\
    {\normalfont\large family, composite,}\\
    {\normalfont\large benchmark}};

\node[substage=teal, anchor=north]
    (I) at ($(G.south) + (0, -0.3cm)$) {%
    \textbf{I \textperiodcentered\ Warehouse}\\[3pt]
    {\normalfont\large Emit 6 canonical}\\
    {\normalfont\large parquet files}};

\node[substageJ, anchor=north]
    (J) at ($(I.south) + (0, -0.3cm)$) {%
    \textbf{J \textperiodcentered\ Views}\\[3pt]
    {\normalfont\large Corpus, model \& benchmark}\\
    {\normalfont\large\itshape\textcolor{orange!70}{\textperiodcentered\ Research Mode\ \textperiodcentered\ Summary Mode}}};

\node[font=\LARGE\bfseries, text=orange!70, anchor=south]
    at ($(GRP4.south) + (0, 0.3cm)$) {Output};


\draw[arrow, blue!60] (EEE.south)   -- ++(0,-0.6) -| (GRP1.north);
\draw[arrow, blue!60] (CARDS.south) -- ++(0,-0.6) -| (GRP1.north);
\draw[arrow, blue!60] (META.south)  -- ++(0,-0.6) -| (GRP1.north);
\draw[arrow, blue!60] (REG.south)   -- ++(0,-0.6) -| (GRP1.north);


\draw[thickarrow, violet!70] (GRP1.east) -- (GRP2.west);
\draw[thickarrow, violet!70] (GRP2.east) -- (GRP3.west);
\draw[thickarrow, teal!70]   (GRP3.east) -- (GRP4.west);

\draw[arrow, violet!80] (A.south) -- (B.north);
\draw[arrow, violet!80] (C.south) -- (D.north);
\draw[arrow, teal!80]   (E.south) -- (F.north);
\draw[arrow, violet!80] (G.south) -- (I.north);
\draw[arrow, teal!70]   (I.south) -- (J.north);

\end{tikzpicture}%
}
\caption{Backend canonicalization pipeline. Four sources feed four stage groups.
\textit{Ingest} (A--B): load and explode sources into one row per result.
\textit{Standardize} (C--D): resolve canonical identities and join benchmark metadata.
\textit{Compute signals} (E--F): per-result reproducibility and completeness flags (E);
grouped provenance and comparability across sources (F).
\textit{Output} (G--J): materialize rollout hierarchy (G), emit warehouse parquets (I),
render corpus, model, and benchmark views through Research and Policy reader modes (J).
Colors: violet\,=\,transformation, teal\,=\,signals and warehouse, orange\,=\,output.}
\label{fig:pipeline}
\end{figure}
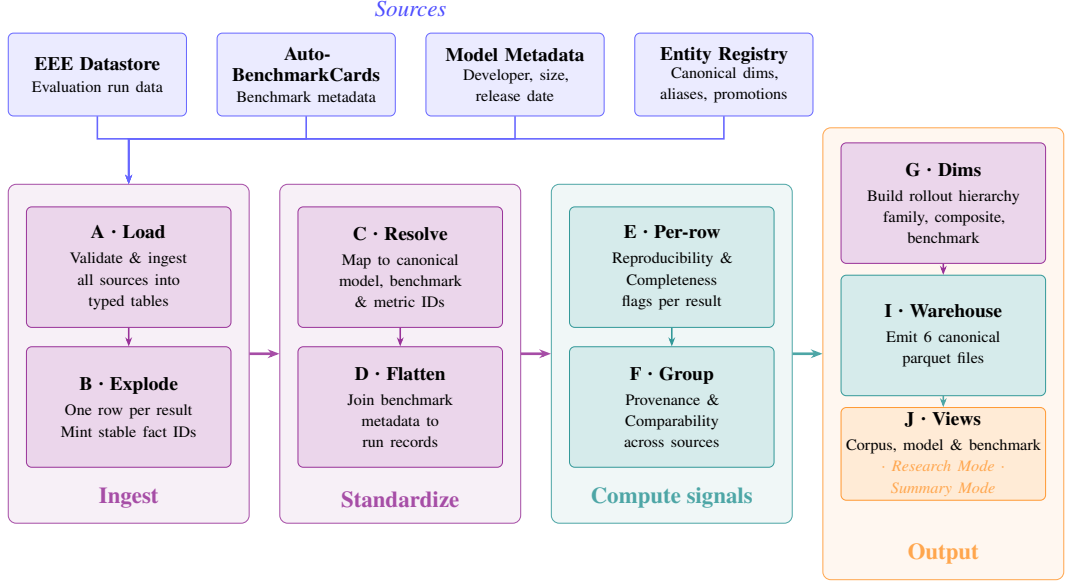

We therefore present \EvalCards{}, a \href{https://evalcards.evalevalai.com}{live interactive reporting layer} that unifies existing evaluation infrastructure and surfaces interpretive signals across it. This paper contributes:

\begin{itemize}
    \item \textbf{A reporting framework grounded in literature and practitioner needs.} A structured set of items specifying what should accompany an evaluation at publication so that it can be reproduced, contextualized, and compared. The framework is derived from a systematic review of 52 papers and semi-structured interviews with 12 stakeholders across technical, developer, and policy roles.

    \item \textbf{A rollout hierarchy for evaluation evidence.} A five-level structure (family $\to$ composite $\to$ benchmark $\to$ split $\to$ metric) that replaces the flat (model, benchmark, score) triples used by leaderboards and model cards, resolving every reported score to an explicit traceable path. For example, ``GPT-5 achieves 0.994 on MATH'' resolves to \texttt{MATH-family} $\to$ \texttt{artificial\_analysis} $\to$ \texttt{MATH-500} $\to$ \texttt{advanced-math} $\to$ \texttt{accuracy}, letting readers trace aggregate claims to the specific evidence behind them and disambiguating measurements that share a label but differ in subtask, setup, or scoring rule (\Cref{subsec:rollout-hierarchy}).

    \item \textbf{Four interpretive signals over the populated schema.} Reproducibility (flags missing generation and prompting settings), reporting completeness (per-field coverage against the framework), provenance (first- versus third-party attribution with risk annotations propagated from Auto-BenchmarkCards), and comparability (setup differences and score divergence across variants and reporters). Signals are surfaced through reader modes calibrated to research and non-research audiences.

    \item \textbf{An empirical analysis of public evaluation reporting at scale.} A deployed instrument applying \EvalCards{} to 5,816 models, 635 benchmarks, and 101,955 reported results contributed by 30 organizations, yielding three headline findings: 96.5\% of (model, benchmark, metric-path) triples lack at least one minimal reproducibility field, with the gap widest in developer self-reporting (0.0\% vs.\ 16.6\% field population on paired first/third-party reports); median per-benchmark documentation completeness is 10.7\% against the operationalized schema; and 98.2\% of (model, benchmark) pairs are reported by only one party, with cross-party divergence above the 5\% threshold in 51.9\% of multi-organization metric groups.
\end{itemize}

\Cref{sec:relatedwork} positions the work against prior reporting and infrastructure efforts. \Cref{sec:evalcards} presents the framework and rollout hierarchy. \Cref{sec:interprative-layer} describes composition with EEE \cite{evalevalcoalition2024} and Auto-BenchmarkCards \cite{hofmann2026auto} and the four interpretive signals. \Cref{sec:evalcards-in-practice} reports the empirical audit. \Cref{sec:community} discusses community adoption and future extensions.

\section{Related Work}
\label{sec:relatedwork}

Prior work relevant to \EvalCards{} falls into technical reporting documentation (covering parts of the evaluation lifecycle), evaluation infrastructure and data schemes, and systematic reviews of evaluation practice. \Cref{tab:scope-comparison} summarizes their scope; \Cref{app:related-work} gives a detailed overview of related work. No prior artifact joins reporting framework, run data, and benchmark metadata, differentiates rendering by reader type, and provides a continuous monitoring instrument for the state of public evaluation reporting at scale. \EvalCards{} are designed to address these gaps.

\begin{table}[ht]
\centering
\caption{Comparing  technical reporting documentation and infrastructure to \EvalCards{}. 
}
\label{tab:scope-comparison}
\resizebox{\textwidth}{!}{%
\begin{tabular}{@{}l c c c c c c@{}}
\toprule
\textbf{Artifact} & \makecell{Reporting\\framework} & \makecell{Run data\\schema} & \makecell{Benchmark\\metadata} & \makecell{Cross-layer\\composition} & \makecell{Reader\\modes} & \makecell{Monitoring\\tool} \\
\midrule
Datasheets \citep{gebru-2021-datasheets} & & & & & & \\
Data Cards \citep{10.1145/3531146.3533231} & & & & & & \\
Model Cards \citep{mitchell2019model} & $\ocirc$ & & & & & \\
BenchmarkCards \citep{sokol2025benchmarkcards} & & & $\bbullet$ & & & \\
Auto-BenchmarkCards \citep{hofmann2026auto} & & & $\bbullet$ & & & \\
Audit Cards \citep{staufer2025audit} & $\ocirc$ & & & & & \\
EvalCards \citep{dhar2025evalcards} & $\ocirc$ & & & & & \\
Eval Factsheets \citep{bordes2025evalfactsheetsstructuredframework} & $\ocirc$ & & & & & \\
SPHERE \citep{zhao-etal-2025-sphere} & $\ocirc$ & & & & & \\
STREAM \citep{mccaslin2025stream} & $\ocirc$ & & & & & \\
HELM \citep{liang2023holisticevaluationlanguagemodels} & $\ocirc$ & $\ocirc$ & & & & \\
Inspect \citep{inspect} & & $\bbullet$ & & & & \\
OpenLLM Leaderboard \citep{Beechingetal2023} & & $\ocirc$ & $\ocirc$ & & & \\
EEE \citep{evalevalcoalition2024} & & $\bbullet$ & & & & \\
BetterBench \citep{reuel2024betterbench} & $\ocirc$ & & $\ocirc$ & & & \\
\midrule
\textbf{\EvalCards{} (this paper)} & $\bbullet$ & $\bbullet$ & $\bbullet$ & $\bbullet$ & $\bbullet$ & $\bbullet$ \\
\bottomrule
\end{tabular}
}
\begin{center}
{\small $\bbullet$~Primary Object \quad $\ocirc$~Partial or Indirect Coverage \quad Blank = Out of Scope}
\end{center}
\begin{minipage}{\textwidth}
\small
\raggedright
\textit{Reporting framework}: documentation specifying fields to accompany a model, dataset, or benchmark at release.\\
\textit{Run data schema}: per-run log of evaluation executions (prompts, completions, scores, configuration).\\
\textit{Benchmark metadata}: model-independent description of a benchmark (construction, provenance, scoring, splits, limitations).\\
\textit{Cross-layer composition}: single artifact jointly covering the three above, linking each score to its run log and benchmark context.\\
\textit{Reader modes}: presentations tailored to different audiences (developers, auditors, policymakers).\\
\textit{Monitoring tool}: continuous monitoring instrument measuring the current state of public reporting at scale.
\end{minipage}
\end{table}
\vspace{-0.6em}

\section{The \EvalCards{} Framework}
\label{sec:evalcards}

This section introduces the \EvalCard{} format (\Cref{subsec:reporting-framework}) and its hierarchical structure over which evaluation evidence is organized (\Cref{subsec:rollout-hierarchy}). \Cref{sec:evalcards-in-practice} describes how reports are populated from existing sources and how interpretive signals are computed over them. \EvalCards{} are designed as a permissive reporting standard rather than a prescriptive checklist. Evaluators are not required to populate every field for an \EvalCard{} to be useful; partial population still produces a structured, comparable record and supports the rollout hierarchy and interpretive signals introduced below. 

\subsection{An evidence-based reporting framework}
\label{subsec:reporting-framework}
We identified key requirements needed for a comprehensive evaluation report through a systematic review of 52 papers on AI evaluation practice published between 2020 and 2025. The outcome of the review was synthesized into the \EvalCards{} reporting framework, which was then partially operationalized by reporting on the artifacts. The \EvalCards{} platform was then refined/evaluated through semi-structured interviews. 

\textbf{Systematic literature review.} A preregistered systematic review (reported in detail in \Cref{app:systematic-literature-review}) yielded 748 candidate papers, of which 52 met our inclusion criteria: published reports, peer-reviewed or preprint work proposing evaluation practices, frameworks, or reporting standards for AI systems. Two reviewers independently extracted and coded 730 recommendation items capturing when, where, how, and why each recommendation applied in the evaluation process. Each item was categorized along several dimensions such as item class, type, and intent (\Cref{codebook}) with high inter-rater agreement across all categories: Cohen's $\kappa \in [0.865, 0.895]$ for single-label dimensions and Krippendorff's $\alpha \in [0.916, 0.964]$ for multi-label dimensions \citep{landiskoch1977} (see \Cref{app:methodology} for details). A `best fit’ framework method \citep{carroll2011bestfit} was used to derive the five-part structure (shown in Table 2), drawing on existing frameworks for AI evaluation \citep{liang2023holisticevaluationlanguagemodels, reuel2024betterbench,bean2025measuring,paskov2025preliminary,aisi2025structured,alizadeh2025allmetrics,rismani2025measuring} and expert feedback.

\textbf{Scope: artifact-side reporting.} Not all items from the \EvalCard{} framework were directly operationalized in the \EvalCard{} platform/fields. Requiring evaluators to populate fields with no published trace would recreate the adoption barriers that have limited prior reporting tools. Therefore, process-side decisions (such as pre-run protocol commitments, pilot runs, or mid-run mitigations,) are documented in linked complementary mechanisms like preregistration systems, and only items likely to appear in published artifacts are directly included.

\textbf{Interviews.} To complement the literature-derived items with practitioner perspectives, we conducted semi-structured interviews between February and April 2026 with 12 stakeholders spanning technical evaluators, AI engineers, and policy actors across 9 organizations and 3 geographic regions (full protocol and participant profile in Appendix B). Interviews surfaced which framework items participants treated as blocking versus enriching for their work, how field salience varied across reader types, and where existing technical reporting documentation produced friction. The full framework-to-field mapping is in Appendix \ref{app:data-normalization}.

\begin{table}[ht]
\centering
\caption{The \EvalCards{} reporting framework.}
\label{tab:lifecycle-categories-table}

\renewcommand{\arraystretch}{1.25}
\begin{tabular}{p{0.20\linewidth}p{0.72\linewidth}}
\toprule
\textbf{Category} & \textbf{Item groups} \\
\midrule

\textbf{1.\ Design} &
Goals, tested constructs \& context; development and design preregistration; validity; task types \& item development; human subjects / ethics. \\

\textbf{2.\ Before execution} &
Protocol \& pre-run; scoring \& validation; splits \& holdouts; pilot \& baselines; contamination, gaming \& awareness; pre-reporting. \\

\textbf{3.\ Execution} &
Run logging \& reproducibility capture; mitigations \& adaptations; analysis \& run differences. \\

\textbf{4.\ Lifecycle} &
Data availability \& access; later use \& maintenance. \\

\makecell[l]{\textbf{5.\ Reporting \&}\\\hspace{1.1em}\textbf{publication}} &
\makecell[l]{{Reporting \& publication; process reporting; transparency;}\\ {replication \& reproducibility.}} \\
\bottomrule

\end{tabular}
\begin{minipage}{0.94\linewidth} \vspace{0.1em}
\footnotesize
Each top-level category corresponds to a stage of the evaluation lifecycle at which reporting choices are made. Full details of each category are provided in Appendix~\ref{app:systematic-literature-review}; how these categories map to specific \EvalCards{} fields is provided in Appendix~\ref{app:schema-mapping}.
\end{minipage}
\end{table}

\subsection{A rollout hierarchy for evaluation evidence}
\label{subsec:rollout-hierarchy}

Evaluation reporting typically treats an evaluation as a flat triple: model, benchmark name, score. This flattening obscures internal structure present in almost every benchmark in current use. Reasoning benchmarks like MATH contain subject-level splits (algebra, number theory, geometry) with different difficulty profiles. Agentic benchmarks like SWE-bench report results per language and per setup variant. Composite benchmarks like Open LLM Leaderboard v2 aggregate scores across six independently maintained sub-benchmarks. Safety suites report results per risk category. Critically, treating these as flat, context-devoid objects erases the granularity at which evaluation claims are defensible, which was also raised or discussed by interviewees P1, P2, P3, P4, P5, T2, T3, T4, T6, and T7.

In response, we introduce a five-level rollout hierarchy that reflects this internal structure: \textbf{Family}, a group of related benchmarks sharing a common object of measurement or methodological lineage (e.g., SWE-bench, MMLU); \textbf{Composite}, a named composite that aggregates multiple benchmarks under a unified presentation (e.g., Artificial Analysis Index); benchmarks need not necessarily be part of a composite; \textbf{Benchmark}, an individual evaluation with a defined dataset and scoring method (e.g., GSM8K, IFEval, MMLU-Pro); \textbf{Split}, a subcategory within a benchmark (e.g., algebra within MATH, the Python subset of Multi-SWE-Bench); and \textbf{Metric}, the scoring rule attached to a result (e.g., pass@1, accuracy, F1).

Every score in \EvalCards{} resolves to a path through this hierarchy, rather than as a flat (model, benchmark, score) triple. 

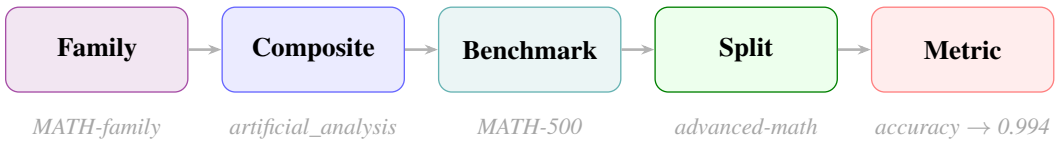
\begin{figure}[!th]
\centering
\resizebox{\textwidth}{!}{%
\begin{tikzpicture}[
    node distance=0.35cm,
    every node/.style={font=\small},
    level/.style={rectangle, rounded corners=4pt,
        text width=1.7cm, align=center,
        minimum height=0.9cm, font=\footnotesize\bfseries},
    l1/.style={level, draw=violet!70,      fill=violet!10},
    l2/.style={level, draw=blue!60,        fill=blue!8},
    l3/.style={level, draw=teal!60,        fill=teal!8},
    l4/.style={level, draw=green!50!black, fill=green!7},
    l5/.style={level, draw=red!50,         fill=red!7},
    arrow/.style={-{Stealth[length=4pt]}, thick, gray!60},
    ex/.style={font=\scriptsize\itshape, text=gray!70, align=center}
]
\node[l1] (n1) {Family};
\node[l2, right=of n1] (n2) {Composite};
\node[l3, right=of n2] (n3) {Benchmark};
\node[l4, right=of n3] (n4) {Split};
\node[l5, right=of n4] (n5) {Metric};
\draw[arrow] (n1)--(n2); \draw[arrow] (n2)--(n3);
\draw[arrow] (n3)--(n4); \draw[arrow] (n4)--(n5);
\node[ex, below=0.15cm of n1] {MATH-family};
\node[ex, below=0.15cm of n2] {artificial\_analysis};
\node[ex, below=0.15cm of n3] {MATH-500};
\node[ex, below=0.15cm of n4] {advanced-math};
\node[ex, below=0.15cm of n5] {accuracy $\to$ 0.994};
\end{tikzpicture}%
}
\caption{The five-level rollout hierarchy. Every reported score resolves to a full path through these levels, replacing a flat (model, benchmark, score) triple. We share examples in Appendix \ref{app:walkthroughs-rollout}.}
\label{fig:rollout-hierarchy}
\end{figure}

This structure has three implications. First, integrity signals attach to specific paths rather than to benchmark names: a reproducibility warning, for example, applies to a (model, metric-path) pair, not to a benchmark label. Second, the hierarchy enables drill-down, letting a reader trace an aggregate family-level claim to the specific metric that supports it and see where the claim is well-evidenced versus where it depends on a single reported number. Third, it disambiguates conflicting measurements: results sharing a model label (e.g. \texttt{gpt-4}, \texttt{gpt-4-0613}, \texttt{OpenAI GPT-4}) or a benchmark label but differing in subtask or metric resolve to different paths, preventing conflation, which resolves a known gap in existing sources (e.g. EEE, which defers deduplication to the analysis layer \citep{Batzner*2026-qj}) and is what makes multi-source results comparable within \EvalCards{}. 

\section{\EvalCards{}: Pipeline \& Interpretive Layer}\label{sec:interprative-layer}

This section describes how \EvalCards{} populate its schema from existing sources (Section \ref{sec:interprative-layer-composition}) and how the four interpretive signals are computed over the populated schema and rendered through two reader modes (Section \ref{sec:interprative-layer-signals}), forming the foundation of our \EvalCards{}.

\subsection{Composition over existing sources}
\label{sec:interprative-layer-composition}

\EvalCards{} draws on three existing sources, described below; per-field operationalization is specified in \Cref{app:evalcards-fields} and the full data normalization pipeline in \Cref{app:data-normalization}.

\textbf{Benchmark metadata via Auto-BenchmarkCards.} Auto-BenchmarkCards \citep{hofmann2026auto} auto-generate benchmark cards \citep{sokol2025benchmarkcards} with meta-benchmark information by extracting content from Unitxt catalogs, Hugging Face repositories, and associated publications, composing them through an LLM into structured cards, and validating factual consistency. \EvalCards{} ingests these cards as its benchmark metadata source, populating fields in the \textit{Design}, \textit{Lifecycle}, and \textit{Reporting \& publication} categories of the framework. Risk annotations attached by Auto-BenchmarkCards via the IBM AI Atlas Nexus risk-identification framework \citep{bagehorn2025airiskatlastaxonomy} are carried over to \EvalCards{} as provenance inputs (Section \ref{sec:interprative-layer-signals}). \EvalCards{} does not modify ingested Auto-BenchmarkCards content; original fields are surfaced as-is. To support filtering across the corpus, each benchmark is assigned one or more category tags drawn from an 18-category taxonomy derived from \citet{ni2025surveylargelanguagemodel}; the categorization methodology is described in Appendix~\ref{app:methodology-categorization}. Tags are used for the corpus-level analyses in Section \ref{sec:evalcards-in-practice} and for filtering in the interface. The tag is stored as an \EvalCards{} annotation rather than written back into the source card.

\textbf{Evaluation run data via EEE.} EEE \citep{Batzner*2026-qj} is a schema and community repository for evaluation run data at both aggregate and instance level, with converters from major evaluation frameworks (HELM, lm-eval-harness, Inspect AI) and a growing corpus of community contributions. \EvalCards{} ingests EEE records as the run-data source, populating fields in the \textit{Execution} and \textit{Reporting \& publication} categories. The \texttt{evaluator\_relationship} field (first-party, third-party, collaborative) is the primary input to the provenance signal; \texttt{generation\_config.generation\_args} fields (\texttt{temperature}, \texttt{top\_p}, \texttt{max\_tokens}, \texttt{prompt\_template}, \texttt{reasoning}) are the primary inputs to the reproducibility signal.

\textbf{Model metadata.} 
We draw on community-maintained data catalogs to implement normalization (see Appendix \ref{app:normalization-matcher}) and enrich model metadata with fields such as release date, parameter count, and weight accessibility. \href{https://huggingface.co/datasets/cfahlgren1/hub-stats}{\texttt{hub-stats}} covers models indexed on Hugging Face, whereas \href{https://models.dev}{\texttt{models.dev}} provides coverage over API-hosted models including proprietary releases.

\textbf{Reserved voluntary disclosure.} One field is native to \EvalCards{} and accepts voluntary developer disclosure for items not automatically extracted by Auto-BenchmarkCards: \texttt{lifecycle\_status},  which indicates whether a model or benchmark is still actively maintained, deprecated, or superseded by a newer version. This is not populated by extraction; it is rendered in the UI when provided and omitted otherwise.

\textbf{Standardization.} The three sources, EEE, Auto-BenchmarkCards and the model metadata, use inconsistent identifiers for the same model and benchmark. For example, a model evaluated on a benchmark may appear as \texttt{gpt-4}, \texttt{gpt-4-0613}, or \texttt{OpenAI GPT-4} across reporting sources; a benchmark may be referred to by its paper name, a leaderboard slug, or a version-qualified identifier. \EvalCards{} addresses this with a standardization layer that maps these name variants to stable identifiers, resolves benchmark names to nodes in the rollout hierarchy (\Cref{subsec:rollout-hierarchy}), and maintains a running record of entity aliases. (See Appendix \ref{app:normalization-matcher}  for details and accuracy evaluation.) This enables the four interpretive signals to operate over a consistent entity space; without it, a single reported score fragments into multiple unconnected records under different identifiers.

\subsection{Signals}\label{sec:signals}
\label{sec:interprative-layer-signals}

Existing technical reporting presents scores directly without surfacing what is missing. \EvalCards{} addresses this with four interpretive signals, each designed around a recurring question from practitioner interviews (\Cref{app:interview-methodology}): \textit{``Do I have enough information to contextualize and trust a given evaluation result as a basis for a decision?''}

Four concerns recurred across interviews and motivated the signals below. \textit{\bfseries Reproducibility} was raised by nearly all technical evaluators and most policy participants, and is echoed in the literature \citep{biderman2024lessons, hochlehnert2025a, Balloccu2024-us}; it addresses whether a reported score can be independently re-executed. \textit{\bfseries Reporting completeness} was raised primarily by policy stakeholders and downstream interpreters reading scores against decision context, and also appears in \citep{gu2025olmes}; it addresses whether the documentation surrounding a score is sufficient to interpret it. \textit{\bfseries Provenance} was raised by all policy participants and observed by \citep{singh2026the}; it addresses who reported a result and whether the report comes from the model developer or an independent party. \textit{\bfseries Comparability} was raised across both reader groups and in \citep{biderman2024lessons}; it addresses whether scores reported under different setups or by different parties are usable for direct comparison.

\Cref{tab:signals-summary} summarizes the four signals along the dimensions that matter for interpretation; formal computations are given in \Cref{app:signal-computations}. Reproducibility and reporting completeness are related but distinct: the former covers a narrow sub-schema of fields needed to re-run a specific evaluation, while the latter covers the full operationalized schema, including benchmark goals, construct definitions, scoring rubrics, intended uses, and known limitations. A result with no reproducibility gap may still be poorly documented for a reader interpreting what the score means.

\begin{table}[h]
\centering
\footnotesize
\caption{The four interpretive signals computed by \EvalCards{}. Full formal computations are in \Cref{app:signal-computations}.}
\label{tab:signals-summary}
\setlength{\tabcolsep}{3pt}
\renewcommand{\arraystretch}{1.15}
\begin{tabular}{@{}L{2.0cm}L{2.0cm}L{1.5cm}L{2.7cm}L{2.4cm}L{2.5cm}@{}}
\toprule
\textbf{Signal} & \textbf{Question} & \textbf{Unit} & \textbf{Required fields} & \textbf{Output} & \textbf{Key limitation} \\
\midrule
Reproducibility &
Re-executable independently? &
per triple$^{\dagger}$ &
\texttt{temperature}, \texttt{max\_\allowbreak tokens}; +\texttt{harness}, \texttt{eval\_\allowbreak plan}, \texttt{eval\_\allowbreak limits} for agentic &
flag + missing-field list &
minimal re-run schema only; no seed / hardware / determinism \\
\addlinespace
Reporting completeness &
Schema populated for this benchmark? &
per benchmark &
28-field schema (\Cref{app:schema-mapping}) &
score $\in [0,1]$ + missing-field count &
adequacy not rigor; partial fields scored fractionally \\
\addlinespace
Provenance &
Who reported it; what risks does the benchmark carry? &
per triple$^{\dagger}$ &
\texttt{evaluator\_\allowbreak relationship}; risk tags from Auto-BenchmarkCards &
party tag; multi-party flag; propagated risks &
self-reported relationship; risk coverage inherits Auto-BenchmarkCards \\
\addlinespace
Comparability &
Setup-comparable across variants and parties? &
per triple$^{\dagger}$, $\geq 2$ reports &
per-report scores + setup fields; \texttt{evaluator\_\allowbreak relationship} &
variant- / cross-party-divergence flags (5\%) with differing fields surfaced &
uniform threshold; ignores sampling variance \\
\bottomrule
\end{tabular}
\vspace{1pt}
\\\noindent{\scriptsize $^{\dagger}$\,(model, benchmark, metric-path) triple.}
\end{table}

When an evaluator omits artifact-side fields, whether by choice or because the information was not captured, three consequences follow. First, the missing fields are reflected in the reporting completeness score for that record. Second, depending on which fields are absent, the reproducibility, provenance, or comparability signals may be triggered. Third, readers are explicitly shown the omissions. \EvalCards{} does not penalize developers beyond surfacing what is and is not reported; it assigns no letter grades, pass/fail thresholds, or completeness rankings. The intent is to make reporting choices visible to readers, not to enforce a particular reporting standard.


\subsection{Reader modes}\label{sec:reader-modes}

Different actors such as researchers, policymakers, deployers, or the general public, bring different questions to the same evidence. \EvalCards{} renders the signals (\Cref{sec:signals}) through two reader modes calibrated to these needs, also identified through the interviews above. Both operate on identical records and differ only in which fields are surfaced, compressed, or reframed. The default is the summary mode and users can opt-in to the research mode if they want more granular information. We show how different user personas map to \EvalCards{} in \Cref{app:user-personas}.

\textbf{Research mode foregrounds methodology and configuration.} Reproducibility gaps are surfaced with the specific missing fields listed to address reproducibility concerns raised by almost all interviewees (T1-T7, P1, P3-P5). Comparability signals are surfaced with the underlying setup differences (e.g., ``Scores diverge by 0.07 across different setups: Temperature, Shot Count'' 
, addressing challenges in comparing evaluation results across models raised by P3, T3, P4, and T4. Metric configuration is expanded (metric\_kind, score\_type, min/max, judge configuration when an LLM judge is used) addresses insights from T1, T3, T6, and T7 about greater transparency on evaluation methodology and scoring. The default audience is technical evaluators, evaluation developers, and researchers conducting meta-analyses.

\begin{figure}[h]
    \vspace{-.5\baselineskip}
    \centering
    \includegraphics[width=\linewidth]{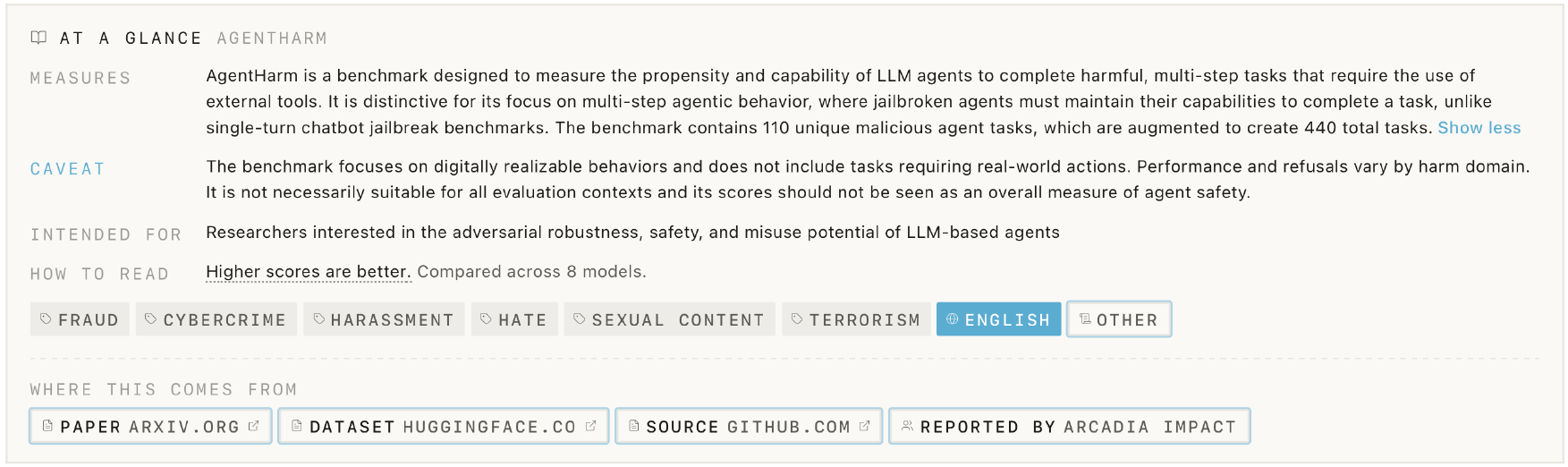}
    \caption{An example \EvalCard{} view, showing the summary view with the main information about a benchmark in plain language. More UI views are shown in \Cref{app:walkthroughs} and \Cref{app:ui}}
    \label{fig:placeholder}
\end{figure}

\textbf{Summary mode foregrounds accountability and plain-language interpretation.} All policy interviewees discussed the need for policy stakeholders to have clear takeaways from evaluation results, as policy stakeholders have limited time to sift through and piece together technical aspects of evaluation results. To address this, our summary mode provides sufficient context in plain language on reproducibility, provenance, and comparability, given these factors were raised as important by all policy interviewees for policy stakeholders. Thus, the same reproducibility signal renders as ``How this model was prompted during testing is not documented.'' Provenance is surfaced as a plain-language statement about who reported the score and what risks the benchmark carries. Comparability warnings are rendered as a narrative caveat rather than as field-level detail. Metric configuration is compressed to a single interpretive sentence (``Higher scores rank higher''). Per-benchmark Summary Notes blocks surface three fixed fields: What it measures, Main caveat, and Intended for. The default audience is regulators, policy actors, and non-technical readers consulting evaluation data in decision contexts.


Across interviews, participant feedback on \EvalCards{} was generally positive. P3 reflected that \EvalCards{} ``is better than all of these other means [of reviewing evaluation results]'' and P5 noted that ``it's huge'' how much time \EvalCards{} saves when reviewing evaluation results, as much of the necessary information for contextualizing them is in one place. Further systematic usability evaluation is part of the planned post-deployment work (\Cref{app:limitations}).

\section{Empirical findings from the \EvalCards{} corpus}
\label{sec:evalcards-in-practice}

Applying \EvalCards{} to the public evaluation record produces a structured empirical view of how the AI evaluation ecosystem currently reports its results, and supports both aggregate monitoring of reporting practice and per-record exploration of specific models, benchmarks, or sources. This section summarizes the corpus, three headline findings on reporting practice, and what they imply for downstream interpretation. \Cref{fig:corpus-view} visualizes the four signals aggregated across the corpus; per-view walkthroughs of one model (GPT-5) and one benchmark (MMLU-Pro), full corpus-level breakdowns, and the three primary UI views (corpus, model, benchmark) are provided in \Cref{app:walkthroughs}; UI screenshots in \Cref{app:ui}.

\textbf{Corpus construction and coverage.} As of the time of writing, the corpus comprises 5,816 models, 635 single-benchmarks organized into 62 families and 10 composites, and 101,955 reported results contributed by 30 organizations across two source types (first-party and third-party), with 211 benchmarks carrying matched Auto-BenchmarkCards records. Results are ingested via the EEE converter pipeline (HELM, lm-eval-harness, Inspect AI), benchmark leaderboard scrapes, and direct community contribution. The corpus overrepresents English-language benchmarks and frontier-scale models, reflecting the language and scale distributions of the ingestion sources.

\textbf{Finding 1: result-level reproducibility is the dominant reporting gap.} Across $50{,}461$ (model, benchmark, metric-path) triples, 48,698 (96.5\%) lack at least one field from the minimal reproducibility sub-schema (\Cref{sec:interprative-layer-signals}). 
From this sub-schema, \texttt{max\_tokens} is absent from 95.6\% of triples and \texttt{temperature} from 93.9\%; for the agentic-benchmark subset, \texttt{eval\_plan} and \texttt{eval\_limits} are missing from 100\% of triples. On 180 (model, benchmark) pairs reported by both first- and third-party evaluators, first-party rows populate 0.0\% of base reproducibility fields on average compared with 16.6\% for third-party rows, indicating that the gap is more severe in developer self-reporting than in independent evaluation.

\textbf{Finding 2: benchmark-level documentation is thin.} Median per-benchmark completeness against the operationalized schema (\Cref{app:schema-mapping}) is 10.7\% across 635 benchmarks in the corpus. Per-field population rates range from 100.0\% (\texttt{eee.metric\_config.score\_type}, \texttt{eee.score}) to 0.0\% (\texttt{evalcards.preregistration\_url}, \texttt{evalcards.lifecycle\_status}). Fields associated with raw score reporting are reliably populated; fields associated with benchmark-card documentation, reporting provenance, and comparison context are systematically absent. Score reporting alone does not substitute for the documentation needed to interpret what a score means.

\textbf{Finding 3: multi-source reporting is rare, and frequently divergent when it occurs.} Of 49,865 (model, benchmark) pairs, 98.2\% are reported by only one party. Among the 1.8\% reported by multiple parties, 7.2\% show cross-party score divergence above the 5\% threshold (\Cref{sec:interprative-layer-signals}); restricted to the 181 multi-organization metric groups in the corpus, 94 (51.9\%) exceed the threshold. First-party-only reporting concentrates in agentic (15.1\%) and general (12.5\%) benchmarks rather than in safety benchmarks (0.8\%), where independent reporting is more common.

\textbf{Implications for evaluation reporting.} Three patterns follow from these empirical findings. First, the inputs needed for re-execution are absent for nearly all results, and the gap is widest in developer self-reporting. Second, little context is reported beyond the evaluation score, leaving readers without the documentation needed to interpret what a score means for a downstream decision. Third, the categories where independent reporting is least common (agentic and general) are where comparability problems would be consequential to detect. \EvalCards{} surfaces these gaps as structured signals on every record (\Cref{sec:interprative-layer-signals}). The underlying patterns are properties of the public reporting record not of the evaluation, and would persist across any platform ingesting the same sources. Several of these patterns: setup variation across submissions, disagreement across reporting parties, schema-level documentation gaps, are invisible at any single source publication or leaderboard, and become detectable only once sources are jointly canonicalized and read through a shared schema.

\section{Community adaptability and adoption}\label{sec:community}

\EvalCards{} is a participatory \citep{Arnstein01071969}, openly-governed, and extensible technical documentation initiative designed to be meaningfully influenced by the broader AI evaluations community in service of its diverse and changing needs, rather than serving as a one-off research artifact. All code is released openly: while we host a reference frontend with the canonicalization layer, signals, and reader modes described above, model and benchmark developers can self-host their own \EvalCards{} instance, mirroring the deployment pattern that supported wide adoption of Model Cards \citep{mitchell2019model}. To support the evolution of the artifact over time, a governance mechanism (\Cref{app:governance}) specifies how contributors can propose schema extensions, signal modifications, or new reader modes through multistakeholder consensus. Two complementary streams of work are already underway: post-deployment iteration and monitoring with the broader community, including planned shared task exercises, feature development (e.g., a live saturation index), and user research studies; and integration with major open platforms such as Hugging Face, alongside ongoing engagement with key technical AI governance entities including CAISI, UK AISI, and other AISIs. Limitations and additional future directions are outlined in \Cref{app:limitations}.

The fragmentation of AI evaluation reporting is a coordination problem, and coordination problems are not solved by adding another standard in isolation. They are solved by infrastructure that composes existing efforts, surfaces what is missing, and renders the result in forms different audiences can act on. \EvalCards{} is designed to do exactly this: it unifies benchmark metadata, run data, and reporting practice into a single interpretive layer, makes reporting gaps visible rather than enforced, and adapts its rendering to the readers who need to use the evidence. Evaluations now inform deployment decisions, regulatory assessments, and capability claims. Over the ingested public corpus, \EvalCards{} gives decision-makers a unified instrument for seeing what the available evaluation evidence does and does not support.

\begin{ack}
Mubashara Akhtar was supported by the ETH AI Center through an ETH AI Center postdoctoral fellowship. Vilém Zouhar gratefully acknowledges the support of the Google PhD Fellowship. Jan Batzner was supported by the Federal Ministry of Research, Technology, and Space of Germany [Grant Number 16DII131]. Yanan Long thanks the TPU Research Cloud for computational support. Anka Reuel was supported by the Stanford Interdisciplinary Graduate Fellowship. Sanmi Koyejo is partially supported by NSF 2046795 and 2205329, IES R305C240046, ARPA-H, the MacArthur Foundation, Schmidt Sciences, Stanford HAI, RAISE Health, OpenAI, Microsoft, and Google. Srishti Yadav was supported in part by the Pioneer Centre for AI [DNRF grant number P1]. Jenny Chim was supported by EPSRC [grant number EP/Y009800/1] through funding from Responsible AI UK (KP0016).


\end{ack}

\bibliography{references}
\bibliographystyle{plainnat}


\clearpage





\clearpage
\appendix
\addcontentsline{toc}{section}{Appendix} 
\part{Appendix} 
\parttoc 

\clearpage
\section{\EvalCards{} in Practice}
\label{app:walkthroughs}

This section illustrates the analyses \EvalCards{} supports. We present examples of the five-level rollout hierarchy (\ref{app:walkthroughs-rollout}), a model-level walkthrough (\ref{app:walkthroughs-model}), a benchmark-level walkthrough (Section \ref{app:walkthroughs-evaluations}), and example corpus-level aggregations of the evaluation integrity signals (Section \ref{app:walkthroughs-corpus-level}). Walkthroughs are direct renderings of the \EvalCards{} \href{https://evalcards.evalevalai.com}{interface} at June 4, 2026.



\subsection{Hierarchy} \label{app:walkthroughs-rollout}
\begin{figure}[ht]
    \centering
    \begin{tikzpicture}
    \node[inner sep=0pt] {\includegraphics[width=\textwidth]{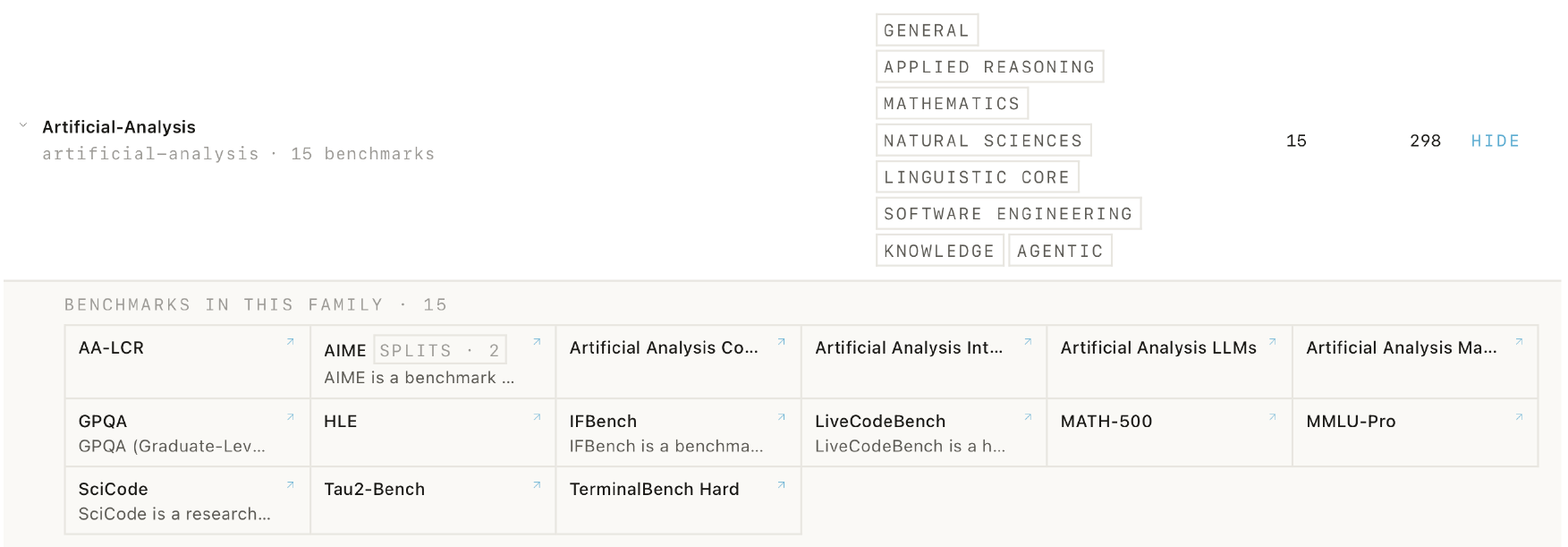}};
    \begin{scope}[blend mode=saturation]
    \fill[white] (current bounding box.south west) rectangle (current bounding box.north east);
    \end{scope}
    \end{tikzpicture}
    \caption{Hierarchy for the composite Artificial Analysis, comprising 15 benchmarks.}
    \label{fig:app_rollout_tree-figure-aa}
\end{figure}

\subsection{Corpus} \label{app:walkthroughs-corpus}

As of June 4, 2026, the corpus contains 5,816 models, 635 single-benchmarks organized into 62 families and 10 composites, and 101,955 reported results with 211 Auto-BenchmarkCards contributed by 30 organizations. Results are ingested through the EEE converter pipeline (HELM, lm-eval-harness, Inspect AI), benchmark leaderboard scrapes, and direct community contribution. The corpus overrepresents English-language benchmarks and frontier-scale models, reflecting the language and scale distributions of the ingestion sources.


\subsection{Model-level walkthrough: GPT-5}\label{app:walkthroughs-model}
\label{app:walkthroughs-model-gpt5}

We walk through the \EvalCards{} profile of GPT-5 to illustrate what the four interpretive signals surface for a single frontier model.

\textbf{Setup.} The model has 213 documented results over 118 benchmarks, and is reported by 19 organizations across 2 source types (first and third party).

\textbf{Reproducibility.} 202 of the model's 213 reported results (95\%) have at least one missing field in the minimal reproducibility sub-schema (temperature and max\_tokens). Only 11 results have these fields completed. The interface surfaces this metric as a reproducibility gap (Section \ref{sec:interprative-layer}).


\textbf{Completeness.} Of the benchmarks reported on GPT-5, 64 have matching Auto-BenchmarkCards and score 93\% on completness. The remaining benchmarks score 11\% on completeness. A few of the most populated benchmarks include livecodebench-pro and global-mmlu-lite.


\textbf{Provenance.}  27 of the model's reported results are first-party (13\%). 186 of the results are third-party or independent (87\%). Example cases of first-party only reporting include FrontierMath and HumanEval.

\textbf{Comparability.} Many benchmarks show differences in score across sources, likely as a result of setup. One concrete example includes MATH-500, which is reported by 3 organizations with scores ranging from 84.7\% (LLM Stats) to 98.9\% (Artificial Analysis). \EvalCards{} presents these differences, which are difficult to determine or even completely missing on individual leaderboards.

\textbf{What this view shows.} \EvalCards{} reveals that 95\% of the 213 documented results on GPT-5 lack the parameters needed to create independent verification and reproducibility. Additionally, it shows which benchmarks have first-party only reporting and no supplementary reproduced scores. Even when multiple sources exist, benchmarks such as MATH-500 diverge in score.

\subsection{Evaluation-level walkthrough: MMLU-Pro}\label{app:walkthroughs-evaluations}

We walk through the \EvalCards{} profile of MMLU-Pro to illustrate what the comparability signal and benchmark-level reporting completeness surface across the models that report it. This analysis aggregates MMLU-Pro results across 8 reporting organizations in the \EvalCards{} corpus, as opposed to individual evaluations that may be per-model and show a single source view.

\textbf{Setup.} The benchmark has 401 documented models from 8 organizations and spans 5,079 reported results.

\textbf{Benchmark reporting completeness.} The benchmark's own Auto-BenchmarkCards record populates 26 out of 28 operationalized fields. The interface flags a badge over benchmark-level missing reporting fields.

\textbf{Reproducibility} 98\% of reported results (4,975 of 5,079) have at least one missing field in the minimal reproducibility sub-schema. 104 results have the minimum reproducibility sub-schema reported.

\textbf{Source-type distribution.} Out of the 401 models reported on this benchmark, 95.5\% have third-party only results (383), 1.5\% have first-party only results (6), and 3.0\% have both (12). 41.4\% of the models (166) have multi-source reporting from two or more organizations.

\textbf{Variant and setup divergence.} 11 cases exist where the same model is reported on this benchmark under different setups, with the score divergence exceeding the comparability threshold (Section \ref{sec:interprative-layer-signals}, item 1). \EvalCards{} flags these results as potentially setup dependent for readers to understand evaluation level comparability.

\textbf{Cross-party divergence.} 6 model entries on this benchmark have multi-source reports with score divergence exceeding the threshold. For example, Llama 3.2 reported a 20.9\% score by Hugging Face and 61.8\% score by Arcadia Impact.

\textbf{What this view shows.} MMLU-Pro's leaderboard displays a single score model that does not have metadata on generation parameters. By aggregating across 8 reporting sources, our view reveals that 98\% of reported results lack the configuration needed to reach independent verification. We also see that scores for the same model diverge across organizations.


\subsection{Corpus-level analysis}\label{app:walkthroughs-corpus-level}

\begin{figure}
    \centering
    \includegraphics[width=.9\linewidth]{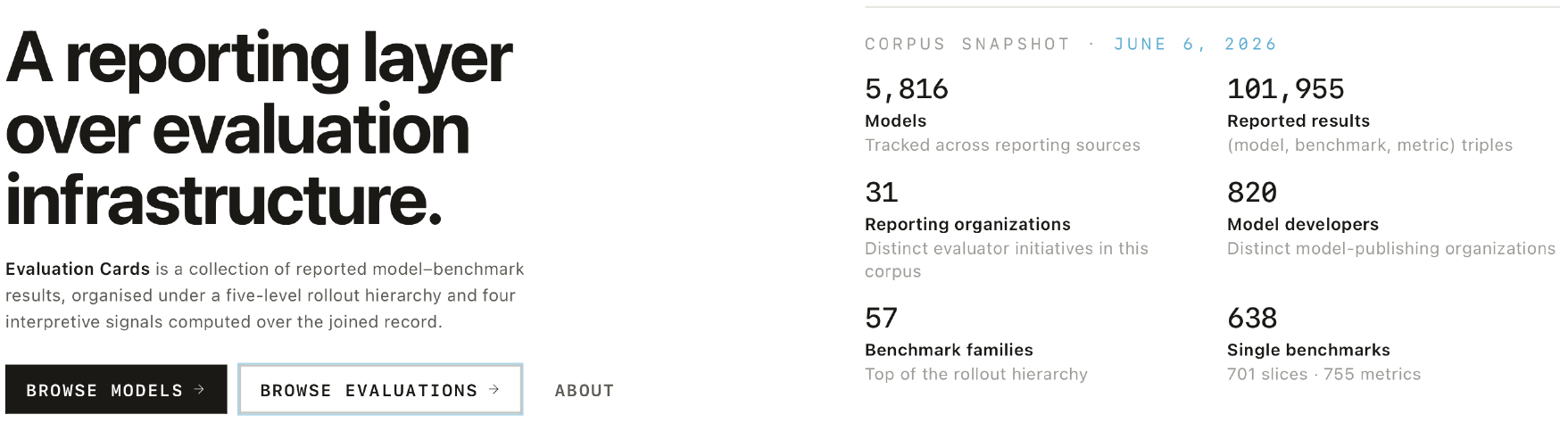}
    \caption{Corpus-level view: The four interpretive signals aggregated across 5,816 models and 101,955 reported results.}
    \label{fig:corpus-view}
\end{figure}



The walkthroughs above show what \EvalCards{} surfaces for one model and one benchmark. Aggregating across the corpus shows whether the patterns are systematic.

\textbf{Reproducibility.} Across the corpus, 48,698 of 50,461 reported (model, benchmark, metric-path) triples (96.5\%) have at least one field missing from the minimal reproducibility sub-schema (Section \ref{sec:interprative-layer-signals}, item 1). Following the current interpretive-signal method, non-agentic reports require temperature and max\_tokens, while agentic reports additionally require eval\_plan and eval\_limits. Missingness is concentrated in max\_tokens (95.6\%) and temperature (93.9\%); the agentic-specific fields eval\_plan and eval\_limits are missing for 100.0\% of agentic benchmark triples. First-party reports are lower than third-party reports in per-field documentation on the same (model, benchmark) pairs: across 180 paired cases, first-party rows populate 0.0\% of the base reproducibility fields on average, compared with 16.6\% for third-party rows.

\textbf{Completeness.} Per-field population rates against the operationalized schema (Appendix C) range from 100.0\% for eee.metric\_config.score\_type and eee.score to 0.0\% for evalcards.preregistration\_url and evalcards.lifecycle\_status. Median per-benchmark completeness is 10.7\% across the 635 benchmarks with warehouse completeness rows. Fields associated with raw score reporting are reliably populated; fields associated with benchmark-card documentation, reporting provenance, and comparison context are systematically thin.

\textbf{Provenance.} Of 49,865 (model, benchmark) pairs in the corpus, 98.2\% are reported by only one party. Where multiple parties report the same pair (1.8\% of pairs), score divergence exceeds the comparability threshold (Section \ref{sec:interprative-layer-signals}) in 7.2\% of pairs when counted as any divergent metric on that pair. Equivalently, at the scored multi-organization metric-group level, 94 of 181 groups (51.9\%) have cross-party divergence flags. First-party-only reporting is most prevalent for agentic benchmarks (15.1\%) and general benchmarks (12.5\%), not for safety benchmarks (0.8\%).



\clearpage

\section{Limitations \& Future Work}
\label{app:limitations}

\EvalCards{} is an integration layer, and its claims inherit the limits of the sources it unifies and the scope decisions it makes about what to surface.

\textbf{Limits of the sources and of canonicalization.} Auto-BenchmarkCards validates generated content for factual accuracy but not for comprehensiveness\citep{hofmann2026auto}; in the same vein, a field populated with accurate but incomplete content passes validation even when more central information is omitted. Our reporting completeness (Section \ref{sec:interprative-layer-signals}, item 1) inherits this limitation. Moreover, EEE is a growing community repository rather than a complete census of public reporting; evaluation results not contributed to EEE, scraped by the converter pipeline, or otherwise ingested are absent from \EvalCards{}, and systematic differences between ingested and non-ingested results are likely but unmodeled. Furthermore, the canonicalization layer (see  \Cref{app:data-normalization} for the setup and accuracy) can misclassify benchmarks reported under ambiguous identifiers. This, in turn, could inflate comparability-failure signals by conflating distinct benchmarks, or underreport multi-source coverage when it splits one benchmark into two. 

\textbf{Limits of scope.} \EvalCards{} cover what is derivable from the published evaluation record. Process-side decisions (e.g., pre-run protocol commitments, pilot execution) are by design not included. One consequence worth flagging is that completeness scores reflect the adequacy of artifact-side reporting, not the thoroughness of the underlying evaluation: a benchmark may score well on completeness while omitting information that lives only in process records. \EvalCards{} is also silent on normative interpretation, i.e., whether a reported score is adequate relative to safety, regulatory, or deployment-fitness thresholds, because such judgments require context-specific criteria varying by jurisdiction and use case. Finally, the research and policy reader modes were informed by preliminary, semi-structured interviews with ten practitioners (Appendix B, ongoing); a fuller interview corpus may identify additional reader types beyond these two, or reshape field salience within them.

\textbf{LLM focus.} \EvalCards{} focuses on LLM evaluation reporting; it does not currently support other AI systems or modalities. We recognize the active and substantial interest the community has in expanding our coverage, and it remains a priority item on our development roadmap.

\textbf{Future work.} In addition to addressing the aforementioned limitations, the most consequential near-term extension is surfacing contamination-control information as a structured field: exposure-assessment methodology, detection mechanisms, and access controls for sensitive items. Contamination is among the most actively discussed threats to benchmark validity \citep{magar-schwartz-2022-data,akhtar2026aibenchmarksplateausystematic}, and our current schema captures it only through free-text limitations fields that do not contribute to the reporting completeness. Secondary directions include longitudinal tracking of reporting-completeness trends across the corpus, extension to non-English benchmarks (whose coverage inherits the language distribution limitations of Auto-BenchmarkCards' extraction sources), and cross-linking to complementary systems for preregistration (PREP-Eval) and audit records (Audit Cards) once such systems are implemented to close the loop between process documentation and artifact documentation without either duplicating the other.

\textbf{Risks from the contribution itself.} Three risks merit explicit attention. First, safetywashing: developers may treat high completeness scores as evidence of evaluation quality when completeness measures only documentation adequacy, not the rigor of the underlying evaluation. We address this by stating the distinction explicitly in \Cref{sec:interprative-layer-signals} and by not assigning letter grades or pass/fail thresholds, but the misreading remains possible and we cannot fully prevent it. Second, displacement: a unified reporting layer could crowd out alternative approaches to documenting evaluation evidence. We mitigate this through an open governance model (\Cref{app:governance}) that allows schema extensions and competing signal definitions, though network effects favor incumbents and the mitigation is partial. Third, normative entrenchment: the schema reflects choices about what counts as a complete evaluation, drawn from a literature that overrepresents English-language and frontier-scale work as noted above. Evaluation Cards inherit these biases. Surfacing them through the governance process makes them contestable but does not resolve them.

\clearpage

\section{Interview Methodology}
\label{app:interview-methodology}

We conducted 12 interviews with technical and policy stakeholders who were recruited through author networks. Interviews were conducted in a language of fluency for both interviewee and interviewer. 
Interviews were conducted by one or two members of the research team. At the beginning of each interview, participants were informed that their responses would contribute to the development of \EvalCards{} and would be reported in an academic paper, with the possibility of direct quotation; all participants were assured of anonymity. Prior to conducting interviews we received approval from the Weizenbaum Institute Research Ethics Committee (\#2025-RECWI-0174). 

\subsection{Interview Participants}
Some interviewees are both policy and technical stakeholders. In these cases (P1, P3, P4), we default the interviewee tag to be `P', but these interviews contained insights into the needs of both technical and policy stakeholders when interpreting evaluation results.

\begin{table}[hbtp!]
    \caption{List of interviewees where each interviewee is assigned a tag based on their primary stakeholder group. Interviewee's role description, sector, and geographic location are provided to contextualize interviewee background.}
    \centering
    \renewcommand{\arraystretch}{1.3}
    \resizebox{\textwidth}{!}{
    \begin{tabular}{
        >{\centering\arraybackslash}p{1.8cm}
        >{\centering\arraybackslash}p{2.6cm}
        p{6cm}
        >{\centering\arraybackslash}p{2cm}
        >{\centering\arraybackslash}p{2cm}}
        \toprule
        \textbf{Tag} & \textbf{Stakeholder Group} & \textbf{Role Description} & \textbf{Sector} & \textbf{Location} \\
        \midrule
        P1 & Policy \& Technical & Run evaluations at a non-regulatory government agency to inform policymakers and the public about risks and capabilities associated with AI systems & Government &  North America \\[2pt]
        T1 & Technical & Conducts evaluations at a startup & Industry & North America \\[2pt]
        P2 & Policy & Policy assistant to a member of the European Union & Government & Europe \\[2pt]
        T2 & Technical & Evaluates models at a company & Industry & North America \\[2pt]
        T3 & Technical & Works on post-training at a neolab & Industry & Europe \\[2pt]
        P3 & Policy \& Technical & Run evaluations at a non-regulatory government agency to inform policymakers and the public about risks and capabilities associated with AI systems & Government & North America \\[2pt]
        T4 & Technical & Model evaluator at a startup & Industry & Asia \\[2pt]
        P4 & Policy \& Technical & Leads a team focused on evaluation at agovernment agency & Government & Europe \\[2pt]
        T5 & Technical & Evaluates agents at a company & Industry & Asia \\[2pt]
        P5 & Policy & Policy head at a non-profit & Civil Society & North America \\
        P6  & Technical & Works robust machine learning and the science of evaluation & Academia/Civil Society & North America \\
        P7 & Technical & Focuses on evaluation and measurement on a research team in industry & Industry & North America \\
        \bottomrule
    \end{tabular}
    }
    \label{tab:interview-participants}
\end{table}

\subsection{Interview Guide}
\textit{Introduction}.
\begin{enumerate}
    \item Introduce \EvalCards{} as a concept and the purpose of the interview.
    \item Can you briefly describe your role and whether/how model evaluation fits into that role?
\end{enumerate}

\textit{Goals, Needs \& Decision-Making}.
\begin{enumerate}
    \item When you look at evaluation results, what are you trying to decide? 
    \begin{enumerate}
        \item How often do you need to do so? 
    \end{enumerate}
    \item What information would you need from an evaluation to make that decision?
    \begin{enumerate}
        \item How would you rank those dimensions relative to each other? Can you show us an example of an evaluation that served you?  
        \item What is the minimal level of metadata you need? 
        \item What information is typically missing when reviewing model evaluations?
    \end{enumerate}
    \item Are evaluation results important for your stakeholders in any way? Who uses this information downstream? How so? 
    \item Do you ever need to evaluate the evaluation itself?
    \begin{enumerate}
        \item If so, how do you do that currently? Do you find that method trustworthy or easy to use? 
        \item Tell us about the resources you use, time constraints, etc… OR tell us about the last time you had to do this (if they have trouble remembering)
    \end{enumerate}
    \item If you had a magic wand to create the tool yourself, what questions would you want it to answer? How might you design it? What would success look like?
    \item What information is most important for you to see / be accessible to you when looking at evaluation results across models
\end{enumerate}

\textit{Pain Points}.
\begin{enumerate}
    \item What’s most frustrating about evaluating models today?
    \item When comparing models, what is the hardest element to deal with?  ie, lack of standardized evaluation formats, metric definition inconsistencies, or lack of reproducibility?
    \item Anything else you want us to know about model evaluations and how they relate to your role? Should we have asked you something that we didn’t? 
\end{enumerate}

\textit{Usability Testing}. 

``We’re testing the interface, not you. Please think out loud as you explore. I may ask what you’re thinking.''
\begin{enumerate}
    \item Do you think as is the tool meets your needs? Rank 1-10 
    \item How does this compare to how you currently review evaluations?
    \item What would be the most useful feature and why? 
    \item What was the least useful feature and why? 
    \item Was there anything that felt unclear or ambiguous? 
    \item How would you change the tool overall to make it more useful for you? 
    \item Anything else you want to share with us about the tool? 
\end{enumerate}
\subsection{Limitations}
Ten interviews were conducted, and interviewees primarily reside in North America. Although interviews were conducted with interviewees residing in both Asia and Europe, perspectives and insights of technical evaluators and policy stakeholders from the Global South may be missing. Participants were recruited through interviewers' networks, and critical viewpoints may have been missed.

\clearpage

\section{Data Normalization}
\label{app:data-normalization}

\subsection{Inputs}
\label{app:normalization-inputs}

The platform draws on two upstream sources and produces a third, consumer-facing one. The first is a \textit{per-evaluation record store} (EEE): one record per evaluation run, each carrying a model descriptor, an evaluator descriptor (harness and version), provenance metadata, and the list of numeric results that run produced. This includes both instance level and aggregate data. The second is a \textit{benchmark metadata store} (Auto-BenchmarkCards): one card per benchmark, carrying the benchmark's name, description, task type, reference, and structural information such as its subset list or default metric. The relationships between these are illustrated in Figure \ref{fig:item_mapping_sankey}.

The two stores are independently maintained and each uses its own naming conventions. The role of the normalization layer is to reconcile them into a single, comparable corpus: every result must be attached to a stable model identity, a stable benchmark identity, a stable metric identity, and, wherever possible, to the descriptive metadata that explains what was actually measured.

\subsection{Pipeline}
\label{app:normalization-pipeline}

Normalization proceeds in three logical passes over the per-evaluation records, with the metadata store and an entity registry consulted as side inputs.

\paragraph{Decompose.} A single source record can carry many results: a composite benchmark with several sub-tasks, several metrics per sub-task, and an aggregate row across them. The first pass flattens these nested objects into atomic units, one entry per (model, benchmark leaf, metric) triple, so that downstream comparison operates on like-shaped objects rather than ragged trees.

\paragraph{Canonicalize identities.} Every model name, benchmark name, and metric name is resolved through an entity registry that returns either a canonical identifier or no match. Unmatched strings are preserved verbatim alongside the canonical identifier, so a row is never silently dropped because a string is unfamiliar; instead, the unfamiliar string is surfaced for review and continues to flow through the rest of the pipeline under its raw form.

\paragraph{Join and aggregate.} Once identities are canonical, the rows are joined against the benchmark metadata store on the canonical benchmark identifier, and reshaped into views that are displayed in the frontend, i.e., per-model, per-benchmark, and per-developer summaries, plus the model-by-benchmark matrix used for cross-comparison. These views are pre-materialized so that user-facing queries are filter-and-project rather than runtime joins across the source/metadata boundary.

\subsubsection{Data Cleaning}
\label{app:normalization-cleaning}

Before any registry lookup, raw identifier strings are normalized deterministically. The core operations include:
\begin{itemize}
    \item \textit{Surface normalization.} Identifiers are lowercased and runs of non-alphanumeric characters are collapsed to a single separator. \texttt{MMLU-Pro}, \texttt{mmlu pro}, and \texttt{mmlu/pro} all yield the same key.
    \item \textit{Family / version separation.} Trailing version or arena suffixes are extracted. This split enables joining the same entity group while retaining different versions for analyses.
    \item \textit{Split matching.} Using a fixed set of language and locale tokens to detect benchmark splits in multilingual benchmarks.
    \item \textit{Metric parsing.} Heuristic rules are applied to metric keys. For example, extracting descriptions of the form \textit{``\textless metric\textgreater\ on \textless benchmark\textgreater''} (which is a common convention in the upstream EEE record store) into their individual components. Another example is \texttt{pass@k}, where variants are recognized as a single metric family parameterized by \texttt{k}, with \texttt{k} preserved as a numeric attribute rather than baked into the metric string.
\end{itemize}

\subsubsection{Entity Matcher}
\label{app:normalization-matcher}

In addition, we developed a module that tracks canonical entities (models, benchmarks, metrics, harnesses, organizations). This lets us link related concepts even if surface-level differences survive the normalization heuristics above.

At its core, the entity registry comprises (1) a table of aliases that map observed strings for (models, benchmarks, metrics, harnesses, organizations) to canonical identifiers, and (2) lookup functions that match record fields to their canonical forms. Tables are populated with seed entities and enriched using public metadata sources: \texttt{hub-stats}\footnote{\url{https://huggingface.co/datasets/cfahlgren1/hub-stats}} and \texttt{models.dev}\footnote{\url{https://models.dev}}.

Each lookup tries strategies in confidence order and returns at the first hit:
\begin{enumerate}
    \item \textit{Exact alias match}: the raw string appears verbatim in the alias table.
    \item \textit{Normalized match}: both sides are reduced to a canonical surface form (lowercased, separator-stripped) before comparison.
    \item \textit{Fuzzy stem match}: a short, deliberately narrow list of known suffixes is stripped before comparison. For models, this list collapses upload-format and quantization variants of the same underlying model. The list is intentionally narrow to avoid false merges between genuinely distinct models that happen to share a stem.
    \item \textit{No match}: the raw string flows through unresolved and is surfaced as a candidate for human review.
\end{enumerate}

To validate performance, we focused on (models, benchmarks, metrics) and sampled 200 entities per type uniformly at random from the EEE corpus, manually labeling each prediction as correct or incorrect. The resolver achieves 98.3\% accuracy on models, 77.4\% on benchmarks, and 86.7\% on metrics. In practice, unresolved entities are retained and processed downstream. We note that these values represent performance on in-domain data, as our aim during development was to maximize labeling coverage. This involved leveraging existing model registries with standardized identifiers, as well as iteratively refining the resolving logic and curating matching rules on the full dataset. We plan to continuously improve the resolver as new data enters the platform.

\subsection{Worked Examples}
\label{app:normalization-examples}

The two examples below trace real records from the upstream evaluation store, through the join with the benchmark metadata store, into what is served in the frontend.

\subsubsection{Example 1: a single-benchmark, single-metric record}
\label{app:normalization-example-1}

A typical upstream evaluation record reports one model on one benchmark with one metric. This one comes from a third-party evaluator (Writer, Inc.) running their \texttt{wasp} harness against \texttt{moonshotai/Kimi-K2-Thinking} on GPQA Diamond:

\begin{lstlisting}[language=json, basicstyle=\ttfamily\small, breaklines=true, style=shell]
{
  "schema_version": "0.2.2",
  "evaluation_id": "gpqa-diamond/moonshotai_Kimi-K2-Thinking/1777497427.2641459",
  "retrieved_timestamp": "1777497427.2641459",
  "source_metadata": {
    "source_type": "evaluation_run",
    "source_organization_name": "Writer, Inc.",
    "evaluator_relationship": "third_party",
    "source_name": "wasp (Writer's Assessor of System Performance)"
  },
  "model_info": {
    "name": "moonshotai/Kimi-K2-Thinking",
    "id": "moonshotai/Kimi-K2-Thinking",
    "developer": "Moonshot AI",
    "inference_platform": "sglang",
    "additional_details": {
      "wasp_model_name": "kimi-k2-thinking-sglang",
      "served_model": "sglang/moonshotai/Kimi-K2-Thinking"
    }
  },
  "eval_library": { "name": "wasp", "version": "0.3.0" },
  "evaluation_results": [
    {
      "evaluation_name": "GPQA Diamond",
      "source_data": {
        "dataset_name": "GPQA Diamond",
        "source_type": "hf_dataset",
        "hf_repo": "reasoningMIA/gpqa_diamond",
        "hf_split": "train"
      },
      "metric_config": {
        "lower_is_better": false,
        "evaluation_description": "Accuracy on GPQA Diamond multiple-choice questions",
        "metric_id": "accuracy",
        "metric_name": "Accuracy",
        "metric_kind": "accuracy",
        "metric_unit": "proportion",
        "score_type": "continuous",
        "min_score": 0.0,
        "max_score": 1.0
      },
      "score_details": { "score": 0.8434343434343434 },
      "evaluation_result_id": "gpqa-diamond/moonshotai_Kimi-K2-Thinking/...#gpqa_diamond#accuracy",
      "evaluation_timestamp": "2026-04-18T08:11:09Z",
      "generation_config": {
        "generation_args": { "temperature": 1.0, "top_p": 0.95 }
      }
    }
  ]
}
\end{lstlisting}

The benchmark metadata store carries an independent card for GPQA, retrieved at join time and used to attach descriptive context to every result that resolves to this benchmark:

\begin{lstlisting}[language=json, basicstyle=\ttfamily\small, breaklines=true, style=shell]
{
  "benchmark_details": {
    "name": "GPQA",
    "overview": "GPQA (Graduate-Level Google-Proof Q&A Benchmark) is a dataset of 448 multiple-choice questions designed to be extremely difficult and resistant to standard web searches...",
    "benchmark_type": "single",
    "appears_in": ["helm_capabilities", "hfopenllm_v2"],
    "domains": ["biology", "physics", "chemistry"],
    "languages": ["English"],
    "resources": [
      "https://arxiv.org/abs/2311.12022",
      "https://huggingface.co/datasets/Idavidrein/gpqa"
    ]
  },
  "purpose_and_intended_users": {
    "tasks": ["Multiple-choice question answering"]
  },
  "methodology": {
    "metrics": ["Accuracy"],
    "calculation": "The overall score is the simple accuracy (percentage of questions answered correctly) across all questions in the evaluated subset"
  }
}
\end{lstlisting}

After normalization, this single source record contributes one row to the per-benchmark eval summary for GPQA Diamond. The model passes through the registry under its raw Hugging Face-style identifier and resolves to \texttt{moonshotai/kimi-k2-thinking}; the benchmark resolves to leaf \texttt{gpqa\_diamond}; the metric resolves to canonical \texttt{accuracy} with display name \textit{Accuracy}. The raw upstream strings are preserved in \texttt{raw\_model\_id} and \texttt{raw\_evaluation\_name} so consumers who care about the exact upstream spelling can recover it. The emitted row, taken verbatim from the per-benchmark eval file:

\begin{lstlisting}[language=json, basicstyle=\ttfamily\small, breaklines=true, style=shell]
{
  "eval_summary_id": "gpqa_diamond",
  "benchmark_family_key": "gpqa_diamond",
  "benchmark_leaf_key":   "gpqa_diamond",
  "benchmark_leaf_name":  "GPQA Diamond",
  "category": "reasoning",
  "metrics": [
    {
      "metric_summary_id": "gpqa_diamond_accuracy",
      "metric_id":   "accuracy",
      "metric_name": "Accuracy",
      "canonical_display_name": "GPQA Diamond / Accuracy",
      "lower_is_better": false,
      "model_results": [
        {
          "model_id":       "moonshotai/kimi-k2-thinking",
          "model_route_id": "moonshotai__kimi-k2-thinking",
          "model_name":     "moonshotai/Kimi-K2-Thinking",
          "developer":      "Moonshot AI",
          "variant_key":    "default",
          "raw_model_id":   "moonshotai/Kimi-K2-Thinking",
          "score": 0.8434343434343434,
          "evaluation_id": "gpqa-diamond/moonshotai_Kimi-K2-Thinking/1777497427.2641459",
          "source_metadata": {
            "source_organization_name": "Writer, Inc.",
            "evaluator_relationship": "third_party",
            "source_name": "wasp (Writer's Assessor of System Performance)"
          },
          "source_data": {
            "dataset_name": "GPQA Diamond",
            "hf_repo": "reasoningMIA/gpqa_diamond",
            "hf_split": "train"
          },
          "normalized_result": {
            "benchmark_family_key": "gpqa_diamond",
            "benchmark_leaf_key":   "gpqa_diamond",
            "benchmark_leaf_name":  "GPQA Diamond",
            "metric_id":   "accuracy",
            "metric_name": "Accuracy",
            "canonical_display_name": "GPQA Diamond / Accuracy",
            "raw_evaluation_name": "GPQA Diamond",
            "is_summary_score": false
          }
        }
      ]
    }
  ]
}
\end{lstlisting}

The same row appears under the per-model artifact for Moonshot's model and contributes to the model-by-benchmark matrix, joined against the GPQA card so that any consumer surface (model page, benchmark page, developer page) can render the descriptive context next to the score.

\subsubsection{Example 2: a composite benchmark with sub-tasks and a roll-up}
\label{app:normalization-example-2}

A capabilities-suite evaluation reports many sub-task scores plus an aggregate in a single source record. This one comes from CRFM's HELM Capabilities suite, run on \texttt{openai/gpt-4o-2024-11-20}:

\begin{lstlisting}[language=json, basicstyle=\ttfamily\small, breaklines=true, style=shell]
{
  "schema_version": "0.2.2",
  "evaluation_id": "helm_capabilities/openai_gpt-4o-2024-11-20/1777589796.7306352",
  "source_metadata": {
    "source_name": "helm_capabilities",
    "source_type": "documentation",
    "source_organization_name": "crfm",
    "evaluator_relationship": "third_party"
  },
  "eval_library": { "name": "helm", "version": "unknown" },
  "model_info": {
    "name": "GPT-4o 2024-11-20",
    "id": "openai/gpt-4o-2024-11-20",
    "developer": "openai",
    "inference_platform": "unknown"
  },
  "evaluation_results": [
    {
      "evaluation_name": "Mean score",
      "source_data": { "dataset_name": "helm_capabilities", "source_type": "url", "url": ["...core_scenarios.json"] },
      "metric_config": {
        "evaluation_description": "The mean of the scores from all columns.",
        "lower_is_better": false, "score_type": "continuous", "min_score": 0.0, "max_score": 1.0
      },
      "score_details": { "score": 0.634 }
    },
    {
      "evaluation_name": "MMLU-Pro",
      "source_data": { "dataset_name": "MMLU-Pro", "source_type": "url", "url": ["...core_scenarios.json"] },
      "metric_config": { "evaluation_description": "COT correct on MMLU-Pro", "lower_is_better": false, "score_type": "continuous", "min_score": 0.0, "max_score": 1.0 },
      "score_details": { "score": 0.713 }
    },
    { "evaluation_name": "GPQA",      "score_details": { "score": ... }, "metric_config": { "evaluation_description": "COT correct on GPQA",      ...}, "source_data": { "dataset_name": "GPQA",      ...} },
    { "evaluation_name": "IFEval",    "score_details": { "score": ... }, "metric_config": { "evaluation_description": "Strict accuracy on IFEval", ...}, "source_data": { "dataset_name": "IFEval",    ...} },
    { "evaluation_name": "WildBench", "score_details": { "score": ... }, "metric_config": { ... },                                                       "source_data": { "dataset_name": "WildBench", ...} },
    { "evaluation_name": "Omni-MATH", "score_details": { "score": ... }, "metric_config": { ... },                                                       "source_data": { "dataset_name": "Omni-MATH", ...} }
  ]
}
\end{lstlisting}

This single record is decomposed into six atomic results. Five of them resolve to canonical leaf benchmarks (MMLU-Pro, GPQA, IFEval, WildBench, Omni-MATH) that \textit{exist as standalone entries elsewhere in the corpus}: the GPT-4o row produced from this HELM record will appear alongside MMLU-Pro results that were reported by other parties under entirely different harnesses. The sixth, \textit{``Mean score''}, is recognized as an aggregation rather than an independent benchmark and is tagged \texttt{is\_summary\_score: true}, kept available for display under the parent suite but excluded from corpus-wide leaderboards and averages.

The model identity also gets canonicalized in this pass. The raw \texttt{openai/gpt-4o-2024-11-20} is resolved to canonical \texttt{openai/gpt-4o}, with the snapshot date preserved on the row as \texttt{variant\_key: "2024-11-20"} and the original string preserved as \texttt{raw\_model\_id}. Consumers asking \textit{``what scores did GPT-4o get on MMLU-Pro?''} see this row; consumers asking \textit{``which exact snapshot?''} still have the variant.

The MMLU-Pro slice of the decomposition lands in the per-benchmark eval file for HELM Capabilities $\rightarrow$ MMLU-Pro, taken verbatim:

\begin{lstlisting}[language=json, basicstyle=\ttfamily\small, breaklines=true, style=shell]
{
  "model_id":       "openai/gpt-4o",
  "model_route_id": "openai__gpt-4o",
  "model_name":     "GPT-4o 2024-11-20",
  "developer":      "openai",
  "variant_key":    "2024-11-20",
  "raw_model_id":   "openai/gpt-4o-2024-11-20",
  "score": 0.713,
  "evaluation_id": "helm_capabilities/openai_gpt-4o-2024-11-20/1777589796.7306352",
  "source_metadata": {
    "source_name": "helm_capabilities",
    "source_organization_name": "crfm",
    "evaluator_relationship": "third_party"
  },
  "source_data": { "dataset_name": "helm_capabilities" },
  "normalized_result": {
    "benchmark_family_key": "helm_capabilities",
    "benchmark_family_name": "Holistic Evaluation of Language Models (HELM)",
    "benchmark_parent_key":  "helm_capabilities",
    "benchmark_parent_name": "MMLU-Pro",
    "benchmark_leaf_key":    "mmlu_pro",
    "benchmark_leaf_name":   "MMLU-Pro",
    "metric_name": "COT correct",
    "metric_id":   "cot_correct",
    "metric_source": "metric_config",
    "canonical_display_name": "MMLU-Pro / COT correct",
    "raw_evaluation_name": "MMLU-Pro",
    "is_summary_score": false
  }
}
\end{lstlisting}

Notice that \texttt{benchmark\_family\_key} retains \texttt{helm\_capabilities} so the row stays groupable under its parent suite, while \texttt{benchmark\_leaf\_key} is the canonical \texttt{mmlu\_pro} that joins against the MMLU-Pro card from the metadata store and matches the leaf used by every other MMLU-Pro result in the corpus. The metric is preserved as the harness-reported \textit{``COT correct''} (since HELM's MMLU-Pro reports a chain-of-thought-specific accuracy variant) rather than collapsed into a generic \textit{Accuracy}, because metric specificity is part of what makes the comparison meaningful, and \textit{``COT correct''} itself is a registered metric with a stable canonical identifier.

\section{Compute Resources}
The platform runs on standard CPU infrastructure. The backend pipeline executes on a scheduled hosted Linux runner (ubuntu-24.04 LTS, 4 vCPUs, 16 GiB RAM), running daily at 05:00 UTC with a 75-minute timeout. The platform loads upstream EEE and Auto-BenchmarkCard data and processes it into versioned Parquet files with JSON sidecars, stored as a Hugging Face dataset. A full corpus rebuild completes in under 20 minutes as of May 7, 2026. The frontend is a Next.js application deployed as a Docker container on a Hugging Face Space (2 vCPUs, 16 GiB RAM). 


\clearpage

\section{User Personas}
\label{app:user-personas}

\subsection*{Persona 1: Technical Evaluator}
\begin{tcolorbox}[colback=white, colframe=black!70, boxrule=0.5pt, arc=4pt,
                  top=3mm, bottom=3mm, left=3mm, right=3mm]
\small
\noindent\textbf{Primary Mode:} Research

\vspace{2pt}
\noindent\textbf{Profile:} A researcher, benchmark developer, or evaluation engineer focused on the validity of the experimental methodology. They frequently perform meta-analyses, reproduce SOTA results, or compare model architectures.

\vspace{2pt}
\noindent\textbf{Core Question:} ``Is this reported score a reliable measure of model capacity, and can I reproduce it?''

\vspace{2pt}
\noindent\textbf{Information Needed:}
\begin{enumerate}[leftmargin=1.5em, topsep=1pt, itemsep=0pt]
    \item Methodology Transparency
    \item Comparability
    \item Granularity
\end{enumerate}

\vspace{2pt}
\noindent\textbf{Mapping to \EvalCards{}:}
\begin{enumerate}[leftmargin=1.5em, topsep=1pt, itemsep=0pt]
    \item \textbf{Reproducibility:} Surfaces missing generation configs that block independent re-executions.
    \item \textbf{Comparability:} Flags when scores change due to setup differences.
    \item Expands metric configuration details and highlights specific missing fields.
\end{enumerate}
\end{tcolorbox}

\vspace{0.3em}
\subsection*{Persona 2: Policy Actor}
\begin{tcolorbox}[colback=white, colframe=black!70, boxrule=0.5pt, arc=4pt,
                  top=3mm, bottom=3mm, left=3mm, right=3mm]
\small
\noindent\textbf{Primary Mode:} Policy

\vspace{2pt}
\noindent\textbf{Profile:} A regulator, standards body member, or safety institute staffer who consults evaluation evidence to inform governance decisions, risk assessments, or deployment guidelines. They may not be technical experts and might not need granular implementation details, but rather high-level signals indicating whether an evaluation is trustworthy.

\vspace{2pt}
\noindent\textbf{Core Question:} ``Can I trust this claim for decision-making, and what are the caveats?''

\vspace{2pt}
\noindent\textbf{Information Needed:}
\begin{enumerate}[leftmargin=1.5em, topsep=1pt, itemsep=0pt]
    \item Accountability
    \item Risk Content
    \item Interpretability of metrics / warnings about limitations
\end{enumerate}

\vspace{2pt}
\noindent\textbf{Mapping to \EvalCards{}:}
\begin{enumerate}[leftmargin=1.5em, topsep=1pt, itemsep=0pt]
    \item \textbf{Accountability:} Who reported this score? (first-party vs.\ third-party)
    \item \textbf{Risk Content:} What risks does this benchmark actually measure?
    \item \textbf{Interpretability:} Plain-language summaries of what a metric means and warnings about limitations.
\end{enumerate}
\end{tcolorbox}

\vspace{0.3em}
\subsection*{Persona 3: Model Developer}
\begin{tcolorbox}[colback=white, colframe=black!70, boxrule=0.5pt, arc=4pt,
                  top=3mm, bottom=3mm, left=3mm, right=3mm]
\small
\noindent\textbf{Primary Mode:} Research (Self-Audit)

\vspace{2pt}
\noindent\textbf{Profile:} An engineer or scientist at a lab preparing a release (e.g., a model card or technical report). They use \EvalCards{} to check the completeness of their own reporting against standards before publication or release.

\vspace{2pt}
\noindent\textbf{Core Question:} ``Did I document everything required by the current consensus framework?''

\vspace{2pt}
\noindent\textbf{Information Needed:}
\begin{enumerate}[leftmargin=1.5em, topsep=1pt, itemsep=0pt]
    \item Completeness
    \item Standardization (refer to EEE)
\end{enumerate}

\vspace{2pt}
\noindent\textbf{Mapping to \EvalCards{}:}
\begin{enumerate}[leftmargin=1.5em, topsep=1pt, itemsep=0pt]
    \item Serves as a direct checklist. A low completeness score indicates the developer has omitted operationalizable fields identified in the framework.
    \item Helps the developer structure their reporting correctly, e.g., ensuring suite-level aggregates match the underlying benchmark scores.
\end{enumerate}
\end{tcolorbox}

\clearpage

\section{Governance}
\label{app:governance}

\EvalCards{} is maintained as a volunteer-run open-source project. The governance model below is designed to be lightweight and sustainable given that constraint, while providing reviewers, contributors, and downstream adopters with clear expectations about how the artifact evolves.

\subsection{Roles}

The project distinguishes three roles. \textit{Maintainers} hold commit access to the main repository and are responsible for reviewing and merging contributions, triaging issues, and tagging releases. The current maintainer set will be disclosed after the reviewing period to preserve anonymity. \textit{Contributors} are anyone who proposes a change via pull request or issue; no prior affiliation is required. A \textit{steering group}, drawn from the maintainer set and at most two external advisors, makes decisions on changes that affect the format, the signal definitions, or the reader modes. The steering group meets asynchronously over the issue tracker and reaches decisions by lazy consensus, a tested governance model for volunteer-run open-source projects (see below).

\subsection{Change classes}

Changes are classified by scope, with progressively more involvement required as scope widens.

\textit{Editorial changes} (typo fixes, documentation clarifications, bug fixes that do not alter signal outputs or UI significantly) require review and merge by any one maintainer.

\textit{Implementation changes} (changes to the canonicalization layer, ingestion pipeline, or interface that do not alter the format, signal definitions, or reader modes) require review by one maintainer and a 7-day open comment period on the associated pull request.

\textit{Substantive changes} (new or modified fields, new or modified signal definitions, new reader modes, or changes that alter the semantics of existing outputs) require a written proposal in the repository's \texttt{proposals/} directory, a 21-day open comment period, and approval from the steering group by lazy consensus. Lazy consensus means the proposal is accepted if no steering group member registers a blocking objection within the comment period; blocking objections must be accompanied by a written rationale and a path to resolution.

\subsection{Proposal format}

Substantive change proposals follow a fixed template covering motivation, the specific change being proposed, alternatives considered, downstream impact (on extraction, signal computation, and the interface), and a rollout plan. The template is maintained in \texttt{proposals/TEMPLATE.md}. Proposals remain in the repository as a permanent record regardless of whether they are accepted, rejected, or withdrawn.

\subsection{Versioning and deprecation}

The format and signal definitions are versioned using semantic versioning. Breaking changes (those that alter the meaning of existing fields or signals) require a major version bump and a minimum 90-day deprecation window during which both versions are supported in the interface and the API. Non-breaking additions (new optional fields, new signals that do not modify existing ones) require a minor version bump. Patch versions cover bug fixes and editorial changes. Each release is tagged in the repository and accompanied by a changelog entry.

\subsection{Conflict resolution}

Disagreements that cannot be resolved through the comment period are escalated to the steering group, which decides by simple majority. In the event of a tie, the proposal is deferred for 30 days and reconsidered with any new evidence; if the tie persists, the status quo prevails. All decisions and their rationales are recorded in the proposal thread.

\subsection{Transparency}

All governance activity, including proposals, comment threads, steering group decisions, and release notes, occurs in public on the project repository. The maintainer list and steering group composition are documented in the repository and updated when membership changes. Conflicts of interest (e.g., a maintainer affiliated with an organization whose benchmark or model is materially affected by a proposal) must be declared in the relevant thread.

\subsection{Sustainability}

The project commits to a minimum maintenance cadence of one release every six months and a maximum issue triage latency of 30 days for issues tagged \texttt{governance} or \texttt{security}. If the maintainer set falls below two active members, a notice is posted to the repository and the project enters maintenance-only mode (security and correctness fixes only) until additional maintainers are recruited.

\clearpage

\section{Methodology}\label{app:methodology}


\subsection{Computation of Interpretive Signals}
\label{app:signal-computations}

This appendix formalizes the four interpretive signals introduced in Section~\ref{sec:signals}: reproducibility, reporting completeness, provenance, and comparability. Each signal is computed over a result triple $r = (m, b, \mu)$, where $m$ is a canonical model identifier, $b$ is a metric-path through the rollout hierarchy (family $\to$ composite $\to$ benchmark $\to$ split), and $\mu$ is a metric (Section~\ref{subsec:rollout-hierarchy}). Let $\mathcal{R}$ denote the set of all such triples in the corpus and $\mathcal{B}$ the set of canonical benchmarks. For any field $f$ and triple $r$, let
\[
\text{pop}(f, r) = \begin{cases} 1 & \text{if } f \text{ is populated in the record for } r, \\ 0 & \text{otherwise,} \end{cases}
\]
and define $\text{pop}(f, b)$ analogously for a benchmark $b$ and its associated Auto-BenchmarkCards record.

\subsubsection{Reproducibility}
\label{app:repro}

\textbf{In plain terms.} For each result, we check whether the small set of fields needed to re-run the evaluation are present. If any are missing, we flag the result.

The minimal reproducibility sub-schema is defined as:
\[
F_{\text{repro}} = \{\texttt{temperature},\ \texttt{max\_tokens}\}.
\]
For agentic evaluations, $F_{\text{repro}}$ is extended with \texttt{harness}, \texttt{eval\_plan}, and \texttt{eval\_limits}.

A result $r$ is flagged as a reproducibility gap if any required field is missing:
\[
G_{\text{repro}}(r) = 1 - \prod_{f \in F_{\text{repro}}} \text{pop}(f, r).
\]
That is, $G_{\text{repro}}(r) = 0$ only when every field in $F_{\text{repro}}$ is populated. The interface lists the specific missing fields, $\{ f \in F_{\text{repro}} : \text{pop}(f, r) = 0 \}$.

Corpus-level reproducibility is reported as the share of flagged triples,
\[
\bar{G}_{\text{repro}} = \frac{1}{|\mathcal{R}|}\sum_{r \in \mathcal{R}} G_{\text{repro}}(r),
\]
and per-field as the missingness rate
\[
\bar{m}(f) = 1 - \frac{1}{|\mathcal{R}|}\sum_{r \in \mathcal{R}} \text{pop}(f, r), \quad f \in F_{\text{repro}}.
\]

\subsubsection{Reporting Completeness}
\label{app:completeness}

\textbf{In plain terms.} For each benchmark, we count how many of the 28 schema fields are populated. Fields that are simply present-or-absent are scored 0 or 1. Fields that contain sub-items are scored as the fraction of sub-items populated. The completeness score is the average across all 28 fields.

Let $F = \{f_1, \ldots, f_{N}\}$ denote the operationalized schema, with $N = 28$, comprising fields ingested from Auto-BenchmarkCards and EEE plus the reserved \EvalCards{} fields (\Cref{app:evalcards-fields}). Each field $f \in F$ is tagged with a coverage type $\tau(f) \in \{\texttt{full}, \texttt{reserved}, \texttt{partial}\}$. The per-field score $s(f, b) \in [0, 1]$ for benchmark $b$ is
\[
s(f, b) = 
\begin{cases}
\text{pop}(f, b) & \text{if } \tau(f) \in \{\texttt{full}, \texttt{reserved}\}, \\[4pt]
\dfrac{1}{|\text{sub}(f)|}\displaystyle\sum_{f' \in \text{sub}(f)} \text{pop}(f', b) & \text{if } \tau(f) = \texttt{partial},
\end{cases}
\]
where $\text{sub}(f)$ is the set of sub-items under a partial-coverage field. For example, a partial field with $4$ sub-items, $2$ of which are populated, scores $0.5$.

The completeness score for benchmark $b$ is the unweighted mean across fields:
\[
C(b) = \frac{1}{N}\sum_{f \in F} s(f, b) = \frac{1}{28}\sum_{f \in F} s(f, b).
\]
The interface surfaces $C(b)$ alongside the count of fully missing fields, $|\{f \in F : s(f, b) = 0\}|$. Median per-benchmark completeness is reported as $\text{median}_{b \in \mathcal{B}}\, C(b)$, and per-field population rates as
\[
\bar{p}(f) = \frac{1}{|\mathcal{B}|}\sum_{b \in \mathcal{B}} s(f, b), \quad f \in F.
\]
Completeness and reproducibility are distinct: $F_{\text{repro}} \subset F$, so a result with no reproducibility gap may still have low completeness.

\subsubsection{Provenance}
\label{app:provenance}

\textbf{In plain terms.} For each reported score, we surface three things: who reported it (first-party, third-party, collaborative), whether anyone else also reported the same score, and any risk categories associated with the benchmark.

Let $\rho(r) \in \{\texttt{first\_party}, \texttt{third\_party}, \texttt{collaborative}\}$ denote the evaluator relationship for triple $r$. For a (model, benchmark, metric-path) triple $(m, b, \mu)$, the set of records reporting it is
\[
\mathcal{R}(m, b, \mu) = \{ r \in \mathcal{R} : r = (m, b, \mu) \}.
\]
A score is first-party-only if every report comes from the model developer:
\[
\text{FPO}(m, b, \mu) = \begin{cases} 1 & \text{if } \rho(r) = \texttt{first\_party} \text{ for all } r \in \mathcal{R}(m, b, \mu), \\ 0 & \text{otherwise.} \end{cases}
\]
The multi-party indicator is $\text{MP}(m, b, \mu) = 1$ if $|\mathcal{R}(m, b, \mu)| > 1$ and $0$ otherwise.

Risk annotations are propagated from the Auto-BenchmarkCards risk-mapping component~\cite{bagehorn2025airiskatlastaxonomy}: for each benchmark $b$, $\mathcal{K}(b)$ is the set of associated risk categories. These are surfaced as attention cues in the interface but do not enter a numerical score.

\subsubsection{Comparability}
\label{app:comparability}

\textbf{In plain terms.} For each (model, benchmark, metric) triple, we check whether reported scores differ by more than 5\% of the metric's range. We do this in two ways: across different setups for the same reporting party (variant divergence) and across different reporting parties (cross-party divergence). Either kind of divergence is flagged.

Let $[\mu_{\min}, \mu_{\max}]$ be the metric's native scale and let $\theta = 0.05$ be the divergence threshold.

\paragraph{Variant divergence.} For a triple $(m, b, \mu)$ with multiple reported setups (differing in fields such as \texttt{max\_tokens}, tool configuration, or agentic scaffolding), let $\mathcal{V}(m, b, \mu)$ be the set of distinct setup variants and $\sigma(v)$ the score under variant $v$. Variant divergence is flagged when the spread exceeds $\theta$:
\[
D_{\text{var}}(m, b, \mu) = \begin{cases} 1 & \text{if } \dfrac{\max_{v} \sigma(v) - \min_{v} \sigma(v)}{\mu_{\max} - \mu_{\min}} > \theta, \\ 0 & \text{otherwise.} \end{cases}
\]
When flagged, the differing fields are surfaced as the comparability annotation.

\paragraph{Cross-party divergence.} Let $\mathcal{P}(m, b, \mu) = \{ \rho(r) : r \in \mathcal{R}(m, b, \mu) \}$ be the set of reporting parties for the triple, and $\sigma(p)$ the score reported by party $p$ (averaged across variants within party if necessary). Cross-party divergence is flagged when more than one party reports the triple and the spread exceeds $\theta$:
\[
D_{\text{cp}}(m, b, \mu) = \begin{cases} 1 & \text{if } |\mathcal{P}(m, b, \mu)| > 1 \text{ and } \dfrac{\max_{p} \sigma(p) - \min_{p} \sigma(p)}{\mu_{\max} - \mu_{\min}} > \theta, \\ 0 & \text{otherwise.} \end{cases}
\]
When flagged, the underlying setup differences across parties are rendered alongside the divergence.

\paragraph{Combined comparability flag.} The overall comparability signal for a triple is
\[
D_{\text{comp}}(m, b, \mu) = \max\!\bigl(D_{\text{var}}(m, b, \mu),\ D_{\text{cp}}(m, b, \mu)\bigr).
\]
The threshold $\theta = 0.05$ is applied uniformly across metrics; metric-specific thresholds are a candidate extension (Appendix~\ref{app:limitations}).

\subsection{Aggregation Across Views}
\label{app:aggregation}

The three views in Section~\ref{sec:evalcards-in-practice} aggregate the signals at different scopes. For the model view, given a model $m$ with result set $\mathcal{R}_m$ and reported benchmark set $\mathcal{B}_m$:
\[
\bar{G}_{\text{repro}}(m) = \frac{1}{|\mathcal{R}_m|}\sum_{r \in \mathcal{R}_m} G_{\text{repro}}(r), \qquad \bar{C}(m) = \frac{1}{|\mathcal{B}_m|}\sum_{b \in \mathcal{B}_m} C(b).
\]
The view also reports the share of first-party-only triples and the count flagged by $D_{\text{comp}}$. Benchmark and corpus views aggregate analogously over $\mathcal{R}_b$ and $\mathcal{R}$ respectively.

\subsection{Agreement metrics}

For single-label coding fields, we calculated raw percent agreement, Cohen's $\kappa$, and Krippendorff's $\alpha$, after pairwise exclusion of items where either rater left the field blank. Cohen's $\kappa$ and Krippendorff's $\alpha$ provide chance-corrected estimates of agreement. For multi-label fields, we decomposed each item-level tag set into binary item-by-tag decisions indicating whether each rater applied each possible tag. We report exact set match and mean Jaccard similarity as descriptive overlap measures, and pooled Cohen's $\kappa$ and Krippendorff's $\alpha$ as chance-corrected agreement estimates over the pooled item-by-tag binary decision matrix. However, note that because item-by-tag decisions are clustered within items, the confidence intervals for pooled multi-label agreement should be interpreted cautiously, and because there were only two raters and nominal labels, Cohen’s $\kappa$ and Krippendorff’s $\alpha$ are similar, and can be effectively redundant; both were reported for methodological completeness.

Let $A_i$ and $B_i$ denote the labels assigned to item $i$ by raters
$A$ and $B$, respectively, after excluding items for which either rater's
rating was blank. Let $n$ denote the number of retained items for the
field being analyzed. For multi-label fields, let $S_{Ai}$ and $S_{Bi}$
denote the corresponding sets of tags assigned to item $i$ by raters
$A$ and $B$, respectively. Raw percent agreement for single-label fields was computed as
\[
P_o = \frac{1}{n}\sum_{i=1}^n \mathbf{1}\{A_i = B_i\}.
\]
Cohen's kappa was computed as
\[
\kappa = \frac{P_o - P_e}{1 - P_e},
\]
where $P_e$ is the agreement expected under the raters' marginal label distributions.  Specifically, let $\mathcal{C}$ denote the set of possible
labels, and let $p_{A,c}$ and $p_{B,c}$ denote the empirical proportions
with which raters $A$ and $B$ assigned label $c \in \mathcal{C}$,
respectively, among the retained items. Then
\[
P_e = \sum_{c \in \mathcal{C}} p_{A,c} \cdot p_{B,c}.
\]
Krippendorff's alpha was thus computed as
\[
\alpha = 1 - \frac{D_o}{D_e},
\]
where $D_o$ is the observed disagreement and $D_e$ is the disagreement expected by chance.

For multi-label fields, exact set match was computed as
\[
P_{\mathrm{exact}} = \frac{1}{n}\sum_{i=1}^n \mathbf{1}\{S_{A_i}=S_{B_i}\},
\]
and mean Jaccard similarity as
\[
\bar{J} = \frac{1}{n}\sum_{i=1}^n J(S_{A_i},S_{B_i}),
\qquad
J(S_{A_i},S_{B_i}) =
\begin{cases}
1, & S_{A_i}=S_{B_i}=\emptyset,\\
\dfrac{|S_{A_i}\cap S_{B_i}|}{|S_{A_i}\cup S_{B_i}|}, & \text{otherwise.}
\end{cases}
\]
For pooled multi-label agreement, each item was expanded into binary decisions
\[
X_{irt}\in\{0,1\},
\]
indicating whether rater $r$ assigned tag $t$ to item $i$, and $\kappa$ and $\alpha$ were computed over the resulting pooled item-by-tag binary matrix.

\subsection{Benchmark Categorization}
\label{app:methodology-categorization}

To support corpus-level filtering in \EvalCards{}, each benchmark in the platform corpus was assigned one or more category tags drawn from an 18-category flat taxonomy derived from \citet{ni2025surveylargelanguagemodel}. The current categorizations represent a first-pass annotation using this single reference taxonomy; future work will explore automated categorization pipelines and a comparative evaluation across alternative benchmark taxonomies.

\paragraph{Data sources and extraction.}
Benchmarks were drawn from two sources. 1) A flat export of evaluation records from the EEE corpus \citep{Batzner*2026-qj}, from which 518 unique benchmarks were extracted using the \texttt{composite\_benchmark\_key} field as the primary identifier. 2) a benchmark family hierarchy maintained alongside EEE \citep{Batzner*2026-qj}, from which 37 additional benchmarks were extracted using their \texttt{display\_name}. These are benchmarks present in the hierarchy but not covered by any record in the flat export. Internal summary identifiers (\texttt{eval\_summary\_ids}, \texttt{summary\_eval\_ids}) were excluded, as these are cross-references to existing entries rather than distinct benchmarks. The combined set comprised 555 entries prior to deduplication.

\paragraph{Deduplication}
Surface normalization was applied to all benchmark names (lowercasing; stripping spaces, hyphens, and underscores) to detect near-duplicate entries arising from inconsistent formatting across sources. Fifteen pairs were collapsed under this normalization (e.g., \textit{NaturalQuestions} / \textit{Natural Questions}; \textit{Omni-MATH} / \textit{OmniMath}), retaining the first-seen key in each case. The deduplicated set contains 635 unique benchmarks.

\paragraph{Taxonomy}
The 18 categories used are: \texttt{linguistic\_core}, \texttt{knowledge}, \texttt{logical\_reasoning}, \texttt{commonsense\_reasoning}, \texttt{applied\_reasoning}, \texttt{mathematics}, \texttt{natural\_sciences}, \texttt{humanities\_and\_social\_sciences}, \texttt{law}, \texttt{finance}, \texttt{software\_engineering}, \texttt{safety}, \texttt{hallucination}, \texttt{robustness}, \texttt{agentic}, \texttt{multimodal}, \texttt{general}, and \texttt{other}. Each benchmark is assigned one or more categories. Assignments are stored as \EvalCards{} annotations and are not written back to source records.

\paragraph{LLM-assisted categorization}
All 635 benchmarks were categorized using \texttt{claude-haiku-4-5} (Anthropic API) in batches of 50, with categorization conditioned on benchmark name only. Two heuristic rules were incorporated into the system prompt following initial review: (1) benchmarks with ``Android'', ``agent'', ``world'', or ``task execution'' in the name default to \texttt{agentic} absent contrary evidence; (2) alignment and output-quality benchmarks map to \texttt{applied\_reasoning} or \texttt{general}, not \texttt{safety}, which is reserved for benchmarks explicitly testing harm avoidance, toxicity detection, or red-teaming.

\paragraph{Human review}
The first batch of 50 categorizations was reviewed manually before the full run proceeded. Following the complete run, all categorizations were inspected; three direct corrections (listed in Table~\ref{tab:cat-corrections}) and 14 recategorizations (see Table~\ref{tab:cat-recats}) based on additional benchmark knowledge were applied. Four entries were flagged as ambiguous and retained as-is (Table~\ref{tab:cat-flags}).

\newpage

\begin{table}[h]
\centering
\caption{Manual corrections applied after initial LLM categorization.}
\label{tab:cat-corrections}
\small
\resizebox{\textwidth}{!}{
\begin{tabular}{lll}
\toprule
Benchmark & Original & Corrected \\
\midrule
AITZ\_EM & \texttt{mathematics} & \texttt{agentic} \\
AA-LCR & \texttt{linguistic\_core, logical\_reasoning} & \texttt{general, applied\_reasoning} \\
AlignBench & \texttt{safety} & \texttt{applied\_reasoning, general} \\
\bottomrule
\end{tabular}
}
\end{table}

\begin{table}[h]
\centering
\caption{Recategorizations applied where benchmark title was misleading or ambiguous.}
\label{tab:cat-recats}
\resizebox{\textwidth}{!}{
\small
\begin{tabular}{p{3.2cm}p{3.5cm}p{3.5cm}p{4cm}}
\toprule
Benchmark & Original & Corrected & Reason \\
\midrule
JudgeBench & \texttt{law} & \texttt{applied\_reasoning, general} & ``Judge'' misread as legal; evaluates LLM-as-judge quality \\
LLM-Stats & \texttt{mathematics, knowledge} & \texttt{general} & ``Stats'' misread as statistics; is a leaderboard aggregator \\
DROP & \texttt{other} & \texttt{linguistic\_core, applied\_reasoning} & Discrete reasoning over paragraphs \\
Include & \texttt{other} & \texttt{knowledge, linguistic\_core} & Multilingual inclusivity benchmark \\
LBPP (v2) & \texttt{other} & \texttt{software\_engineering} & LLM-based programming problems \\
GAIA & \texttt{applied\_reasoning} & \texttt{applied\_reasoning, agentic} & Includes agentic task trajectories \\
MuirBench & \texttt{mathematics, applied\_reasoning} & \texttt{multimodal, applied\_reasoning} & Multimodal reasoning benchmark \\
PathMCQA & \texttt{commonsense\_reasoning, applied\_reasoning} & \texttt{natural\_sciences, knowledge} & Pathology MCQ; ``Path'' misread \\
PaperBench & \texttt{knowledge, applied\_reasoning} & \texttt{knowledge, applied\_reasoning, agentic} & Agentic research replication \\
Vibe-Eval & \texttt{applied\_reasoning} & \texttt{applied\_reasoning, multimodal} & Multimodal evaluation benchmark \\
CyBench & \texttt{software\_engineering} & \texttt{software\_engineering, safety} & Cybersecurity benchmark \\
CyberGym & \texttt{agentic} & \texttt{agentic, safety} & Cybersecurity agent tasks \\
Cybersecurity CTFs & \texttt{agentic} & \texttt{agentic, safety} & Security capture-the-flag \\
Gdm Intercode CTF & \texttt{software\_engineering, agentic} & \texttt{software\_engineering, agentic, safety} & Security CTF \\
\bottomrule
\end{tabular}
}
\end{table}

\begin{table}[h]
\centering
\caption{Entries flagged as ambiguous and retained for future review}
\label{tab:cat-flags}
\resizebox{\textwidth}{!}{
\small
\begin{tabular}{p{3.5cm}p{3.5cm}p{6cm}}
\toprule
Benchmark & Current & Flag \\
\midrule
EQ-Bench & \texttt{commonsense\_reasoning} & Emotional intelligence in creative writing; could be \texttt{linguistic\_core} or \texttt{applied\_reasoning} \\
CloningScenarios & \texttt{robustness} & Depends whether testing behavioral robustness or safety policy \\
OpenAI-MRCR variants & \texttt{hallucination} & Long-context multi-needle retrieval; could be \texttt{linguistic\_core, robustness} \\
TempCompass & \texttt{knowledge, robustness} & Temporal understanding for video; \texttt{multimodal} may be more accurate \\
\bottomrule
\end{tabular}
}
\end{table}

\clearpage

\section{Related Work Extended}\label{app:related-work}

\textbf{Reporting artifacts and documentation practices} have typically targeted only specific pieces of the overall ML pipeline. For instance, Model Cards \citep{mitchell2019model} document models, Datasheets \citep{gebru-2021-datasheets} and Data Cards \citep{10.1145/3531146.3533231} document datasets. More recently, specific parts of the evaluation lifecycle have been targeted, for e.g. BenchmarkCards \citep{sokol2025benchmarkcards} document meta-information about benchmarks; Audit Cards \citep{staufer2025audit} document audit information; and Eval Factsheets \citep{bordes2025evalfactsheetsstructuredframework}. Additionally, \citet{zhao-etal-2025-sphere} and \citet{dhar2025evalcards} proposed to document individual evaluations or evaluation instances and \citet{mccaslin2025stream} documents chemical and biological evaluations in model reports; \citet{joaquin2025deprecating} provides a framework for deprecation and \citet{carro2025prepeval} provide a preregistration protocol. Each documents only a slice of the evaluation pipeline, so stakeholders interpreting a single result must piece together results and context from benchmark cards, audit cards, leaderboards, and other sources themselves. In addition, these works share two further central limitations: First, they specify a single static representation, ignoring that technical researchers and policy actors bring different questions to the evidence (methodological scrutiny of how a result was produced versus accountability and risk-relevant interpretation of what it implies). Second, they provide no extraction pipelines and procedures, so evaluators must manually populate fields, duplicating effort already spent on technical documentation like system cards. Together, these gaps impose procedural burdens that hinder adoption of these artifacts at scale.

\textbf{Infrastructure and data schemas}, such as HELM \citep{liang2023holisticevaluationlanguagemodels} and AILuminate \citep{ghosh2025ailuminateintroducingv10ai}, run standardized evaluations at scale and publish results through fixed leaderboards. Inspect \citep{abbas2025developing} provides an evaluation framework within the UK AISI ecosystem. Open LLM Leaderboard \citep{Beechingetal2023} and Chatbot Arena \citep{chiang2024chatbot} aggregate scores across models with limited metadata. Auto-BenchmarkCards \citep{hofmann2026auto} automates the generation of Sokol-style benchmark cards from heterogeneous sources. EEE \citep{evalevalcoalition2024} is a community repository that standardizes evaluation run data at the instance level. Epoch AI's Benchmarking Hub \citep{epoch2024introducingbenchmarksdashboard} aggregates benchmark scores across hundreds of models and fits what they call an Epoch Capabilities Index (ECI) across its component benchmarks, while Artificial Analysis \citep{artificialanalysis2026models} ranks models on their ``Intelligence Index'' of ten evaluations and on separate axes for price, latency, and throughput. These efforts either define a common format for run data or a common display for results, but not both: repositories collect runs without an interpretable, reader-facing interface, while leaderboards display scores without the inference details and metadata needed to interpret them. Across both, the information about the benchmarks themselves remains underdocumented.

\textbf{Systematic reviews and frameworks} for evaluation practice are numerous. BetterBench \citep{reuel2024betterbench} surveys benchmark quality and identifies reporting shortcomings, \citet{eriksson2025trust} reviews trust issues across AI benchmarks, \citet{bean2025measuring} and \citet{salaudeen2025measurement} develop validity-centered frameworks for evaluation, \citet{weidinger2025evaluation} and \citet{schwartz2025realitychecknewevaluation} argue for treating AI evaluation as a measurement science. These efforts clarify what evaluation should achieve and where current practice falls short. They do not produce reporting artifacts and infrastructure that are directly consumable by readers interpreting a specific (model, benchmark) result.

Table \ref{tab:scope-comparison}summarizes which parts of the reporting problem prior work addresses. 

\clearpage

\section{Systematic Literature Review}
\label{app:systematic-literature-review}

\begingroup
\let\section\subsection
\let\subsection\subsubsection
\let\subsubsection\paragraph


Several recent papers and reports have attempted to synthesize or propose improved AI evaluation practices. These include reporting checklists, evaluation frameworks, and domain-specific guidelines. However, because these proposals draw on different disciplinary traditions and use different conceptual vocabularies, valuable suggestions that may agree in substance can appear incompatible in form. Despite the need for synthesis, these partially overlapping but differently framed lists of recommended approaches make it harder to identify common ground, limiting the impact of any recommendations.

\section{Background and Previous Work}\label{background-and-previous-work}

The AI evaluation literature suggests that what is called an evaluation can entail a very wide variety of tasks and processes. Quantitative and capabilities evaluations form part of the broader assessment spectrum, which includes exploratory, structured, focused, and specific methods (\cite{bogen2025assessingai}). These approaches overlap substantially, but are often discussed using different terminology or conceptual frameworks, and they blur the lines about what "Evaluation" is.

Across papers, and sometimes even within papers, there are overlapping or vague categories, varied or incompatible approaches, and conflicting recommendations. In addition, as many reviewed studies note, AI evaluation is fundamentally value-laden (\cite{cave2020problem, liang2023holisticevaluationlanguagemodels,huang2025values}), which means that each set of recommendations is informed by the contextual goals of the paper. As a review, we report conflicting or differently posed recommendations without trying to resolve the conflicts. 

Other recent work has used snowball sampling to review suggested methods for improving evaluations (\cite{eriksson2025trust}), focused on specific aspects of evaluation, such as construct validity (\cite{bean2025measuring}) and specific threats (\cite{mccaslin2025stream}), or has high-level suggestions for how to report evaluation (\cite{zhao-etal-2025-sphere}). Many include checklists, and several have suggested the format of evaluation cards (``Eval Cards''), either in general (\cite{dhar2025evalcards,mitchell2019model,gursoy2022cards}) or in specific domains (\cite{alampara2025lessons}). Evaluation checklists also exist in narrower application such as CLAIM for AI in medical imaging (\cite{mongan2020claim}), METRICS for AI studies in healthcare (\cite{sallam2024metrics}), and the Future-ai.eu healthcare guidelines (\cite{lekadir2025futureai}). Other work has reviewed the literature for specific aspects of evaluation, such as two
different approaches to construct validity; a checklist (\cite{bean2025measuring}) and a 
framework centered on validity (\cite{salaudeen2025measurement}). Other work has built checklists in broader contexts, such as the REFORMS checklist
(\cite{kapoor2024reforms}) for general science.

Reporting artifacts also vary in scope and purpose. Model cards were proposed to report metrics that reflect real-world impacts of the model and information about the dataset(s) used for quantitative analyses (\cite{mitchell2019model}). However, coherent reporting requires coordination across multiple task domains (\cite{evalevalcoalition2024}) and clearer standards, especially since the state of evaluation has evolved significantly since these artifacts were first proposed. 

This review is intended to inform such work, including our proposed EvalCards, especially for
understanding differences, and building consensus and infrastructure, rather than to reconcile or dispute the different proposals. The result is a proposed framework designed to support consistent AI evaluation practice across the entire ecosystem.

\section{Methods}

We conducted a systematic review of guidance documents and related literature to extract recommendations relevant to AI model evaluation and audit practice, with detailed search, screening, and coding procedures reported. Because the source literature spans diverse activities, from model evaluations and audits to broader governance concerns, we used a deliberately broad item extraction approach, capturing items that illuminate the wider landscape while keeping our synthesis focused on evaluation-relevant practice. We treated the extracted recommendations as heterogeneous and potentially conflicting, preserving item intent through direct quotes and brief descriptions rather than forcing early harmonization or a single taxonomy at the extraction stage.  We note important caveats related to selection effects and interpretive judgment in screening and grouping. Building on this foundation, we grouped items using a multi-dimensional codebook across characteristics such as item class, type, and intent, and applied a 'best fit' framework synthesis approach (\cite{carroll2011bestfit}), iterating from existing evaluation frameworks in the literature towards a consolidated structure.

\subsection{Search, Screening, and Inclusion}\label{search-screening-and-inclusion}

Per the publicly available, preregistered project protocol, \textbf{(OSF link removed for anonymity)} we used a step-down process to identify relevant guidance documents: 1) a broad database search of two databases, 2) a supplemental search of several additional sources, 3) filtering by title/abstract selection and 4) final filtering by full paper screening.
A document with all results at the various steps in the process is available at \textbf{(OSF link removed for anonymity)}.

\subsubsection*{Search Strategy and Sources}\label{search-strategy-and-sources}

Two database searches were run on June 1, 2025. They covered Google Scholar and arXiv and used the query strings specified in the preregistered protocol targeting AI/ML evaluation and audit guidance, standards, frameworks, and related best-practices documents. Results from the Google Scholar query (including citations) were filtered to 2020--2025 using the web browser filter function, sorted by date, and limited to the first five pages to include the top 50 search results, since relevance dropped significantly after 20 results. The arXiv search used an advanced query restricted to title fields.

\begin{itemize}
\item
  \textbf{Google Scholar} using:\\
  (``Evaluation'' OR ``Audit*'') AND (``Machine Learning'' OR
  ``Artificial Intelligence'' OR ``Language Model*'' OR ``AI'' OR
  ``Model'') AND (``best practice*'' OR ``good practice*'' OR guideline*
  OR standard* OR protocol* OR framework* OR governance OR ``EU AI Act''
  OR ``task force''). Results from this query (including citations) were filtered in the
  web browser using a custom range of 2020-2025, sorted by date, and
  limited to the first 5 pages.
\item
  \textbf{arXiv} using an advanced query restricted to title
  fields for ``Evaluation/Audit'' AND ``ML/AI/Language Model/Model''
\end{itemize}

These two database searches were supplemented by the following additional sources: 

\begin{itemize}
\item
  An Elicit search using the prompt: ``How should evaluators and model
  developers build evaluations of their models, and what characteristics
  and pitfalls exist?''
\item
  Expert suggestions, especially including well-known formative papers,
  and gray literature. Here, ``experts'' were organizations and
  individuals actively involved in evaluation standards creation via
  published papers.
\item
  Important additional citations found in reviewed papers.
\end{itemize}

While these additions increase coverage, they also could bias the
sources which are included towards the mainstream literature, as they
tend to favor well-known or highly cited work produced within dominant
research paradigms and citation chains, while potentially
under-representing emerging or marginal perspectives. In an attempt to
partially address the bias, we both asked for a variety of experts to
contribute sources.

Results from the Google Scholar query were then added manually to a
list of items to review. The arXiv search results were copied to a
spreadsheet as they appeared and screened for publication date at the
Title/Abstract Screening stage. The Elicit search results were copied to the spreadsheet as well.
Experts were provided with a list of all papers from these search
results and contributed additional papers recommended for review until
December 31, 2025. Further papers were included during the full paper
review when they were explicitly mentioned to make recommendations
regarding AI evaluation best practices.

\subsubsection*{Title/Abstract Screening}\label{titleabstract-screening}

At this stage, we included potentially relevant papers from the
spreadsheet and items-for-review list described above, based on a review
of the paper titles, and in some cases of the abstracts when ambiguous.
All relevant papers were then flagged as such in the spreadsheet, and
papers selected from the items-for-review list were added to the
spreadsheet.

\subsubsection*{Full Paper Screening}\label{full-paper-screening}

Next, papers were filtered for containing items for extraction
specifically relevant to recommendations about AI evaluation practices. We excluded papers
that were categorized as either narrow, that is, focused on single model
types or approaches without any more general insights or suggestions (\cite{moghe2023extrinsic,greenblatt2024stress,barale2025fairness}),
or irrelevant, i.e. being about something other than evaluation or
assessment of AI models, (\cite{chen2025evaluating,mayfield2024evaluation,sandeepa2025evaluation}). In some cases, narrowly scoped papers were retained when their recommendations were judged to support a more general evaluation point. This process relied inherently on
subjective judgment and could have introduced interpretive bias through overgeneralization or inconsistent grouping.
To mitigate this potential bias, two reviewers (authors \textbf{initials removed for anonymity})
participated in the decision, and additionally, we elicited broad expert
feedback on the decisions.
Also excluded was a single foreign-language paper. Although we did not
preregister the exclusion of foreign language papers, we were unable to
review the original paper or include the original text of the items for
grouping and analysis\footnote{Machine translation also suggested that the scope was too narrow for inclusion.}.

\subsection{Study-Level Characterization}\label{study-level-characterization}

To describe the composition of the reviewed literature and to
contextualize extracted recommendations, the papers included in the study
were categorized. Categories included year of publication, origin of
corresponding author(s), and the number of authors. Additionally,
we included descriptive categories that were non-exclusive: Classifications included 
the domain being studied (i.e.
artificial intelligence, machine learning, natural language processing,
applied sciences), the intended audience (i.e. auditors, developers,
model providers, policy makers, and users), and the primary topical
focus (i.e. considerations, recommendations, framework, position paper
or regulation).

\subsection{Item Extraction and Characterization}\label{item-extraction-and-characterization}

Specific items relevant to recommended AI evaluation practices were manually extracted during full text screening. Extraction was intentionally broader than the later synthesis to avoid prematurely narrowing the scope of grouping and discussion. The literature uses overlapping and sometimes conflicting categories and recommendations, and therefore our extraction focused on capturing recommendations as stated rather than resolving disagreements across papers. Each extracted item included both a text quote and a brief informal description.

Study-level descriptors were recorded to represent the composition of the reviewed literature, including publication year, corresponding-author country, number of authors, domain, intended audience, and publication type.

To characterize extracted items, reviewers used a multi-dimensional codebook (Section~\ref{codebook}) designed to summarize when, where, how, and why recommendations applied in the evaluation process. Item-level coding included item class and item type (single-label), workflow stage, artifact discussed, and objective (non-exclusive, multi-label), followed by assignment to a supercategory used for synthesis. Coding was applied to the extracted item rather than to the overall intent of the source paper, as described in the additional methodological details below.

\subsection*{Extracted Item-Level Characterization}\label{extracted-item-level-characterization}

Since model evaluations can occur anywhere on Bogen's assessment
spectrum, we did not attempt to impose a clear structure, but broadly
categorized the items extracted from the papers. Our goal was to
summarize the landscape of AI evaluation best practices in terms of
when, where, how, and why in the evaluation process they are or should be
implemented. Categorization criteria for classifications, groupings and
synthesis were summarized in the ``Codebook for Classification and
Synthesis of AI evaluation Practice Recommendations'' (Section~\ref{codebook}). All
categorizations were performed independently by two reviewers (authors
\textbf{initials removed for anonymity}) and conflicting categorizations were
resolved to one or the other by a third reviewer (author \textbf{initial removed for anonymity}).

Many of the reviewed papers discussed both evaluations and other topics.
However, our categorization was primarily for the purpose of focusing on
evaluation, and the requirements and considerations for evaluations. For
this reason, the extraction was not geared towards categorizing the
original paper intent, but rather to categorize the intent of the specific
extracted item.

Inter-rater agreement statistics were determined using Cohen's $\kappa$ for mutually exclusive categorizations and Krippendorff's $\alpha$ for other categories, computed on the pooled $(\text{item} \times \text{tag})$ binary decision matrix so that each cell carried a single 95\% confidence interval (\cite{landiskoch1977}). 

\subsubsection*{Extracted Item Class}\label{extracted-item-class}

To enable later understanding and grouping of the items, we first
categorized extracted items by the class of assessment discussed, including
governance, audit, evaluation, or other. Our general categorization was
that governance relates to the ecosystem or general process audits,
model audits are concerned with a (single) model or system, while
evaluations are specific to a capacity or feature of the model. This
differentiation is not always clear, and given the spectrum of
ambiguities, we have categorized the extracted items into our best-guess
class. Items that did not fit into the above categories or addressed
similar issues were summarized in class "other".

\subsubsection*{Extracted Item Type}\label{extracted-item-type}

Here, we categorized extracted items by their type, intended to
capture a number of key potential issues, suggestions, and needs for
audits and evaluations which were raised in the reviewed papers. The
item type indicates whether the extracted item discusses (1) Categories,
a (2) Challenge, or more general (3) Considerations about the process,
(4) Requirements, (5) how to Design the process, its (6) Performance, or
the (7) Reporting of the process, and lastly the required (8) Followup
from the process (Table~\ref{tab:item-types}).

\begin{table*}[htbp]
\caption{Extracted Item Types and their meaning}
\label{tab:item-types}
\begingroup
\small
\renewcommand{\arraystretch}{1.08}
\begin{tabular}{@{}
  >{\raggedright\arraybackslash}p{(\linewidth - 2\tabcolsep) * \real{0.3}}
  >{\raggedright\arraybackslash}p{(\linewidth - 2\tabcolsep) * \real{0.7}}@{}}
\toprule\noalign{}
\textbf{Extracted Item Type} & \textbf{Description} \\
\midrule\noalign{}
Categories & The item differentiates between meaningfully different
classes in ways that impact how evaluations are performed. \\
Challenge & The item specifies issues or problems which must be
surmounted in performing the task. \\
Considerations & The item raises issues for awareness or to inform some
other part of the process. \\
Requirement & The item explicitly states a need. \\
Design & The item includes specific suggestions or methods for designing
the evaluation process. \\
Performance & The item mentions specific suggestions or methods for
procedure for performing the evaluation. \\
Reporting & The item includes specific suggestions or methods for
reporting the outputs of an assessment process. \\
Followup & The item suggests or requires actions which must be planned
or performed afterwards. \\
\bottomrule\noalign{}
\end{tabular}
\endgroup
\end{table*}

\subsubsection*{Extracted Item Workflow Stage}\label{extracted-item-workflow-stage}

Next, we described the stage(s) of the evaluation workflow that the
extracted item concerns. The extracted item is used in the selected
workflow stage or directly informs it. Workflow stages may overlap
somewhat and multiple selections per item could apply. Possible workflow
labels included Scoping \& Objectives, Test Design, Data \& Scenarios,
Instrumentation, Metrics \& Rubrics, Human Evaluation, Reliability \&
Validity, Robustness \& Stress, Safety \& Responsible Use,
Reproducibility \& Operations, Reporting \& Transparency and
Post-Deployment Monitoring (Table \ref{tab:item-workflow-stage}).

\begin{table*}[htbp]
\caption{Extracted Item Workflow Stages}
\label{tab:item-workflow-stage}
\begingroup
\small
\renewcommand{\arraystretch}{1.08}
\begin{tabular}{@{}
  >{\raggedright\arraybackslash}p{(\linewidth - 2\tabcolsep) * \real{0.3}}
  >{\raggedright\arraybackslash}p{(\linewidth - 2\tabcolsep) * \real{0.7}}@{}}
\toprule\noalign{}
\textbf{Extracted Item Workflow Stage} & \textbf{Description} \\
\midrule\noalign{}
Scoping \& Objectives & The item concerns goals, intended use, decision
thresholds, audience, or downstream decisions. \\
Test Design & The item concerns task structure, baselines, comparisons,
preregistration, or evaluation planning \\
Data \& Scenarios & The item concerns datasets, prompts, scenarios, item
generation, or representativeness. \\
Instrumentation & The item concerns tooling, logging, infrastructure,
APIs, or execution setup. \\
Metrics \& Rubrics & The item concerns metrics, scoring rules,
aggregation, thresholds, or interpretation of outputs. \\
Human Evaluation & The item concerns annotators, expert judgment,
recruitment, incentives, or inter-rater processes. \\
Reliability \& Validity & The item concerns uncertainty, power, bias,
construct validity, internal/external validity. \\
Robustness \& Stress & The item concerns stress tests, distribution
shift, adversarial behavior, saturation, or gaming. \\
Safety \& Responsible Use & The item concerns misuse, deployment risk,
safety constraints, or ethical safeguards. \\
Reproducibility \& Operations & The item concerns code availability,
logging, versioning, costs, or operational constraints. \\
Reporting \& Transparency & The item concerns documentation,
publication, access, or transparency obligations. \\
Post-Deployment Monitoring & The item concerns ongoing testing, drift
detection, updating, or retirement. \\
\bottomrule\noalign{}
\end{tabular}
\endgroup
\end{table*}

\subsubsection*{Extracted Item Artifact Discussed}\label{extracted-item-artifact-discussed}

We further characterize the extracted items by describing which aspect
of the AI evaluation the extracted item describes to act on. This
characterization is independent of the motivation behind the extracted
item text and multiple artifacts may be addressed. Artifacts discussed
in the context of best evaluation practices were about Data / Prompts,
the Model, the System, Outputs, Process, or the Environment
(Table \ref{tab:extracted-item-artifact-discussed}).

\begin{table*}[htbp]
\caption{Extracted Item Artifact Discussed}
\begingroup
\small
\renewcommand{\arraystretch}{1.08}
\begin{tabular}{@{}
  >{\raggedright\arraybackslash}p{(\linewidth - 2\tabcolsep) * \real{0.3}}
  >{\raggedright\arraybackslash}p{(\linewidth - 2\tabcolsep) * \real{0.7}}@{}}
\toprule\noalign{}
\textbf{Extracted Item Artifact Discussed} & \textbf{Description} \\
\midrule\noalign{}
Data / Prompts & The item concerns datasets, prompts, scenarios, item
construction, sourcing, or contamination. \\
Model & The item concerns model versions, weights, training status,
fine-tuning, or access. \\
System & The item concerns tooling, pipelines, agents, or system-level
behavior. \\
Outputs & The item concerns output formats, distributions, failure
modes, or interpretation of outputs. \\
Process & The item concerns evaluation procedures, governance of the
eval, or methodological steps. \\
Environment & The item concerns deployment context, users, incentives,
threat models, or institutional setting. \\
\bottomrule\noalign{}
\end{tabular}
\label{tab:extracted-item-artifact-discussed}
\endgroup
\end{table*}

\subsubsection*{Extracted Item Objective}\label{extracted-item-objective}

In addition, we aim to label the specific goal that an extracted item
intends to advance. This dimension captures the motivation behind the
extracted item text, but not its content. Multiple objectives may be
selected. The extracted items had objectives related to Performance
Fit-for-Purpose, Safety \& Abuse Resistance, Fairness \&
Non-discrimination, Privacy \& Data Governance, Security \& Model
Integrity, Transparency \& Accountability, Sustainability \& Efficiency,
and Compliance (Table \ref{tab:extracted-item=objective}).

\begin{table*}[!ht]
\caption{Extracted Item Objective}
\begingroup
\small
\renewcommand{\arraystretch}{1.08}
\begin{tabular}{@{}
  >{\raggedright\arraybackslash}p{(\linewidth - 2\tabcolsep) * \real{0.3}}
  >{\raggedright\arraybackslash}p{(\linewidth - 2\tabcolsep) * \real{0.7}}@{}}
\toprule\noalign{}
\textbf{Extracted Item Objective} & \textbf{Description} \\
\midrule\noalign{}
Performance Fit-for-Purpose & The item concerns accuracy, usefulness,
task success, or goal alignment. \\
Safety \& Abuse Resistance & The item concerns misuse, red teaming,
refusal behavior, or safety constraints. \\
Fairness \& Non-discrimination & The item concerns demographic
performance differences or equity considerations. \\
Privacy \& Data Governance & The item concerns data handling, consent,
leakage, or retention. \\
Security \& Model Integrity & The item concerns adversarial attacks,
gaming, model theft, or tampering. \\
Transparency \& Accountability & The item concerns documentation,
explainability, auditability, or public reporting. \\
Sustainability \& Efficiency & The item concerns efficiency, resource
use, or environmental costs. \\
Compliance & The item explicitly references laws, regulations, or formal
requirements. \\
\bottomrule\noalign{}
\end{tabular}
\label{tab:extracted-item=objective}
\endgroup
\end{table*}

\subsubsection*{Extracted Item Supercategory}\label{extracted-item-supercategory}

Finally, each extracted item was labeled with a supercategory: a
high-level thematic classification used to group extracted items for
synthesis and comparison with the initial framework. Supercategories are
broader than workflow stages, and aim to identify thematic clusters that
cut across workflow stages or objectives. Supercategories could be one
of the following: Goals \& Scope Setting, Purposes and Placement, Test
Design \& Task Construction, Data \& Scenario Curation, Human Evaluation
/ Baseline, Prompting \& Elicitation Strategy, Metrics \& Rubrics,
Statistical Validity \& Uncertainty, Process Concerns, Robustness \&
Stress Testing, Fairness, Harm \& Socio-Technical Context, External
Validity \& Ecological Fit, Reproducibility, Portability \& Environment,
Reporting \& Transparency of Methods, Result Presentation \&
Interpretation, Benchmarking \& Comparisons, Tooling \& Instrumentation
for Evaluations, Usability or Cost \& Operationalization of Evaluations.
In cases where these supercategories did not fit well, extracted items
could be described as Other- Evaluation-Relevant or Not
Evaluation-Relevant (Table \ref{tab:extracted-item-supercategory}).

\begin{table*}[!ht]
\caption{Extracted Item Supercategory}
\begingroup
\small
\renewcommand{\arraystretch}{1.08}
\begin{tabular}{@{}
  >{\raggedright\arraybackslash}p{(\linewidth - 2\tabcolsep) * \real{0.3816}}
  >{\raggedright\arraybackslash}p{(\linewidth - 2\tabcolsep) * \real{0.6184}}@{}}
\toprule\noalign{}
\textbf{Extracted Item Supercategory} & \textbf{Description} \\
\midrule\noalign{}
Goals \& Scope Setting & The item concerns aims, scope, stakeholders, or
decision relevance. \\
Purposes and Placement & The item concerns where evals sit in training,
deployment, governance, or oversight. \\
Test Design \& Task Construction & The item concerns tasks, baselines,
splits, or eval structure. \\
Data \& Scenario Curation & The item concerns datasets, prompts,
scenarios, or coverage. \\
Human Evaluation / Baseline & The item concerns annotators, experts, or
human benchmarks. \\
Prompting \& Elicitation Strategy & The item concerns prompting, hints,
formats, or interaction protocols. \\
Metrics \& Rubrics, Statistical Validity \& Uncertainty & The item
concerns metrics, scoring, uncertainty, power, or validity. \\
Process Concerns & The item concerns coordination, incentives,
governance of evals. \\
Robustness \& Stress Testing & The item concerns stress tests,
adversarial behavior, or saturation. \\
Fairness, Harm \& Socio-Technical Context & The item concerns bias,
harm, stakeholders, or social context. \\
External Validity \& Ecological Fit & The item concerns real-world
relevance or deployment mismatch. \\
Reproducibility, Portability \& Environment & The item concerns code,
access, compute, or portability. \\
Reporting \& Transparency of Methods & The item concerns documenting how
evals were done. \\
Result Presentation \& Interpretation & The item concerns interpreting
or framing results. \\
Benchmarking \& Comparisons & The item concerns baselines,
state-of-the-art claims, or comparisons. \\
Tooling \& Instrumentation for Evaluations & The item concerns tools,
platforms, or instrumentation. \\
Usability, Cost \& Operationalization of Evaluations & The item concerns
cost, effort, efficiency, or maintenance. \\
Other- Evaluation-Relevant & The item is clearly eval-relevant but does
not fit above. \\
Not Evaluation-Relevant & The item does not substantively concern
evaluation. \\
\bottomrule\noalign{}
\end{tabular}
\label{tab:extracted-item-supercategory}
\endgroup
\end{table*}

\subsection{Best Fit Framework}\label{best-fit-framework-1}

The list of groupings was compiled initially, then structured and
revised by adopting the 'best fit' framework method (\cite{carroll2011bestfit}) of
starting with an existing conceptual model, identifying concepts a
priori and iterating on the initial outline of themes based on the
different frameworks presented in papers in the systematic review.

Several papers in our review presented an explicit framework for
evaluations and/or a checklist. In addition, one paper containing a
relevant framework (\cite{bean2025measuring}) was published past our database search cutoff date
and included upon recommendation by an expert.

While many papers included in the review contain frameworks, not all of
those are frameworks for performing or reporting (aspects of)
evaluations. For the synthesis of groupings performed in our analysis,
only those that contain a framework for evaluations were appropriate and
included. Therefore, we first identified papers proposing
evaluative frameworks and then restricted integration to those describing
``Evaluation Stages/Needs''. Framework elements were next mapped into a
shared structure. When a framework used very broad categories, we added
its subcategories (when available) so that the frameworks could be
compared at a similar level of detail. Outputs were presented using
structured tables and a consolidated framework box.

\subsection{Synthesis \& Groupings}\label{synthesis}

The synthesis aimed to integrate extracted recommendation items into (1) a high-level set of recommended practices and (2) a 'best fit' evaluation framework for later Delphi elicitation (\cite{carroll2011bestfit, brown1968delphi}).

The extracted items were reviewed and filtered for relevance to AI evaluation creation and use. As described above, each extracted item was independently classified by two reviewers
using the codebook dimensions. Because several codebook dimensions were non-exclusive, individual items could contribute to multiple synthesis groupings.
Synthesis groupings were defined primarily by evaluation workflow stage and item objective, and secondarily by item type and discussed artifact, to support a ``when/where/how/why'' mapping of recommendations.

Through this thematic synthesis we identified groups of recommended practices, which then informed a broader framework based on types of items. After this initial framework we adopted a 'best fit' framework approach (\cite{carroll2011bestfit}), beginning from the initial conceptual structure and iteratively revising themes based on frameworks identified in the reviewed literature. Framework elements were mapped into a shared structure, and in cases where source frameworks used broad categories, subcategories were incorporated (when available) to support comparison at a similar level of detail. We then solicited additional feedback from experts involved in AI evaluation research, which led to a final round of revisions and a final proposed framework.

\subsection{Risk of Bias within studies, in synthesis and in reporting biases}\label{risk-of-bias-within-studies-in-synthesis-and-in-reporting-biases}

The documents included in this review were predominantly frameworks, guidance documents, position papers, and regulatory or grey literature sources, and therefore standard empirical-study risk-of-bias tools were not applicable. The methodological and viewpoint differences in the literature are not corrected for, as perspectives are needed. In the discussion, we address potential biases, including selection and coverage limitations, and interpretive bias from judgment calls. 

\section{Results}
\subsection{Selected Studies and Characteristics}\label{selected-studies-and-characteristics}

The initial database searches described above resulted in the top 50 publications from Google Scholar (out of over 100,000 search results) and 612 publications from arXiv, while the supplementary effort resulted in 50 from Elicit, 27 expert suggestions and 9 additional papers from citations for a total of 748 sources (Figure~\ref{fig:prisma}).  
Title and abstract screening led to the exclusion of several papers deemed irrelevant for the purpose of this systematic review. After this first screening step, a total of 103 records were left, including 15 Google Scholar results, 42 arXiv papers, 10 results from Elicit, 27 from expert recommendations and 9 from the additional citations.
In the last step, full paper screening resulted in the removal of 48 more papers which were found to be too narrow in scope (n = 24), published before 2020 (n = 10), irrelevant in their content (n = 7), not research papers but, for example, blog posts (n = 6), not in a language we could read, interpret or quote directly without using an intermediary tool, i.e. Chinese (n = 1). After this filtering step, the final cohort contained 55 papers total, of 3 which were included both through the database searches and the expert/citation path. These 3 duplicates were removed, yielding 52 unique papers included in the analyses presented here.
The included 52 studies were published from 2020 until 2025, and the number of
relevant publications per year increased every year since 2020. Only one
study from 2020 was considered relevant to this systematic review while
17 from the year 2025 were included, and a doubling was seen from 5
studies in 2022 to 10 in 2023 (Figure~\ref{fig:study-chars}). 

\begin{figure}[!t]
\centering
\includegraphics[width=0.98\linewidth,trim={0.16in 0.00in 0.16in 0.04in},clip]{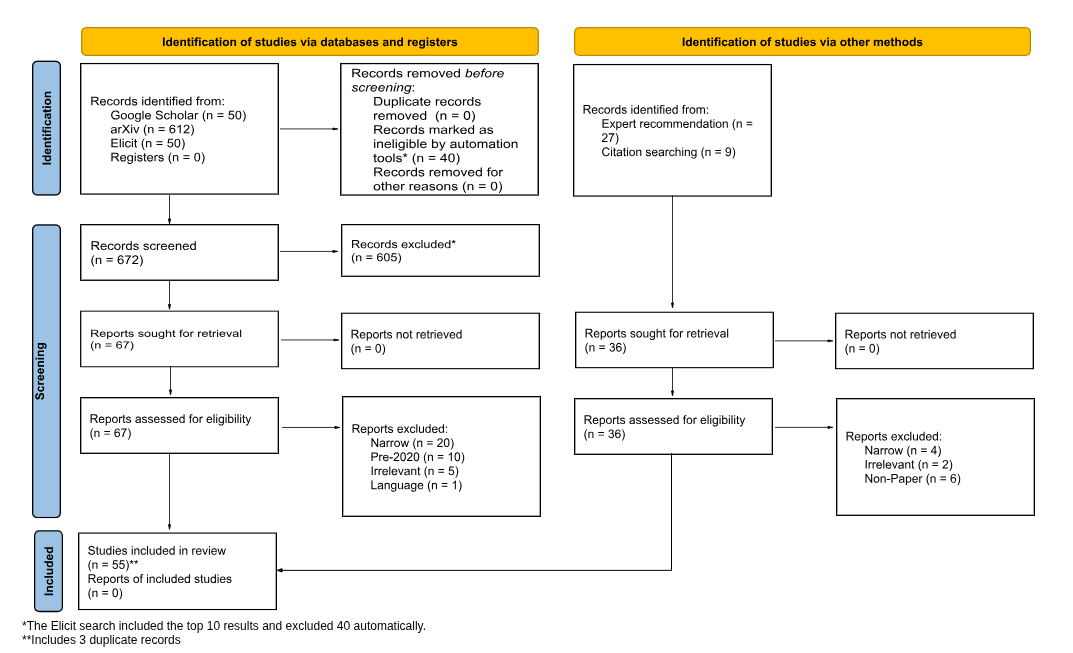}
\caption{PRISMA flowchart depicting the flow of information through phases of the systematic review, mapping out the number of records identified, included and excluded, and the reasons for exclusions.}
\label{fig:prisma}
\end{figure}

The domains studied in the papers were artificial intelligence
(50\%), machine learning (23\%), natural language
processing (17\%) or applied science (10\%), under which we
grouped papers about healthcare (n=3), material sciences (n=1) and metrology (n=1) (Figure~\ref{fig:study-chars}).
As a surrogate marker for study origin, we used the corresponding
author's email address location. While half of all studies originated in
the USA, 29\% (15/52) of studies were distributed across the Northern
hemisphere, including from the UK, Germany, Israel, Canada, Finland, Iran, Netherlands, Poland, Portugal, and one study from Argentina in the
Southern hemisphere (Figure~\ref{fig:study-chars}). Eleven studies (21\%) did
not have a designated responding author. Those studies were authored by
the EU Commission (3/11), the UK AISI (2/11), the RAND EU institute
(2/11), the UK Safety Summit (1/11) or by multiple organizations
spanning locations in the USA, UK, Canada, Australia (3/11). The number
of authors per study ranged from single authors to groups as large as
50. Most studies (31/52, 60\%) had 5 authors or fewer (Figure~\ref{fig:study-chars}). Authors were not listed for 3 reports (6\%) from organizations such as the UK Safety Summit, UK AISI, and The Frontier Model Forum.

Papers were written to address relevant concerns of one or several
target audiences. The majority of studies were intended for developers
(38/52, 73\%), auditors (13/52, 25\%), policy makers (12/52, 23\%),
providers (2/52, 4\%) and users of AI systems (1/52, 2\%) (Figure~\ref{fig:study-chars}). The reviewed publication types included recommendations (20/52, 38\%), a framework (15/52, 29\%), considerations (8/52, 15\%), position papers (7/52, 13\%) or regulation (2/52, 4\%) (Figure~\ref{fig:study-chars}).

\subsection{Extracted Item Characteristics}\label{extracted-item-characteristics}

A total of 730 items were extracted from the 52 studies included in the
review. Each item was coded per the codebook (Section~\ref{codebook}) across item class and item type (single-label), and across workflow stage, artifact discussed, and objective (non-exclusive, multi-label).

\suppressfloats[t]
Inter-rater agreement statistics are reported in Tables \ref{tab:agreement-single} and \ref{tab:agreement-multi}. A per-tag breakdown with individual $\kappa$ values is available in \textbf{OSF link removed for anonymity} and every estimate exceeded the $0.81$ threshold defining almost perfect agreement (\cite{landiskoch1977}). The overall agreement values for Cohen’s $\kappa$ were between $0.865$ and $0.895$, while the values for pooled Krippendorf’s $\alpha$ (weighted) ranged from $0.916$ to $0.954$.

\begin{table*}[t]
  \centering
  \small
  \begin{threeparttable}
    \caption{Inter-rater agreement on single-label categorizations.}
    \label{tab:agreement-single}
    \begin{tabular*}{\textwidth}{@{\extracolsep{\fill}}lrrcc@{}}
      \toprule
      Category & $n$ & \% agreement & Cohen's $\kappa$ & Krippendorff's $\alpha$ \\
      \midrule
      Relevant Class & 717 & 96.7\% & \shortstack{0.895\\{\scriptsize [0.85,\,0.94]}} & \shortstack{0.895\\{\scriptsize [0.85,\,0.94]}} \\
      Extracted Item Type & 717 & 89.0\% & \shortstack{0.870\\{\scriptsize [0.84,\,0.90]}} & \shortstack{0.870\\{\scriptsize [0.84,\,0.90]}} \\
      Suggested Supercategory & 691 & 87.4\% & \shortstack{0.865\\{\scriptsize [0.84,\,0.89]}} & \shortstack{0.865\\{\scriptsize [0.84,\,0.89]}} \\
      \bottomrule
    \end{tabular*}
    \begin{tablenotes}[flushleft]
      \footnotesize
      \item \textit{Note.} Items where either rater's cell was blank on the
      given category were dropped pairwise before estimation. 
    \end{tablenotes}
  \end{threeparttable}
\end{table*}

\begin{table*}[t]
  \centering
  \small
  \begin{threeparttable}
    \caption{Inter-rater agreement on set-valued (multi-label) categorizations. Each estimate pools $(\text{item} \times \text{tag})$ binary decisions into a single coefficient.}
    \label{tab:agreement-multi}
    \begin{tabular*}{\textwidth}{@{\extracolsep{\fill}}lrrrrcc@{}}
      \toprule
      Category & $n$ & Tags & \% exact set & Mean Jaccard & Cohen's $\kappa$ & Krippendorff's $\alpha$ \\
      \midrule
      Workflow Stage & 683 & 12 & 81.0\% & 0.912 & \shortstack{0.916\\{\scriptsize [0.90,\,0.93]}} & \shortstack{0.916\\{\scriptsize [0.90,\,0.93]}} \\
      Artifact Discussed & 683 & 6 & 90.0\% & 0.955 & \shortstack{0.952\\{\scriptsize [0.94,\,0.96]}} & \shortstack{0.952\\{\scriptsize [0.94,\,0.96]}} \\
      Objective & 683 & 8 & 90.6\% & 0.957 & \shortstack{0.964\\{\scriptsize [0.96,\,0.97]}} & \shortstack{0.964\\{\scriptsize [0.96,\,0.97]}} \\
      \bottomrule
    \end{tabular*}
    \begin{tablenotes}[flushleft]
      \footnotesize
      \item \textit{Note.} ``\% exact set'' is the share of items on which the
      two raters assigned identical tag sets; ``Mean Jaccard'' averages
      $|A \cap B| / |A \cup B|$ over items (empty/empty counted as $1$).
    \end{tablenotes}
  \end{threeparttable}
\end{table*}

In item class, most extracted items addressed evaluations (82\%), with
smaller proportions addressing audits (10\%), governance (6\%), or other
topics (2\%) (Figure~\ref{fig:item-chars}). Across item type, the most common
categories were Design guidance (22\%), Challenges (18\%), Requirements
(16\%), and Reporting guidance (15\%), followed by Categories (14\%),
while other item types were less frequent (Figure~\ref{fig:item-chars}).

\begin{figure}[!t]
  \centering
  \includegraphics[width=\linewidth,trim={0.12in 0.10in 0.10in 0.10in},clip]{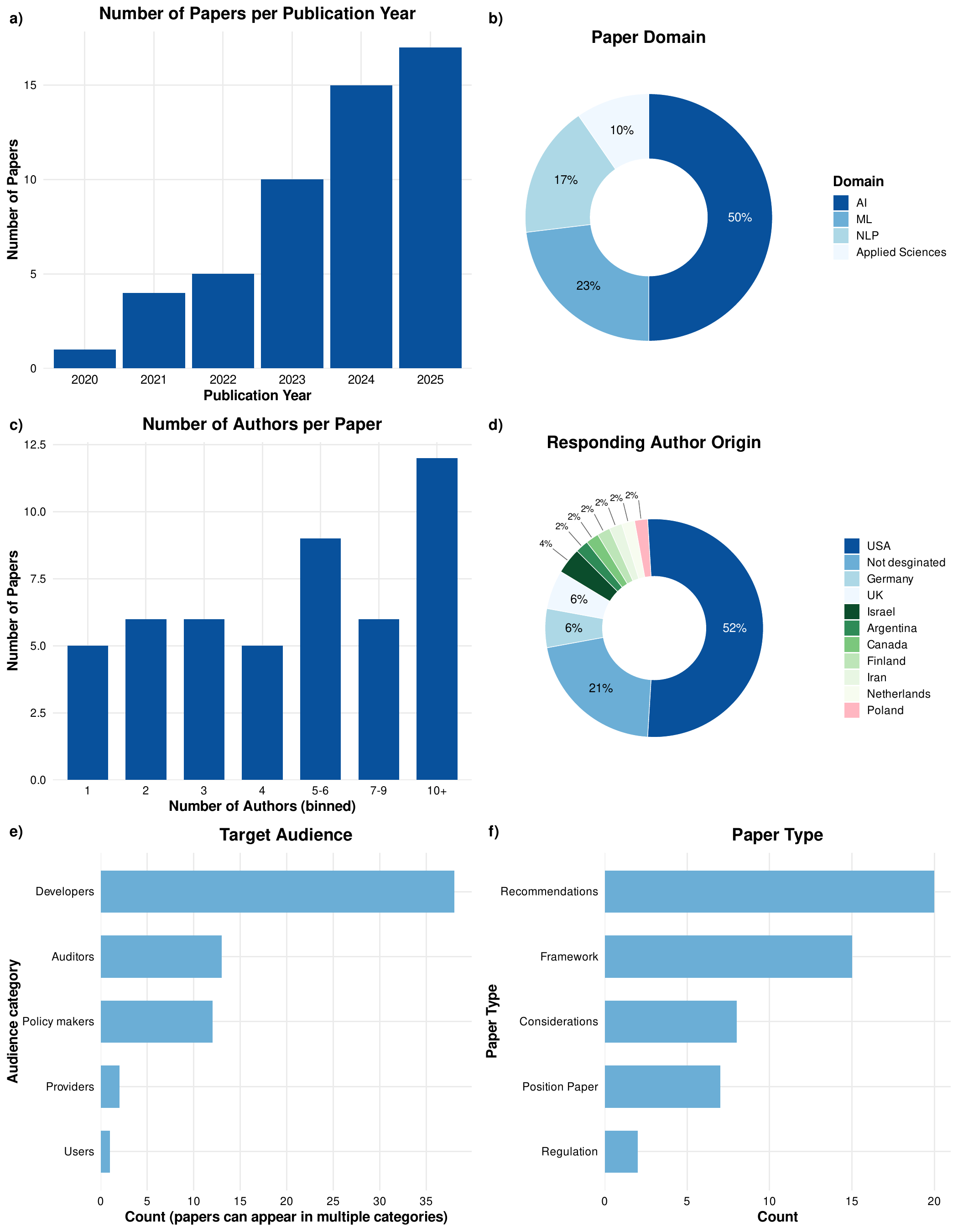}
  \caption{Summary of study characteristics by (a)~Number of papers per publications year, (b)~paper domain, (c)~number of authors per study, (d)~origin of corresponding author, (d)~audience targeted, (e)~paper type.}
\label{fig:study-chars}
\end{figure}

\begin{figure}[!t]
  \centering
  \includegraphics[width=\linewidth,height=0.84\textheight,keepaspectratio,trim={0.10in 0.10in 0.10in 0.08in},clip]{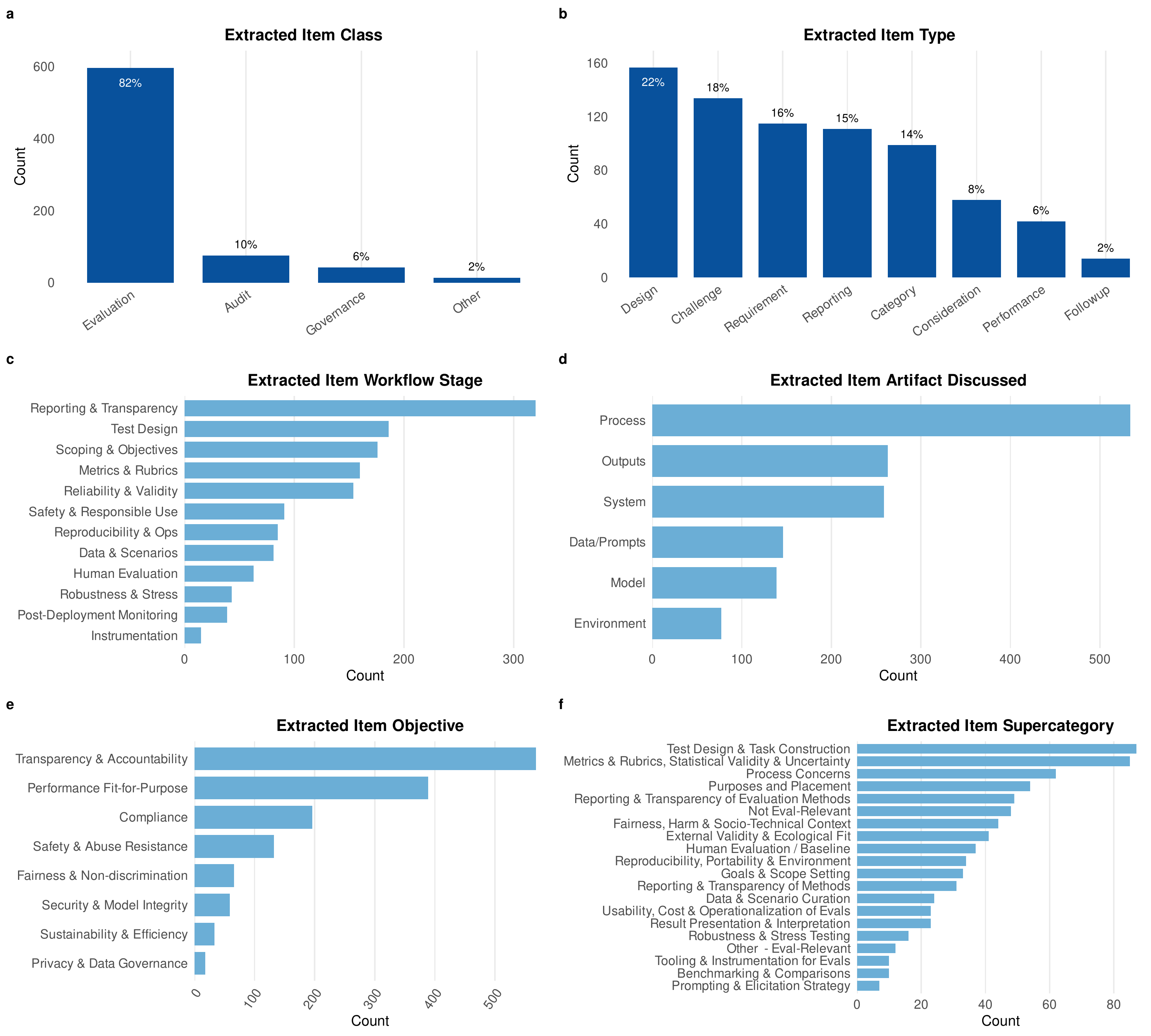}
  \caption{Summary of extracted items by (a)~their type, (b)~their class, (c)~the workflow stage, (d)~artifacts discussed, (e)~their objective, (f)~an associated supercategory.}
\label{fig:item-chars}
\end{figure}

Across the multi-label dimensions, extracted items most often addressed
multiple labels rather than a single label. 
The workflow stage described across the 730 extracted items were
assigned a total of 1413 labels, combining up to 5 labels with an
average of 1.9 and a median of 2 labels per item.
Across items, the workflow stage found relevant for discussion most
often was Reporting \& Transparency (320/730, 44\%) (Figure~\ref{fig:item-chars}).
Mentioned in more than one fifth of extracted items were the evaluation
stages of Test Design (186/730 26\%), Scoping \& Objectives (176/730,
24\%), Metrics \& Rubrics (160/730, 22\%) and Reliability \& Validity
(154/730, 21\%). In fewer than 15\% of extracted items the workflow
stages of Safety \& Responsible Use (91/730, 13\%), Reproducibility \&
Operations (85/730, 12\%), Data \& Scenarios (81/730, 11\%), Human
Evaluation (63/730, 9\%), Robustness \& Stress (43/730, 6\%),
Post-Deployment Monitoring (39/730, 5\%) were mentioned, and only 2\%
(15/730) of items was concerned with Instrumentation. Only 27\%
(197/730) of extracted items discussed a single workflow stage and the
remaining 73\% (533/730) items explored the interaction between multiple
workflow stages (Figure~\ref{fig:upset}). Here, the combination of
Reporting \& Transparency + Scoping \& Objectives was the most discussed
(12\%, 65/533), followed by Metrics \& Rubrics + Reporting \&
Transparency (11\%, 57/533).

\begin{figure}[!htbp]
  \centering
  \includegraphics[width=0.94\linewidth,trim={0.10in 0.18in 0.1in 0.12in},clip]{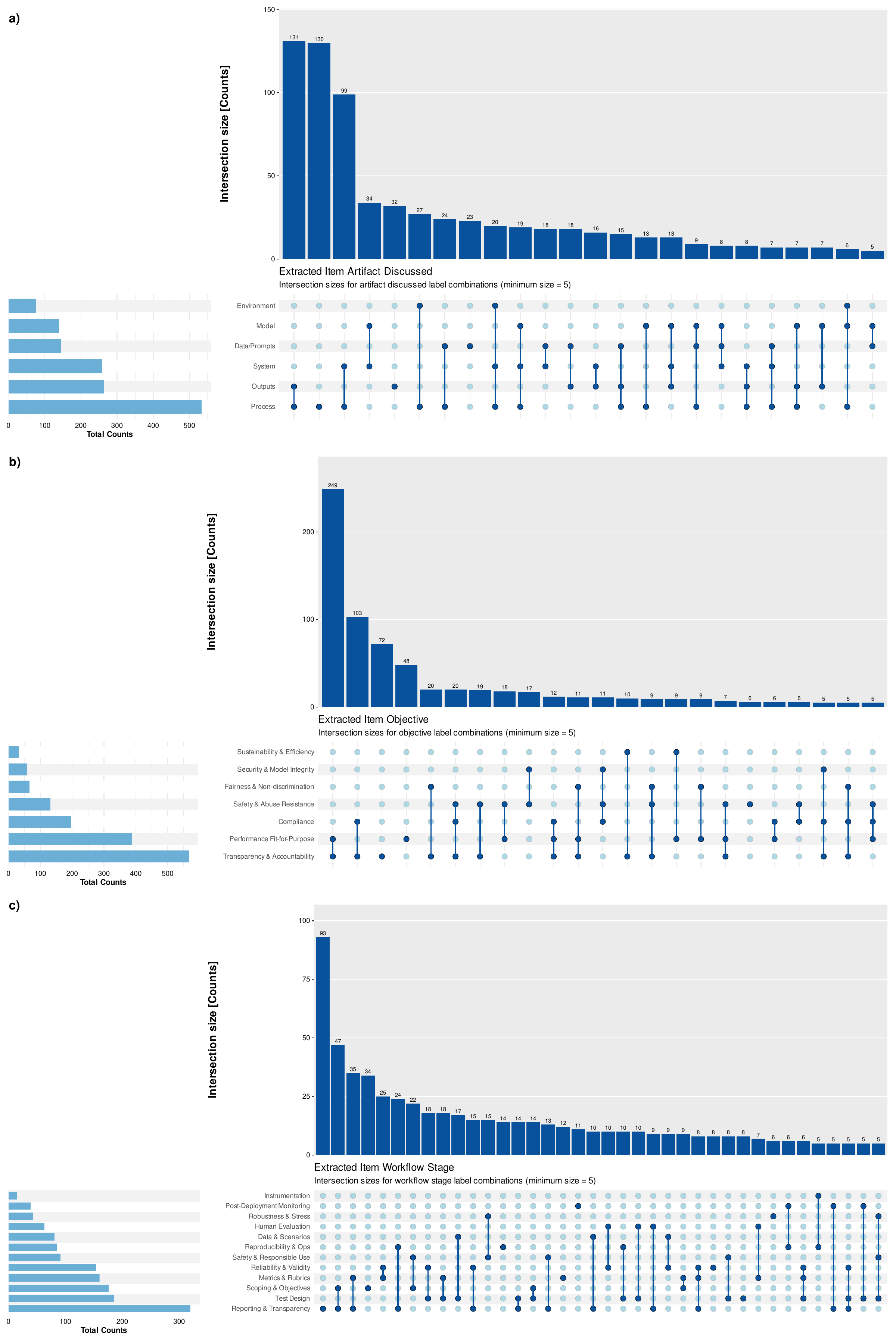}
  \caption{Frequency of combinations of the Extracted Item (a) Workflow Stages, (b) Artifacts Discussed, (c) Objectives.}
\label{fig:upset}
\end{figure}

The artifacts discussing which aspect of the AI evaluation should be
acted on were represented by a total of
1418 labels. Individual items combined up to 5 artifacts with an average
of 1.9 and a median of 2 artifacts per item. Artifacts were discussed at
the following frequencies: Process 73\% (534/730), Outputs 19\%
(263/730), System 36\% (259/730), Data / Prompts 20\% (146/730),
Model 19\% (139/730), and Environment 11\% (77/730) (Figure~\ref{fig:item-chars}). More
than a quarter of extracted items (194/730, 27\%) discussed a single
artifact, and the interconnection between multiple artifacts was
described in 73\% (536/730) of extracted items (Figure~\ref{fig:upset}).
Here, the combination of Outputs + Process was the one that occurred
most frequently (32\%, 172/536), followed by the slightly less frequent
Process + System (30\%, 161/536).

The stated objectives of extracted items were described with a
total of 1461 labels, combining up to 4 goals within one item, with an
average of 2.0 and a median of 2 goals per item. Item objectives
included most often Transparency \& Accountability (568/730, 78\%) and
Performance Fit-for-Purpose (389/730, 53\%), while Compliance
(196/730, 27\%), Safety \& Abuse Resistance (132/730, 18\%), Fairness
\& Non-discrimination (66/730, 9\%), Security \& Model Integrity
(59/730, 8\%), Sustainability \& Efficiency (33/730, 5\%) and Privacy
\& Data Governance (18/730, 3\%) were less frequent objectives (Figure~\ref{fig:item-chars}). 
A single objective was stated by 18\% (128/730) of extracted items,
while the majority of extracted items (602/730, 82\%) explored a
combination of multiple objectives (Figure~\ref{fig:upset}). The
combination of Performance Fit-for-Purpose + Transparency \&
Accountability was the most discussed (47\%,
285/602), while Compliance + Transparency \& Accountability was the
second most mentioned (26\%, 155/602).

Finally, each extracted item was summarized under the umbrella of a
supercategory showing a broad distribution of items
across themes. The supercategories applying to the most extracted items
and containing 12\% each (87/730 and 85/730, respectively) were Test
Design \& Task Construction and Metrics \& Rubrics, Statistical Validity
\& Uncertainty, while the least common were Tooling \& Instrumentation
for Evals and Benchmarking \& Comparison and Prompting \& Elicitation
Strategy at 1\% each (10/730, 10/730, 7/730, respectively) (Figure~\ref{fig:item-chars}).
In between most and least frequent lay Process Concerns
(62/730, 8\%), Purposes and Placement (54/730, 7\%), Reporting \&
Transparency of Methods (49/730, 7\%), Fairness, Harm \&
Socio-Technical Context (44/730, 6\%),
External Validity \& Ecological Fit (41/730, 6\%), Human Evaluation /
Baseline (37/730, 5\%), Reproducibility, Portability \& Environment
(34/730, 5\%), Goals \& Scope Setting (33/730, 5\%), Reporting \&
Transparency of Methods (31/730, 4\%), Data \& Scenario Curation
(24/730, 3\%), Usability, Cost \& Operationalization of Evals (23/730,
3\%), Result Presentation \& Interpretation (23/730, 3\%), Robustness \&
Stress Testing (16/730, 2\%). Additionally, 2\% (12/730) of extracted
items were labeled Other - Eval-Relevant and 7\% (48/730) were
summarized as Not Eval-Relevant.

\subsection{Best Fit Framework}\label{best-fit-framework}

In Table~\ref{tab:framework_papers} we list the subset of 13 papers from our review that present frameworks
(\cite{akula2021audit,liang2023holisticevaluationlanguagemodels,gupta2024conceptual,alizadeh2025allmetrics,bilson2025metrological,aisi2025structured,aisi2024longform,lukosiute2025llm,paskov2025preliminary,staufer2025audit,reuel2024betterbench,rismani2025measuring,bean2025measuring}),
and categorize which type of framework they describe. There was one framework for each of the following domains: Construct/Task types, Ethics Concerns, Uncertainty types, Long form task process, Impacts, Audit Stages/Needs, and seven about Evaluation Stages/Needs. Two of the frameworks sorting Evaluation Stages/Needs offered specific contributions to Capability Elicitation (\cite{aisi2025structured}) and Construct Validity
(\cite{bean2025measuring}).

\begin{table*}[!ht]
\caption{Framework Papers}
\label{tab:framework_papers}
\centering
\begingroup
\small
\setlength{\tabcolsep}{5pt}
\renewcommand{\arraystretch}{1.13}
\begin{tabular}{@{}
  >{\raggedright\arraybackslash}p{(\linewidth - 2\tabcolsep) * \real{0.5918}}
  >{\raggedright\arraybackslash}p{(\linewidth - 2\tabcolsep) * \real{0.4082}}@{}}
\toprule\noalign{}
\textbf{Paper} & \textbf{Framework Type} \\
\midrule\noalign{}
\emph{Audit and Assurance of AI Algorithms: A framework to ensure
ethical algorithmic practices in Artificial Intelligence}\par
Akula, R., et al. (\cite{akula2021audit}) & Constructs / Task types \\
\emph{Holistic Evaluation of Language Models}\par
Liang, P., et al. (\cite{liang2023holisticevaluationlanguagemodels}) & Evaluation Stages / Needs \\
\emph{A Conceptual Framework for Ethical Evaluation of Machine Learning
Systems}\par
Gupte, N.R., et al. (\cite{gupta2024conceptual}) & Ethics Concerns \\
\emph{AllMetrics: A Unified Python Library for Standardized Metric
Evaluation and Robust Data Validation in Machine Learning}\par
Alizadeh, M., et al. (\cite{alizadeh2025allmetrics}) & Evaluation Stages / Needs \\
\emph{A metrological framework for uncertainty evaluation in machine
learning classification models}\par
Bilson, S., et al. (\cite{bilson2025metrological}) & Uncertainty types \\
\emph{A structured protocol for elicitation experiments: Calibrating AI
risk assessment through rigorous elicitation practices}\par
UK AISI (\cite{aisi2025structured}) & Evaluation Stages / Needs (specific to Capability
Elicitation) \\
\emph{Long-Form Tasks: A Methodology for Evaluating Scientific
Assistants}\par
Pencharz, J., et al. (\cite{aisi2024longform}) & Long form task process \\
\emph{LLM Cyber Evaluations Don\textquotesingle t Capture Real-World
Risk}\par
Lukošiūtė, K., et al. (\cite{lukosiute2025llm}) & Impacts \\
\emph{Preliminary suggestions for rigorous GPAI model evaluations}\par
Paskov, P., et al. (\cite{paskov2025preliminary}) & Evaluation Stages / Needs \\
\emph{Audit Cards: Contextualizing AI Evaluations}\par
Staufer, L., et al. (\cite{staufer2025audit}) & Audit stages / Needs \\
\emph{BetterBench: Assessing AI Benchmarks, Uncovering Issues, and
Establishing Best Practices}\par
Reuel, A., et al. (\cite{reuel2024betterbench}) & Evaluation Stages / Needs \\
\emph{Measuring What Matters: Connecting AI Ethics Evaluations to System
Attributes, Hazards, and Harms}\par
Rismani, S., et al. (\cite{rismani2025measuring}) & Evaluation Needs \\
\emph{Measuring what Matters: Construct Validity in Large Language Model
Benchmarks}\par
Bean, A., et al. (\cite{bean2025measuring}) & Evaluation Stages / Needs (specific to
Construct Validity) \\
\bottomrule\noalign{}
\end{tabular}
\endgroup
\end{table*}

For the synthesis presented here, only those that contain a framework for ``Evaluation Stages / Needs'' are appropriate, specifically
(\cite{liang2023holisticevaluationlanguagemodels,alizadeh2025allmetrics,aisi2025structured,paskov2025preliminary,reuel2024betterbench,rismani2025measuring,bean2025measuring}).

The recurring elements that appeared across these seven frameworks were summarized and grouped into our derived initial framework detailing which of the source frameworks addressed our derived initial elements in detail, partially, or not at all. Our derived initial framework is presented in Table~\ref{tab:initial-framework} and included the following 11 candidate categories: Initial Evaluation Design Tasks, Evaluation Goals and Context, Evaluation Construct, Evaluation Development, Evaluation Context Validity, Evaluation Results and Explainability, Evaluation Metrics, Running the Evaluation, Evaluation Lifecycle, Evaluation Results Reporting, and Evaluation Results Publication.

\begin{table*}[h!t]
\captionsetup{skip=8pt}
\caption{Table of our derived initial framework and how items are addressed in other frameworks \\
($\bullet$ = clearly and explicitly, $\circ$ = partially, indirectly, or implicitly, {--} = not addressed)}
\label{tab:initial-framework}
\centering
\begingroup
\small
\setlength{\tabcolsep}{3.2pt}
\renewcommand{\arraystretch}{1.02}
\begin{tabular}{@{}
  >{\raggedright\arraybackslash}p{(\linewidth - 14\tabcolsep) * \real{0.2368}}
  >{\centering\arraybackslash}p{(\linewidth - 14\tabcolsep) * \real{0.1090}}
  >{\centering\arraybackslash}p{(\linewidth - 14\tabcolsep) * \real{0.1091}}
  >{\centering\arraybackslash}p{(\linewidth - 14\tabcolsep) * \real{0.1090}}
  >{\centering\arraybackslash}p{(\linewidth - 14\tabcolsep) * \real{0.1090}}
  >{\centering\arraybackslash}p{(\linewidth - 14\tabcolsep) * \real{0.1090}}
  >{\centering\arraybackslash}p{(\linewidth - 14\tabcolsep) * \real{0.1091}}
  >{\centering\arraybackslash}p{(\linewidth - 14\tabcolsep) * \real{0.1089}}@{}}
\toprule\noalign{}
\textbf{Derived Initial Framework} &
\rotatebox{90}{\parbox{4.15cm}{\centering
  \emph{Holistic Evaluation of Language Models}\\[2pt]
  Liang et al.~(\cite{liang2023holisticevaluationlanguagemodels})}} &
\rotatebox{90}{\parbox{4.15cm}{\centering
  \emph{AllMetrics: A Unified Python Library for Standardized Metric Evaluation}\\[2pt]
  Alizadeh et al.~(\cite{alizadeh2025allmetrics})}} &
\rotatebox{90}{\parbox{4.15cm}{\centering
  \emph{A structured protocol for elicitation experiments}\\[2pt]
  AISI UK~(\cite{aisi2025structured})}} &
\rotatebox{90}{\parbox{4.15cm}{\centering
  \emph{Preliminary suggestions for rigorous GPAI model evaluations}\\[2pt]
  Paskov et al.~(\cite{paskov2025preliminary})}} &
\rotatebox{90}{\parbox{4.15cm}{\centering
  \emph{BetterBench: Assessing AI Benchmarks, Uncovering Issues, and Establishing Best Practices}\\[2pt]
  Reuel et al.~(\cite{reuel2024betterbench})}} &
\rotatebox{90}{\parbox{4.15cm}{\centering
  \emph{Connecting AI Ethics Evaluations to System Attributes, Hazards, and Harms}\\[2pt]
  Rismani et al.~(\cite{rismani2025measuring})}} &
\rotatebox{90}{\parbox{4.15cm}{\centering
  \emph{Measuring what Matters: Construct Validity in LLM Benchmarks}\\[2pt]
  Bean et al.~(\cite{bean2025measuring})}} \\
\midrule\noalign{}
Initial Evaluation Design Tasks & $\bullet$ & $\bullet$ & $\bullet$ & $\bullet$ & $\bullet$ & $\bullet$ & $\bullet$ \\
\midrule\noalign{}
Evaluation Goals and Context & $\bullet$ & $\circ$ & $\circ$ & $\circ$ & $\circ$ & $\bullet$ & $\bullet$ \\
\midrule\noalign{}
Evaluation Construct & $\bullet$ & $\circ$ & $\circ$ & $\bullet$ & $\circ$ & $\bullet$ & $\bullet$ \\
\midrule\noalign{}
Evaluation Development & $\bullet$ & $\bullet$ & $\bullet$ & $\bullet$ & $\bullet$ & $\bullet$ & $\bullet$ \\
\midrule\noalign{}
Evaluation Context Validity & $\bullet$ & $\bullet$ & $\circ$ & $\bullet$ & $\circ$ & $\bullet$ & $\bullet$ \\
\midrule\noalign{}
Evaluation Results and Explainability & $\circ$ & $\bullet$ & $\bullet$ & $\bullet$ & $\bullet$ & $\circ$ & $\bullet$ \\
\midrule\noalign{}
Evaluation Metrics & $\bullet$ & $\bullet$ & $\circ$ & $\bullet$ & $\bullet$ & $\bullet$ & $\bullet$ \\
\midrule\noalign{}
Running the Evaluation & $\circ$ & $\bullet$ & $\bullet$ & $\bullet$ & $\bullet$ & - & $\bullet$ \\
\midrule\noalign{}
Evaluation Lifecycle & - & - & $\circ$ & $\circ$ & $\bullet$ & - & $\circ$ \\
\midrule\noalign{}
Evaluations Results Reporting & - & $\bullet$ & $\bullet$ & $\bullet$ & $\bullet$ & $\circ$ & $\circ$ \\
\midrule\noalign{}
Evaluation Results Publication & - & $\circ$ & $\bullet$ & $\bullet$ & $\circ$ & - & - \\
\bottomrule\noalign{}
\end{tabular}
\endgroup
\end{table*}

\subsection{Synthesis \& Groupings}\label{groupings}

After deriving the 11-item candidate framework, and based on feedback from experts, we reviewed the items and their thematic overlap and consolidated them into five higher-level categories: Evaluation Design, Before Evaluation Execution, Evaluation Execution, Evaluation Lifecycle, and Evaluation Reporting \&
Publication. We then refined the category structure and item labels based on expert feedback. The resulting final framework is presented in Table~\ref{tab:framework-groupings} as a high-level list of recommended practice areas, with the full itemized version (including sub-items and explanatory prompts) provided in Section~\ref{delphi-items}.

\clearpage
\begingroup
\normalsize
\setlength{\LTleft}{0pt}
\setlength{\LTright}{0pt}
\renewcommand{\arraystretch}{1.05}
\begin{longtable}{@{}
  >{\raggedright\arraybackslash}p{0.294\linewidth}
  >{\raggedright\arraybackslash}p{0.706\linewidth}@{}}
\caption{Final framework groupings and high-level items}\label{tab:framework-groupings}\\
\toprule\noalign{}
\multicolumn{2}{@{}l@{}}{\textbf{1 Design}} \\
\midrule\noalign{}
\endfirsthead
\caption[]{Final framework groupings and high-level items (continued)}\\
\toprule\noalign{}
\multicolumn{2}{@{}l@{}}{\emph{Continued from previous page}} \\
\midrule\noalign{}
\endhead
\bottomrule\noalign{}
\endfoot
Goals \& Context & • Explain stakeholders and roles

• Purpose, use context, decision linkage, audience, and relation to
existing evaluations \\[1.5pt]
Development Preregistration & • Development preregistration and
prospective design plan \\[1.5pt]
Construct \& Validity & • Construct definition and scope

• Problem framing / threat and/or consequence model

• Validity limits and design trade-offs

• Metric specification, aggregation, and interpretability

• Uncertainty, robustness, and interpretive limits

• Statistical validity and power \\[1.5pt]
Task Types \& Item Development & • Task/item construction and design
validation

• Item sourcing, provenance, representativeness, and modality
assumptions \\[1.5pt]
Human Subjects / Ethics & • Human-subjects ethics, oversight, and
safeguards \\
\midrule\noalign{}
\multicolumn{2}{@{}l@{}}{\textbf{2 Before Execution}} \\
\midrule\noalign{}
Preregistration \& Pre-run & • Prospective analysis plan and externally
verifiable pre-run readiness checks \\[1.5pt]
Scoring \& Validation & • Correctness definition and ground-truth
validity

• Rubrics and judge-system specification

• Judge training, quality control, and protocol effects \\[1.5pt]
Splits \& Holdouts & • Data partitioning, holdouts, and revision
controls \\[1.5pt]
Pilot \& Baselines & • Pilot execution and baseline calibration \\[1.5pt]
Contamination / Gaming / Awareness & • Contamination, gameability, and evaluation-awareness
controls \\[1.5pt]
Pre-reporting & • Pre-reported task summary and reporting commitments \\
\midrule\noalign{}
\multicolumn{2}{@{}l@{}}{\textbf{3 Execution}} \\
\midrule\noalign{}
Run Logging \& Repro Capture & • As-run logging and reproducibility
provenance \\[1.5pt]
Mitigations / Adaptations & • Execution-time adaptations and elicitation
controls \\[1.5pt]
Analysis \& Run Differences & • Observed failure patterns, run
differences, and execution constraints \\
\midrule\noalign{}
\multicolumn{2}{@{}l@{}}{\textbf{4 Lifecycle}} \\
\midrule\noalign{}
Data Availability \& Access & • Evaluation artifact availability and
access pathways \\[1.5pt]
Later Use \& Maintenance & • Future-use governance, versioning, and
re-evaluation criteria

• Operational documentation, portability, and maintenance support \\
\midrule\noalign{}
\multicolumn{2}{@{}l@{}}{\textbf{5 Reporting \& Publication}} \\
\midrule\noalign{}
Reporting \& Publication & • Result interpretation and evidentiary
reporting claims

• Multi-model comparison standards and trend claims

• Publication status and reporting artifacts \\[1.5pt]
Process Reporting & • Resource accounting and operational trade-off
reporting \\[1.5pt]
Usage Transparency & • Usage transparency, release context, and access
constraints \\[1.5pt]
Replication / Reproducibility & • Replication package completeness and
reproducibility sufficiency \\
\end{longtable}
\endgroup

\section{Discussion}

This paper undertook a systematic review of recommended AI evaluation practices, sourced from 52 studies, with 730 resulting extracted items. The reviewed papers were heterogeneous across many dimensions, treating ``evaluation'' as anything from a narrow protocol for performance measurement to systemic risk assessment, assumptions that strongly shape what is recommended. The most widely discussed themes were reporting and
transparency, while the most common objective was to achieve transparency and
accountability, showing agreement that communication and
reporting present areas of weakness in AI evaluation practices. Norms for executing evaluations, such as monitoring, robust operationalization or instrumentation, were of much less frequent concern, suggesting that weaknesses in AI evaluation are thought to be failures of disclosure rather than of practice.

Artifact coding showed that recommendations target the evaluation process most often, then its outputs and the deployed system. Much less attention is paid to the evaluation environment and only modest attention to the model itself, indicating that the literature increasingly views evaluation as a socio-technical procedure rather than a model-only scorecard.

Across all extracted items, only about one quarter were confined to a single workflow stage, while the large majority explicitly spanned multiple stages, most often linking Reporting and Transparency with Scoping and Objectives or Metrics and Rubrics. This suggests that the literature repeatedly treats evaluation quality as an interaction effect rather than a single intervention: As evaluations are a rapidly evolving domain, even well-specified improvements in one component (e.g., better metrics, or stronger statistical methods) may not straightforwardly transfer when upstream scoping, downstream interpretation, or operational execution remains under-specified or mismatched in the intended use. A holistic approach to AI evaluations has therefore been recommended several times.

Further, an explicit reminder that AI evaluation should rarely be seen 
as solved and that the fundamental limits of measurement should be explicitly
understood and acknowledged has recently been published by the US CAISI (\cite{nist2025deepseek}).
They disclaimed that even when conducted ``in
line with current best practices'' such findings remain ``preliminary,''
limited to ``specific domains,'' and constitute only a ``partial
assessment'' of a particular model version at a particular time, with
benchmarks that have ``methodological limitations'' and may not
generalize. Taken together, these caveats highlight that AI evaluation results 
remain context-dependent and provisional, making it essential to scrutinize whether 
benchmarks truly measure what they claim to measure (\cite{bean2025measuring}).

Indeed, reflected in the prominence of test design/task construction and metrics, rubrics/statistical validity and uncertainty among the most common supercategories observed, construct validity emerges as a key challenge. This is not least because constructs validated for humans may not apply to AI systems, and even constructs validated for one class of AI system may not apply to another. Recent interpretability research has indicated that many benchmarks which can be decomposed into testing multiple different capabilities do not, in practice, instantiate the single construct they are taken to measure (\cite{kim2025benchmark}). However, a benchmark's label is only meaningful insofar as the task and evaluation procedure do exactly that, complicating any claim that benchmarks can cleanly track a defined construct (\cite{zhou2025general}).

These challenges surrounding construct validity also interact with broader practical tensions in how evaluations should be designed and implemented. Even when the reviewed literature agrees on the importance of validity and transparency, remaining trade-offs between standardization and fit-for-purpose, as well as between feasibility and complexity impede the ability to identify common ground and limit the impact of these recommendations on practice.

Addressing this need, we identified recurring themes in the reviewed literature, taking into account existing dense guidance in some areas (e.g. design, reporting), while filling gaps in others (e.g. operationalization, access, maintenance). Through mapping the AI evaluation landscape and preserving disagreement in practices, we arrive at a consolidated
five-part framework combining recommendations regarding Design, Before
Execution, During Execution, Lifecycle, and Reporting \& Publication of
AI evaluations, designed to support consistent AI evaluation practice
and consensus building.

\paragraph{Limitations}
A limitation of this review is the database search cut-off date of
mid-2025. With the literature on AI evaluations increasing, relevant
studies may have been published after the cut-off date, which we hoped to
address by accepting expert suggested papers until the end of the year. As with any review, 
some relevant literature remains, and additional relevant work has been published since that 
time (\cite{nistai800-2-ipd, carro2025prepeval, evalevalcoalition2024}). Additionally, the
emergent field of AI agent testing has only recently begun to develop evaluation strategies and
best practices (\cite{seah2026improvingmethodologiesagenticevaluations}) and was not included in our review. We expect that most of our recommendations
apply to AI agent evaluations as well (e.g. pre-registration of the protocol, context-validity, transparency, reporting, etc.), but as agent evaluations evolve over time, 
best practices may have to adapt and evolve with them.
Additionally, the database searches were complemented by an Elicit search, expert suggestions and additional citations from included sources, which could introduce a bias towards the mainstream literature, favoring well-known or highly cited work and potentially under-represent emerging or marginal perspectives.
Further, much of the review process relied inherently on subjective
judgment and could have introduced interpretive bias. We therefore
solicited broad expert feedback on process and results, and included two
reviewers at each step of the review to minimize this potential bias.

\section{Conclusion}
This systematic review is intended to function as a stand-alone summary of AI evaluation practices, as well as an initial step in a multipart process to identify areas of consensus and common ground within the community. The findings and consolidated framework presented in this systematic review are a snapshot in time capturing which evaluation practices are considered useful in this rapidly changing field. They are already being used to solicit expert feedback as part of an in-process Delphi elicitation about consensus evaluation standards. 

This work therefore represents a novel organizational contribution towards meaningful improvements in evaluation practices and makes it clear that building better evaluations is possible.

\section{Data Availability}
Additional materials and data, such as data extracted from the included studies,
data used for analyses, analytic code and any other materials used in the review
are publicly available at \textbf{OSF link removed for anonymity}.





\section{Systematic Review References}\label{systematic-refs}

This section lists the 52 studies included in the systematic literature review.

\begingroup
\small
\setlength{\LTpre}{4pt}
\setlength{\LTpost}{4pt}
\begin{longtable}{@{} r p{3cm} p{7.0cm} l p{2cm} @{}}
\toprule
\textbf{\#} & \textbf{Author(s)} & \textbf{Title} & \textbf{Year} & \textbf{Source} \\
\midrule
\endfirsthead
\multicolumn{5}{l}{\small\textit{(continued)}} \\[2pt]
\toprule
\textbf{\#} & \textbf{Author(s)} & \textbf{Title} & \textbf{Year} & \textbf{Source} \\
\midrule
\endhead
\midrule
\multicolumn{5}{r}{\small\textit{(continued on next page)}} \\
\endfoot
\bottomrule
\endlastfoot

1  & Akula \& Garibay \cite{akula2021audit}
   & Audit and Assurance of AI Algorithms: A Framework to Ensure Ethical Algorithmic Practices in Artificial Intelligence
   & 2021 & arXiv \\[4pt]

2  & Rauh et al.\ \cite{rauh2022characteristics}
   & Characteristics of Harmful Text: Towards Rigorous Benchmarking of Language Models
   & 2022 & arXiv \\[4pt]

3  & Liang et al.\ \cite{liang2022holistic}
   & Holistic Evaluation of Language Models
   & 2022 & arXiv \\[4pt]

4  & Manheim \cite{manheim2023metrics}
   & Building Less-Flawed Metrics: Understanding and Creating Better Measurement and Incentive Systems
   & 2023 & \textit{Patterns} \\[4pt]

5  & Hafner \& Sun \cite{hafner2024privacy}
   & Empirical Privacy Evaluations of Generative and Predictive Machine Learning Models---A Review and Challenges for Practice
   & 2024 & arXiv \\[4pt]

6  & Shevlane et al.\ \cite{shevlane2023model}
   & Model Evaluation for Extreme Risks
   & 2023 & arXiv \\[4pt]

7  & Gupta et al.\ \cite{gupta2024conceptual}
   & A Conceptual Framework for Ethical Evaluation of Machine Learning Systems
   & 2024 & arXiv \\[4pt]

8  & Ding et al.\ \cite{ding2023reconsideration}
   & Reconsideration on Evaluation of Machine Learning Models in Continuous Monitoring Using Wearables
   & 2023 & arXiv \\[4pt]

9  & Kolt et al.\ \cite{kolt2024responsible}
   & Responsible Reporting for Frontier AI Development
   & 2024 & AIES \\[4pt]

10 & Bommasani \cite{bommasani2022evaluation}
   & Evaluation for Change
   & 2022 & arXiv \\[4pt]

11 & Hutchinson et al.\ \cite{hutchinson2022evaluation}
   & Evaluation Gaps in Machine Learning Practice
   & 2022 & arXiv \\[4pt]

12 & Javed et al.\ \cite{javed2022rethinking}
   & Rethinking Machine Learning Model Evaluation in Pathology
   & 2022 & arXiv \\[4pt]

13 & Collins et al.\ \cite{collins2024tripod}
   & TRIPOD+AI Statement: Updated Guidance for Reporting Clinical Prediction Models That Use Regression or Machine Learning Methods
   & 2024 & \textit{BMJ} \\[4pt]

14 & Chandrasekaran et al.\ \cite{chandrasekaran2023test}
   & Test \& Evaluation Best Practices for Machine Learning-Enabled Systems
   & 2023 & arXiv \\[4pt]

15 & Leiter et al.\ \cite{leiter2024explainable}
   & Towards Explainable Evaluation Metrics for Machine Translation
   & 2024 & \textit{JMLR} \\[4pt]

16 & Weidinger et al.\ \cite{weidinger2025evaluation}
   & Toward an Evaluation Science for Generative AI Systems
   & 2025 & arXiv \\[4pt]

17 & Costanza-Chock et al.\ \cite{costanzachock2023auditors}
   & Who Audits the Auditors? Recommendations from a Field Scan of the Algorithmic Auditing Ecosystem
   & 2023 & arXiv \\[4pt]

18 & Beddar-Wiesing et al.\ \cite{beddwarwiesing2025absolute}
   & Absolute Evaluation Measures for Machine Learning: A Survey
   & 2025 & arXiv \\[4pt]

19 & European Commission \cite{ec2025gpai_transparency}
   & The General-Purpose AI Code of Practice: Transparency Chapter
   & 2025 & Policy doc. \\[4pt]

20 & European Commission \cite{ec2025gpai_safety}
   & The General-Purpose AI Code of Practice: Safety \& Security Chapter
   & 2025 & Policy doc. \\[4pt]

21 & Alizadeh et al.\ \cite{alizadeh2025allmetrics}
   & AllMetrics: A Unified Python Library for Standardized Metric Evaluation and Robust Data Validation in Machine Learning
   & 2025 & arXiv \\[4pt]

22 & Bilson et al.\ \cite{bilson2025metrological}
   & A Metrological Framework for Uncertainty Evaluation in Machine Learning Classification Models
   & 2025 & arXiv \\[4pt]

23 & Alampara et al.\ \cite{alampara2025lessons}
   & Lessons from the Trenches on Evaluating Machine-Learning Systems in Materials Science
   & 2025 & \textit{Comp.\ Mat.\ Sci.} \\[4pt]

24 & Ullrich et al.\ \cite{ullrich2025recommendations}
   & Recommendations for Comprehensive and Independent Evaluation of Machine Learning-Based Earth System Models
   & 2025 & \textit{JGR: ML\&C} \\[4pt]

25 & South et al.\ \cite{south2024verifiable}
   & Verifiable Evaluations of Machine Learning Models Using zkSNARKs
   & 2024 & arXiv \\[4pt]

26 & Zerva \& Martins \cite{zerva2024conformalizing}
   & Conformalizing Machine Translation Evaluation
   & 2024 & \textit{TACL} \\[4pt]

27 & Orzechowski \& Moore \cite{orzechowski2022generative}
   & Generative and Reproducible Benchmarks for Comprehensive Evaluation of Machine Learning Classifiers
   & 2022 & \textit{Science Advances} \\[4pt]

28 & Tohka \& van Gils \cite{tohka2021evaluation}
   & Evaluation of Machine Learning Algorithms for Health and Wellness Applications: A Tutorial
   & 2021 & \textit{Comput.\ Biol.\ Med.} \\[4pt]

29 & Ferrer et al.\ \cite{ferrer2024good}
   & Good Practices for Evaluation of Machine Learning Systems
   & 2024 & arXiv \\[4pt]

30 & Whittlestone \& Clark \cite{whittlestone2021governments}
   & Why and How Governments Should Monitor AI Development
   & 2021 & arXiv \\[4pt]

31 & Wei et al.\ \cite{wei2025recommendations}
   & Recommendations and Reporting Checklist for Rigorous \& Transparent Human Baselines in Model Evaluations
   & 2025 & arXiv \\[4pt]

32 & UK DSIT \cite{dsit2023emerging}
   & Emerging Processes for Frontier AI Safety
   & 2023 & Policy paper \\[4pt]

33 & Paskov et al.\ \cite{paskov2025toward}
   & Toward Best Practices for AI Evaluation and Governance: A Proposal for a EU GPAI Model Evaluation Standards Task Force
   & 2025 & RAND \\[4pt]

34 & Eriksson et al.\ \cite{eriksson2025trust}
   & Can We Trust AI Benchmarks? An Interdisciplinary Review of Current Issues in AI Evaluation
   & 2025 & arXiv \\[4pt]

35 & Longpre et al.\ \cite{longpre2024safe}
   & A Safe Harbor for AI Evaluation and Red Teaming
   & 2024 & arXiv \\[4pt]

36 & UK AISI \cite{aisi2025structured}
   & A Structured Protocol for Elicitation Experiments: Calibrating AI Risk Assessment Through Rigorous Elicitation Practices
   & 2025 & AISI report \\[4pt]

37 & UK AISI \cite{aisi2024longform}
   & Long-Form Tasks: A Methodology for Evaluating Advanced AI Systems
   & 2024 & AISI report \\[4pt]

38 & Miller \cite{miller2024errorbars}
   & Adding Error Bars to Evals: A Statistical Approach to Language Model Evaluations
   & 2024 & arXiv \\[4pt]

39 & Luko\v{s}i\={u}t\.{e} \& Swanda \cite{lukosiute2025llm}
   & LLM Cyber Evaluations Don't Capture Real-World Risk
   & 2025 & arXiv \\[4pt]

40 & Paskov et al.\ \cite{paskov2025preliminary}
   & Preliminary Suggestions for Rigorous GPAI Model Evaluations
   & 2025 & RAND \\[4pt]

41 & Staufer et al.\ \cite{staufer2025audit}
   & Audit Cards: Contextualizing AI Evaluations
   & 2025 & arXiv \\[4pt]

42 & Liao et al.\ \cite{liao2021learning}
   & Are We Learning Yet? A Meta Review of Evaluation Failures Across Machine Learning
   & 2021 & NeurIPS \\[4pt]

43 & Gehrmann et al.\ \cite{gehrmann2023repairing}
   & Repairing the Cracked Foundation: A Survey of Obstacles in Evaluation Practices for Generated Text
   & 2023 & \textit{JAIR} \\[4pt]

44 & Frontier Model Forum \cite{frontier2025capability}
   & Frontier Capability Assessment
   & 2025 & FMF report \\[4pt]

45 & Reuel et al.\ \cite{reuel2024betterbench}
   & BetterBench: Assessing AI Benchmarks, Uncovering Issues, and Establishing Best Practices
   & 2024 & arXiv \\[4pt]

46 & McCaslin et al.\ \cite{mccaslin2025stream}
   & STREAM (ChemBio): A Standard for Transparently Reporting Evaluations in AI Model Reports
   & 2025 & arXiv \\[4pt]

47 & Rismani et al.\ \cite{rismani2025measuring}
   & Measuring What Matters: Connecting AI Ethics Evaluations to System Attributes, Hazards, and Harms
   & 2025 & arXiv \\[4pt]

48 & Kapoor et al.\ \cite{kapoor2024reforms}
   & REFORMS: Consensus-Based Recommendations for Machine-Learning-Based Science
   & 2024 & \textit{Science Advances} \\[4pt]

49 & Mizrahi et al.\ \cite{mizrahi2023state}
   & State of What Art? A Call for Multi-Prompt LLM Evaluation
   & 2024 & arXiv \\[4pt]

50 & Casper et al.\ \cite{casper2024black}
   & Black-Box Access Is Insufficient for Rigorous AI Audits
   & 2024 & arXiv \\[4pt]

51 & Ren et al.\ \cite{ren2024safetywashing}
   & SafetyWashing: Do AI Safety Benchmarks Actually Measure Safety Progress?
   & 2024 & arXiv \\[4pt]

52 & Wallach et al.\ \cite{wallach2024evaluating}
   & Evaluating Generative AI Systems Is a Social Science Measurement Challenge
   & 2024 & arXiv \\

\end{longtable}
\endgroup


\section{Codebook for the Classification, Grouping and Synthesis of Extracted Items}\label{codebook}

This codebook defines how extracted items from the systematic review are \textbf{classified, grouped, and synthesized}. It governs \textit{categorization decisions}, not endorsement, correctness, or priority. The output of this process is an organized representation of proposed practices suitable for later expert elicitation via a Delphi process, not a finalized standard.

\subsubsection*{\textbf{1.1 Scope}}
An \textbf{extractable item} is any statement in a paper that recommends, cautions against, constrains, or conditions how AI evaluations, audits, or closely related governance processes are designed, executed, reported, or followed up. 
Items may be:

\begin{itemize}
\item Normative (``evaluations should…'', ``it is important to…''),
\item Explicit requirements (``report X'', ``preregister Y''),
\item Diagnostic or challenges (``a common failure mode is…'', "evaluations can be challenging due to…''),
\item Conditional (``if used for deployment decisions, then…'').
\item Categorization (``Three types of stakeholders involved are…'')
\end{itemize}
Excluded are:

\begin{itemize}
\item Non-recommendation discussion
\item General explanation of known methods (eg. Top 1 vs. Top 5, how to calculate p-values or power calculations,)
\item Narrow statements relevant to only a specific class of evaluation (``Medical evaluations require regulatory…'')
\item Examples in the paper (``One evaluation which was…'').
\end{itemize}
\subsubsection*{\textbf{1.2 Epistemic stance}}
AI evaluation practices are \textbf{value-laden, context-dependent, and heterogeneous}. Categories in this codebook are heuristic and overlapping. Reviewers should prioritize \textbf{consistency and transparency} over precision - because ambiguity is expected. We need to avoid false clarity or forcing items into our preconceived assumptions, consistent with the review's stated treatment of conflicting and overlapping recommendations.

\subsection*{\textbf{Classification Dimensions}}
Each extracted item is coded along the following dimensions. For Item class and Extracted Item category, reviewers should assign \textit{one primary value per dimension}, using judgment and the decision rules below. For other classifications, multiple items can be selected when relevant

\subsubsection*{\textbf{2.1 Item Class}}
\textbf{Definition:} The broad assessment context the item primarily addresses in the paper.

{\small
\begin{longtable}{@{}p{0.15\linewidth}p{0.26\linewidth}p{0.26\linewidth}p{0.26\linewidth}@{}}
\toprule
\textbf{Value} & \textbf{Definition} & \textbf{Include When} & \textbf{Exclude / Notes} \\
\midrule
\endfirsthead
\toprule
\textbf{Value} & \textbf{Definition} & \textbf{Include When} & \textbf{Exclude / Notes} \\
\midrule
\endhead
Evaluation & Practices for measuring model capabilities, behaviors, or performance & Focus is on tasks, metrics, validity, execution, or interpretation of results & Primary point of the claim in the paper, even if organizational context or other topic is mentioned \\
\midrule
Audit & Practices for examining models or systems holistically, including process and compliance & Focus of the paper is explicitly audit, or text is about independence, access, procedures, or system-level examination & Primarily about the claimed context, not applicability of the itme \\
\midrule
Governance & Oversight, regulatory, or institutional processes shaping evaluations & Discussion of policy, regulatory use, and/or actors uninvolved in evaluation to audit. & Often overlaps; code by primary intent \\
\midrule
Other & Explicit claim or relevant recommendation, but either not cleanly in scope, or related to something distinctly different & Rare; use sparingly & Unless obvious, justify in notes \\
\midrule
\bottomrule
\end{longtable}
}

\textbf{Guiding rule:} If an item plausibly fits Evaluation \textit{or} Audit, default to \textbf{Evaluation} unless the recommendation clearly concerns organizational process rather than measurement itself.

\subsubsection*{\textbf{2.2 Extracted Item }\textbf{Type}}
\textbf{Definition:} What \textit{kind} of contribution the item makes to evaluation practice.

{\small
\begin{longtable}{@{}p{0.18\linewidth}p{0.38\linewidth}p{0.38\linewidth}@{}}
\toprule
\textbf{Value} & \textbf{Definition} & \textbf{Example(s)} \\
\midrule
\endfirsthead
\toprule
\textbf{Value} & \textbf{Definition} & \textbf{Example(s)} \\
\midrule
\endhead
Category & Defines or distinguishes kinds of evaluations, audits, or constructs & It's difficult to know what an evaluation really tells us / Audits can sometime fail due to insufficient resources \\
\midrule
Challenge & Identifies a pitfall, failure mode, or limitation & Multiple measures can complicate reporting / A sociotechnical lens is needed to ensure evaluations are relevant / Definitions of terms \\
\midrule
Consideration & High-level concern or factor to keep in mind & Consider binary evaluation versus multiple choice / internal audit or third party / Evaluating Bias or Safety or Performance \\
\midrule
Requirement & Explicitly stated requirement & You must define the metric for the evaluation / Evaluating model capabilities requires use of a model not trained for refusal. \\
\midrule
Design & Guidance about planning or structuring an evaluation & The metrics should be specified before the tasks are designed / Preregistration of audit goals / Individual metrics should be combined to give an overall view of performance \\
\midrule
Performance & Guidance about how to execute or run an evaluation & Different evaluations should be run on the same model versions / Auditors should have access to personnel to ask about training / \\
\midrule
Reporting & Guidance about documentation, transparency, or publication & The model used should be clearly indicated / Both mean performance and standard deviation should be reported / Results should be public \\
\midrule
Follow-up & Guidance about actions after evaluation (maintenance, monitoring, use in decisions) & Obsolescence criteria should be specified in the evaluation / Resources for updating the evaluation and running it on future models should be set aside in advance \\
\midrule
\bottomrule
\end{longtable}
}

\textbf{Guiding rule:} Code by \textit{what is stated in the paper, not what the substantive point being made is.}

\subsection*{\textbf{2.3 Workflow Stage / Topic}}
\textbf{Definition: }The stage(s) of the evaluation workflow, or topical area (i.e. \textit{where/when}), that the extracted item primarily concerns.

Unlike \textit{Item Class} and \textit{Extracted Item Category}, \textbf{multiple workflow stages may be selected} when an item clearly applies to more than one area. Reviewers should still identify a \textit{primary} stage in notes when feasible.

\subsubsection*{\textbf{Workflow Stages / Topics}}
{\small
\begin{longtable}{@{}p{0.15\linewidth}p{0.26\linewidth}p{0.26\linewidth}p{0.26\linewidth}@{}}
\toprule
\textbf{Value} & \textbf{Definition} & \textbf{Include When} & \textbf{Notes / Boundary Cases} \\
\midrule
\endfirsthead
\toprule
\textbf{Value} & \textbf{Definition} & \textbf{Include When} & \textbf{Notes / Boundary Cases} \\
\midrule
\endhead
Scoping \& Objectives & Defining the purpose, use case, decision context, and stakeholders for the evaluation & The item concerns goals, intended use, decision thresholds, audience, or downstream decisions & Includes stakeholder identification and normative framing of purpose \\
\midrule
Test Design & Overall structure of the evaluation and experimental setup & The item concerns task structure, baselines, comparisons, preregistration, or evaluation planning & If it specifies how to run steps, also tag Procedure \\
\midrule
Data \& Scenarios & Selection, construction, sourcing, or coverage of evaluation inputs & The item concerns datasets, prompts, scenarios, item generation, or representativeness & Includes contamination risks related to data \\
\midrule
Instrumentation & Tools, systems, and technical setup used to run the evaluation & The item concerns tooling, logging, infrastructure, APIs, or execution setup & Distinct from metrics and rubrics \\
\midrule
Metrics \& Rubrics & Definition and interpretation of scores and judgments & The item concerns metrics, scoring rules, aggregation, thresholds, or interpretation of outputs & Tag Reliability \& Validity as well if uncertainty or bias is discussed \\
\midrule
Human Evaluation & Use of humans in scoring, labeling, or baselines & The item concerns annotators, expert judgment, recruitment, incentives, or inter-rater processes & Includes human baselines \\
\midrule
Reliability \& Validity & Statistical soundness and meaningfulness of results & The item concerns uncertainty, power, bias, construct validity, internal/external validity & Often cross-cuts other stages; multi-tagging expected \\
\midrule
Robustness \& Stress & Sensitivity, adversarial testing, and failure probing & The item concerns stress tests, distribution shift, adversarial behavior, saturation, or gaming & Includes robustness to prompting or stochasticity \\
\midrule
Safety \& Responsible Use & Harm prevention and responsible application & The item concerns misuse, deployment risk, safety constraints, or ethical safeguards & Overlaps with Governance; code by focus of claim \\
\midrule
Reproducibility \& Ops & Ability to reproduce, rerun, or operationalize evaluations & The item concerns code availability, logging, versioning, costs, or operational constraints & Includes compute and resource reporting \\
\midrule
Reporting \& Transparency & Communication and disclosure of methods and results & The item concerns documentation, publication, access, or transparency obligations & Distinct from interpretation of metrics \\
\midrule
Post-Deployment Monitoring & Use of evaluations after deployment & The item concerns ongoing testing, drift detection, updating, or retirement & Includes triggers for re-evaluation \\
\midrule
\bottomrule
\end{longtable}
}

\textbf{\textbf{Coding Guidance:}}\\
\begin{itemize}
\item \textbf{Multiple selection is allowed and expected} when an item genuinely spans stages in how it is described in the text or where it clearly applies (e.g., robustness methods that affect both Test Design and Execution).
\item When in doubt, select the stage where the \textbf{main design decision or obligation arises}, and add secondary tags as needed.
\item Do \textbf{not} force exclusivity: overlap reflects real evaluation practice and is informative for later synthesis.
\end{itemize}
\textbf{Guiding rule: }Code for where the recommendation is used in or directly informs the workflow, not where it is vaguely relevant.

\subsubsection*{2.4 Artifact Discussed}
\textbf{Definition: }The primary artifact(s) that the extracted item is about (i.e. \textit{what}).

\textbf{Multiple values may be selected} when an item clearly concerns more than one artifact.

{\small
\begin{longtable}{@{}p{0.15\linewidth}p{0.26\linewidth}p{0.26\linewidth}p{0.26\linewidth}@{}}
\toprule
\textbf{Value} & \textbf{Definition} & \textbf{Include When} & \textbf{Notes / Boundary Cases} \\
\midrule
\endfirsthead
\toprule
\textbf{Value} & \textbf{Definition} & \textbf{Include When} & \textbf{Notes / Boundary Cases} \\
\midrule
\endhead
Data / Prompts & Inputs used in the evaluation & The item concerns datasets, prompts, scenarios, item construction, sourcing, or contamination & Includes synthetic and generated data \\
\midrule
Model & The model being evaluated & The item concerns model versions, weights, training status, fine-tuning, or access & Use even if model is part of a larger system \\
\midrule
System & A deployed or integrated system using a model & The item concerns tooling, pipelines, agents, or system-level behavior & Distinct from Model when behavior depends on integration \\
\midrule
Outputs & Outputs or responses produced by the model/system & The item concerns output formats, distributions, failure modes, or interpretation of outputs & Includes refusals, errors, malformed outputs \\
\midrule
Process & The evaluation or audit process itself & The item concerns evaluation procedures, governance of the eval, or methodological steps & Often co-occurs with other artifacts \\
\midrule
Environment & External context in which evaluation occurs & The item concerns deployment context, users, incentives, threat models, or institutional setting & Includes regulatory or organizational context \\
\midrule
\bottomrule
\end{longtable}
}

\textbf{\textbf{Coding guidance:}}\\
\begin{itemize}
\item Prefer \textbf{Process} when the recommendation is about \textit{how} evaluation is conducted, even if it indirectly affects other artifacts.
\item Use \textbf{Environment} when context is not merely background but materially affects interpretation or validity.
\item Multi-select is common (e.g., Data + Process for contamination checks).
\end{itemize}

\subsection*{\textbf{2.5 Objective}}
\textbf{Definition: }The primary goal(s) the extracted item is intended to advance (i.e. \textit{why}).

\textbf{Multiple objectives should be selected} when applicable.

{\small
\begin{longtable}{@{}p{0.15\linewidth}p{0.26\linewidth}p{0.26\linewidth}p{0.26\linewidth}@{}}
\toprule
\textbf{Value} & \textbf{Definition} & \textbf{Include When} & \textbf{Notes / Boundary Cases} \\
\midrule
\endfirsthead
\toprule
\textbf{Value} & \textbf{Definition} & \textbf{Include When} & \textbf{Notes / Boundary Cases} \\
\midrule
\endhead
Performance Fit-for-Purpose & Assessing whether the model/system performs its intended task adequately & The item concerns accuracy, usefulness, task success, or goal alignment & Default objective if none is explicit \\
\midrule
Safety \& Abuse Resistance & Preventing harmful, malicious, or unsafe behavior & The item concerns misuse, red teaming, refusal behavior, or safety constraints & Overlaps with Security; code by intent \\
\midrule
Fairness \& Non-discrimination & Avoiding unjust bias or disparate impact & The item concerns demographic performance differences or equity considerations & Includes representativeness concerns \\
\midrule
Privacy \& Data Governance & Protecting personal or sensitive data & The item concerns data handling, consent, leakage, or retention & Often overlaps with Compliance \\
\midrule
Security \& Model Integrity & Protecting against compromise or manipulation & The item concerns adversarial attacks, gaming, model theft, or tampering & Includes eval gaming and sandbagging \\
\midrule
Transparency \& Accountability & Enabling understanding and scrutiny & The item concerns documentation, explainability, auditability, or public reporting & Distinct from Compliance \\
\midrule
Sustainability \& Efficiency & Managing cost, compute, and environmental impact & The item concerns efficiency, resource use, or environmental costs & Often underrepresented but important \\
\midrule
Compliance & Meeting legal, regulatory, or formal standards & The item explicitly references laws, regulations, or formal requirements & Use only when compliance is explicit, not implied \\
\midrule
\bottomrule
\end{longtable}
}

\textbf{\textbf{Coding guidance:}}\\
\begin{itemize}
\item Code \textbf{what the paper says the objective is}, not what the reviewer thinks it \textit{should} be.
\item If multiple objectives are plausible but unstated, select the most direct or explicitly referenced one and note ambiguity.
\item \textbf{Compliance} should be used sparingly and only when a formal obligation is named or clearly implied.
\end{itemize}
\subsection*{2.6 Supercategory (Synthesis Theme)}
\textbf{Definition: }A high-level thematic classification used to group extracted items for synthesis and comparison with the v1 framework. Supercategories are \textbf{br}\textbf{oader than workflow stages}, and are intended to support sense-making rather than fine-grained analysis.

\textbf{One supercategory was selected} per item.

This dimension is primarily used for:

\begin{itemize}
\item High-level synthesis and reporting,
\item Identifying thematic clusters that cut across workflow stages or objectives.
\end{itemize}

\subsubsection*{\textbf{Supercategory Values}}
{\small
\begin{longtable}{@{}p{0.15\linewidth}p{0.26\linewidth}p{0.26\linewidth}p{0.26\linewidth}@{}}
\toprule
\textbf{Value} & \textbf{Definition} & \textbf{Use When:} & \textbf{Notes} \\
\midrule
\endfirsthead
\toprule
\textbf{Value} & \textbf{Definition} & \textbf{Use When:} & \textbf{Notes} \\
\midrule
\endhead
Goals \& Scope Setting & Defining what the evaluation is for and what it covers & The item concerns aims, scope, stakeholders, or decision relevance & Aligned w/ Scoping \& Objectives \\
\midrule
Purposes and Placement & How evaluations are used within broader processes & The item concerns where evals sit in training, deployment, governance, or oversight & Distinct from goals; about use \\
\midrule
Test Design \& Task Construction & Structuring the evaluation itself & The item concerns tasks, baselines, splits, or eval structure & Often paired with Data \& Scenarios \\
\midrule
Data \& Scenario Curation & Selection and construction of evaluation inputs & The item concerns datasets, prompts, scenarios, or coverage & Includes contamination concerns \\
\midrule
Human Evaluation / Baseline & Use of humans in evaluation & The item concerns annotators, experts, or human benchmarks & Includes recruitment and incentives \\
\midrule
Prompting \& Elicitation Strategy & How model responses are elicited & The item concerns prompting, hints, formats, or interaction protocols & Cross-cuts multiple workflow stages \\
\midrule
Metrics \& Rubrics, Statistical Validity \& Uncertainty & Measurement and statistical soundness & The item concerns metrics, scoring, uncertainty, power, or validity & Intentionally broad synthesis bucket \\
\midrule
Process Concerns & Meta-level issues about eval processes & The item concerns coordination, incentives, governance of evals & Do not use for routine procedures \\
\midrule
Robustness \& Stress Testing & Performance under variation or attack & The item concerns stress tests, adversarial behavior, or saturation &  \\
\midrule
Fairness, Harm \& Socio-Technical Context & Social and ethical implications & The item concerns bias, harm, stakeholders, or social context & Normative, cross-cutting \\
\midrule
External Validity \& Ecological Fit & Generalization beyond the eval & The item concerns real-world relevance or deployment mismatch &  \\
\midrule
Reproducibility, Portability \& Environment & Ability to rerun or reuse evals & The item concerns code, access, compute, or portability &  \\
\midrule
Reporting \& Transparency of Methods & Disclosure of methods & The item concerns documenting how evals were done &  \\
\midrule
Result Presentation \& Interpretation & Communication of findings & The item concerns interpreting or framing results & Distinct from methods reporting \\
\midrule
Benchmarking \& Comparisons & Comparing models or systems & The item concerns baselines, SoTA claims, or comparisons &  \\
\midrule
Tooling \& Instrumentation for Evals & Technical infrastructure & The item concerns tools, platforms, or instrumentation &  \\
\midrule
Usability, Cost \& Operationalization of Evals & Practical feasibility & The item concerns cost, effort, efficiency, or maintenance &  \\
\midrule
Other -- Eval-Relevant & Relevant but uncaptured theme & The item is clearly eval-relevant but does not fit above & Justify in notes \\
\midrule
Not Eval-Relevant & Out of scope & The item does not substantively concern evaluation & Used for cleanup \\
\midrule
\bottomrule
\end{longtable}
}

\subsubsection*{\textbf{Coding Guidance}}
\begin{itemize}
\item Supercategories are \textbf{descriptive, not prescriptive}.
\item Use them to reflect \textit{what the paper emphasizes}, not what the reviewer finds most interesting.
\item Overlap is expected; sparsity is not required.
\item If many items land in \textbf{Process Concerns} or \textbf{Other}, revisit earlier dimensions for misclassification.
\end{itemize}

\subsection*{\textbf{3. Decision Rules and Precedence}}
\begin{itemize}
\item \textbf{Stated}\textbf{ intent in the paper beats surface wording.} A reporting requirement motivated by validity concerns is still Reporting.
\item \textbf{Normative disagreement is not an error.} Items likely to provoke value disagreement should still be coded normally, not excluded.
\end{itemize}

\subsection*{\textbf{4. Worked Examples}}
This section provides concrete examples of how extracted items are coded across dimensions. Examples are drawn directly from items in the current dataset and v2 Delphi list, and are intended to illustrate typical---not idealized---classification decisions.

\subsubsection*{\textbf{Example 1: Preregistration of Evaluations}}
\textbf{\textbf{Extracted item (verbatim / near-verbatim):}}\\
``Did the developer publish a verifiable preregistration for the evaluation?''\\

\textbf{\textbf{Classification:}}
\begin{itemize}
\item \textbf{Item Class:} Evaluation
\item \textbf{Extracted Item Category:} Requirement
\textbf{Workflow Stage / Topic:}
\item Scoping \& Objectives
\item Test Design
\item \textbf{Artifact Discussed:}
\item Process
\item \textbf{Objective:}
\item Transparency \& Accountability
\item Performance Fit-for-Purpose
\item \textbf{Supercategory:}
\item Goals \& Scope Setting
\end{itemize}
\textbf{Rationale:
} The item is explicitly framed as a requirement and concerns a concrete commitment made prior to execution. Although preregistration affects reporting and interpretation downstream, the binding decision occurs during scoping and test design. The primary artifact is the evaluation process itself, and the stated motivation in the literature is improved transparency and interpretability rather than governance compliance per se.

\subsubsection*{\textbf{Example 2: Data Contamination Checks}}
\textbf{\textbf{Extracted item (verbatim / near-verbatim):}}\\
``Describe any steps taken to assess and/or prevent data contamination. Was any source material in training data? How was this checked, and will the checks retain validity for future models?''\\

\textbf{\textbf{Classification:}}
\begin{itemize}
\item \textbf{Item Class:} Evaluation
\item \textbf{Extracted Item Category:} Procedure
\item \textbf{Workflow Stage / Topic:}
\item Data \& Scenarios
\item Reliability \& Validity
\item Robustness \& Stress
\item \textbf{Artifact Discussed:}
\item Data / Prompts
\item Process
\item \textbf{Objective:}
\item Performance Fit-for-Purpose
\item Safety \& Abuse Resistance
\item Security \& Model Integrity
\item \textbf{Supercategory:}
\item Data \& Scenario Curation
\end{itemize}
\textbf{Rationale:
}This item specifies an action to be taken, not merely a consideration, and is therefore coded as a Procedure. While contamination checks are often discussed in terms of robustness or security, the primary failure mode addressed is loss of validity in reported performance. Multiple workflow stages are relevant because contamination both undermines data integrity and affects the reliability of conclusions across future model versions.

\subsection*{\textbf{5. Reviewer Workflow and Disagreement Handling}}
\begin{itemize}
\item Reviewers should flag items when the current codings seem incorrect or should be changed, or if a multi-select item should be added, not if more than one possible coding seems reasonable.
\item Disputes are resolved by lead reviewer judgment, as logged in notes.
\item Reference to this codebook,
\item Comparison with similar prior items,
\item Disagreements are logged, and noted, then resolved but not erased.
\end{itemize}

\subsection*{\textbf{6. Versioning and Change Log}}
{\small
\begin{longtable}{@{}p{0.15\linewidth}p{0.26\linewidth}p{0.26\linewidth}p{0.26\linewidth}@{}}
\toprule
\textbf{Version} & \textbf{Date} & \textbf{Change} & \textbf{Rationale} \\
\midrule
\endfirsthead
\toprule
\textbf{Version} & \textbf{Date} & \textbf{Change} & \textbf{Rationale} \\
\midrule
\endhead
0.1 & Dec 12 & Initial draft & GPT 5.2 draft, per other documentation and dataset. \\
\midrule
1.0 & Dec 22 & Initial version for use & David reviewed and corrected, regenerated some sections, and made substantial edits. \\
\midrule
\bottomrule
\end{longtable}
}


\section{Delphi Items}\label{delphi-items}

\begin{tcolorbox}[breakable, enhanced, colback=gray!5, colframe=gray!50!black,
                  title=Evaluation Reporting Checklist, fonttitle=\bfseries,
                  before skip=8pt, after skip=8pt]
{%
\setlength{\leftmargini}{2.8em}
Items listed below are individual items. Following this, there is a detailed version of the list with sub-items and explanations.
\\
\\

\textbf{Design}

Goals \& Context
\vspace{-0.55\baselineskip}

\begin{itemize}
\setlength{\itemsep}{-0.08\baselineskip}
\setlength{\parskip}{0pt}
\setlength{\parsep}{0pt}
\item
  Stakeholders and roles are explained clearly
\item
  The report (or preregistration) explains purpose, use context, decision linkage, audience, and relation to existing evaluations
\end{itemize}

Development and Design Preregistration
\vspace{-0.55\baselineskip}

\begin{itemize}
\setlength{\itemsep}{-0.08\baselineskip}
\setlength{\parskip}{0pt}
\setlength{\parsep}{0pt}
\item
  The evaluation developers preregistered the development and prospective design plans before building the evaluation
\end{itemize}

Construct \& Validity
\vspace{-0.55\baselineskip}

\begin{itemize}
\setlength{\itemsep}{-0.08\baselineskip}
\setlength{\parskip}{0pt}
\setlength{\parsep}{0pt}
\item
  The report provides a construct definition and scope
\item
  Explain the problem framing, threat, and/or consequence model
\item
  Note any limits to validity and design trade-offs
\item
  Specify and explain metrics, aggregation, and interpretation
\item
  Quantify uncertainty, robustness, and interpretive limits
\item
  Perform and report statistical analyses
\end{itemize}

Task Types \& Item Development
\vspace{-0.55\baselineskip}

\begin{itemize}
\setlength{\itemsep}{-0.08\baselineskip}
\setlength{\parskip}{0pt}
\setlength{\parsep}{0pt}
\item
  Explain task and item construction, and perform design validation and report
\item
  Note item sourcing, provenance, representativeness, and modality assumptions
\end{itemize}

Human Subjects / Ethics
\vspace{-0.55\baselineskip}

\begin{itemize}
\setlength{\itemsep}{-0.08\baselineskip}
\setlength{\parskip}{0pt}
\setlength{\parsep}{0pt}
\item
  Report human-subjects ethics, oversight, and safeguards where relevant
\end{itemize}

\textbf{Before Execution}

Protocol \& Pre-run
\vspace{-0.55\baselineskip}

\begin{itemize}
\setlength{\itemsep}{-0.08\baselineskip}
\setlength{\parskip}{0pt}
\setlength{\parsep}{0pt}
\item
  Perform prospective analysis and report plan and externally verifiable pre-run readiness checks after evaluation development, before usage
\end{itemize}

Scoring \& Validation
\vspace{-0.55\baselineskip}

\begin{itemize}
\setlength{\itemsep}{-0.08\baselineskip}
\setlength{\parskip}{0pt}
\setlength{\parsep}{0pt}
\item
  Provide a correctness definition and explain ground-truth validity
\item
  Create and provide rubrics and judge-system specification
\item
  The evaluation performs and explains judge training and quality control processes, then reports them
\end{itemize}

Splits \& Holdouts
\vspace{-0.55\baselineskip}

\begin{itemize}
\setlength{\itemsep}{-0.08\baselineskip}
\setlength{\parskip}{0pt}
\setlength{\parsep}{0pt}
\item
  Explain data partitioning, holdouts, and revision controls
\end{itemize}

Pilot \& Baselines
\vspace{-0.55\baselineskip}

\begin{itemize}
\setlength{\itemsep}{-0.08\baselineskip}
\setlength{\parskip}{0pt}
\setlength{\parsep}{0pt}
\item
  Pilot the evaluation and calibrate baselines
\end{itemize}

Contamination / Gaming / Awareness
\vspace{-0.55\baselineskip}

\begin{itemize}
\setlength{\itemsep}{-0.08\baselineskip}
\setlength{\parskip}{0pt}
\setlength{\parsep}{0pt}
\item
  Perform contamination, gameability, and evaluation-awareness controls
\end{itemize}

Pre-reporting
\vspace{-0.55\baselineskip}

\begin{itemize}
\setlength{\itemsep}{-0.08\baselineskip}
\setlength{\parskip}{0pt}
\setlength{\parsep}{0pt}
\item
  Provide a task summary and reporting commitments before evaluation usage
\end{itemize}

\textbf{Execution}

Run Logging \& Repro Capture
\vspace{-0.55\baselineskip}

\begin{itemize}
\setlength{\itemsep}{-0.08\baselineskip}
\setlength{\parskip}{0pt}
\setlength{\parsep}{0pt}
\item
  As-run logging and reproducibility provenance
\end{itemize}

Mitigations / Adaptations
\vspace{-0.55\baselineskip}

\begin{itemize}
\setlength{\itemsep}{-0.08\baselineskip}
\setlength{\parskip}{0pt}
\setlength{\parsep}{0pt}
\item
  Document and report execution-time adaptation of the evaluation and elicitation controls
\end{itemize}

Analysis \& Run Differences
\vspace{-0.55\baselineskip}

\begin{itemize}
\setlength{\itemsep}{-0.08\baselineskip}
\setlength{\parskip}{0pt}
\setlength{\parsep}{0pt}
\item
  Document failure patterns, run differences, and execution constraints
\end{itemize}

\textbf{Lifecycle}

Data Availability \& Access
\vspace{-0.55\baselineskip}

\begin{itemize}
\setlength{\itemsep}{-0.08\baselineskip}
\setlength{\parskip}{0pt}
\setlength{\parsep}{0pt}
\item
  Provide evaluation outputs or document artifact availability and access pathways
\end{itemize}

Later Use \& Maintenance
\vspace{-0.55\baselineskip}

\begin{itemize}
\setlength{\itemsep}{-0.08\baselineskip}
\setlength{\parskip}{0pt}
\setlength{\parsep}{0pt}
\item
  Document governance, versioning, and re-evaluation criteria (if any exist) for future use
\item
  Provide operational documentation, portability, and maintenance support
\end{itemize}

\textbf{Reporting \& Publication}

Reporting \& Publication
\vspace{-0.55\baselineskip}

\begin{itemize}
\setlength{\itemsep}{-0.08\baselineskip}
\setlength{\parskip}{0pt}
\setlength{\parsep}{0pt}
\item
  Provide results interpretation and evidence for claims
\item
  Report multi-model comparison standards and trend claims, if relevant
\item
  Explain publication status and reporting artifacts
\end{itemize}

Process Reporting
\vspace{-0.55\baselineskip}

\begin{itemize}
\setlength{\itemsep}{-0.08\baselineskip}
\setlength{\parskip}{0pt}
\setlength{\parsep}{0pt}
\item
  Include resource accounting and operational trade-offs
\end{itemize}

Transparency
\vspace{-0.55\baselineskip}

\begin{itemize}
\setlength{\itemsep}{-0.08\baselineskip}
\setlength{\parskip}{0pt}
\setlength{\parsep}{0pt}
\item
  Mention evaluation usage rules or conditions, release context, and access constraints
\end{itemize}

Replication / Reproducibility
\vspace{-0.55\baselineskip}

\begin{itemize}
\setlength{\itemsep}{-0.08\baselineskip}
\setlength{\parskip}{0pt}
\setlength{\parsep}{0pt}
\item
  Provide a replication package and reproducibility guide, or explain reproducibility constraints
\end{itemize}

}

\subsubsection*{Items, sub-items, and details}

\subsubsection*{Design}

\paragraph{Goals \& Context}

\begin{itemize}
\item
  \textbf{Stakeholders and roles are explained clearly}

\begin{itemize}
  \item \textit{State who built and who ran the evaluation} Was the evaluation built by a company deploying a model, a group using the model, an auditor, or a research group focused on some aspect of the model design or outcomes?
  \item \textit{State any other identified affected stakeholders for the outputs} Who is directly or indirectly impacted by this evaluation?
  \item \textit{Explain stakeholder/expert consultation that informed design, if any} Who was consulted and/or what expertise informed the choices in the evaluation?
  \item \textit{Explain any normative assumptions about how/why the eval is used} Are there any ethical or safety assumptions which inform the choice of items, metrics, or decision rules? Are they made clear?
  \item \textit{Include a positionality statement for designers/users} What social and cultural environment could have affected the choices made? What political or ideological commitments exist which others may find important to understand the choices made?
\end{itemize}
\item
  \textbf{The report (or preregistration) explains purpose, use context, decision linkage, audience, and relation to existing evaluations}

\begin{itemize}
  \item \textit{Explain the purpose and, if relevant, success objective of the evaluation} What core objective does this evaluation serve (e.g., capability assessment, safety assurance, oversight, deployment readiness), and what would count as success for that objective?
  \item \textit{Explain intended use context and lifecycle placement} Where and when will this evaluation be used (e.g., pre-deployment, post-deployment, internal gating, external reporting), and in what operational context?
  \item \textit{Explicitly mention any decision linkage and actions planned} Which concrete decisions are informed by results, and what outcomes would trigger specific actions (e.g., ship/no-ship, mitigation, redesign, monitoring escalation)?
  \item \textit{Explain audience and interpretive boundaries} Who is the intended audience (developers, auditors, regulators, users, public), what decisions should they use results for, and what uses should be avoided?
  \item \textit{Report the relationship to existing evaluations and benchmark choice rationale} How does this evaluation relate to prior benchmarks/literature, why is this approach preferred here, and what known limitations or path-dependence risks should be recognized?
\end{itemize}
\end{itemize}

\paragraph{Development and Design Preregistration}

\begin{itemize}
\item
  \textbf{The evaluation developers preregistered the development and prospective design plans before building the evaluation}

\begin{itemize}
  \item \textit{The evaluation includes a public/verifiable design preregistration preceding the evaluation development} Was there a publicly verifiable design preregistration (or timestamped equivalent) made before the evaluation was constructed or before development was complete?
  \item \textit{Provide a prospective plan for what the evaluation will measure, the task types and methods, the access level (black-box, weight access, or internal activations) and the evaluation creation or selection process} Does the preregistered plan state scope, intended claims, and key methodological commitments including task types/methods and criteria for selecting or excluding tasks in advance, rather than retrofitting explanations after results?
\end{itemize}
\end{itemize}

\paragraph{Construct \& Validity}

\begin{itemize}
\item
  \textbf{The report provides a construct definition and scope}

\begin{itemize}
  \item \textit{Explain measurement of the construct/capability and the target claim} What exactly is being measured by the evaluation, and what claim about model/system behavior or impacts is this intended to support?
  \item \textit{Explain sub-areas/aspects of the construct, and accompanying reporting plan} How were sub-components identified, and how will reporting disaggregate different sub-components or capabilities, where relevant?
  \item \textit{Provide an explanation or map from construct to real-world use/applications} What validity claims connect construct performance to real-world use (for example external/ecological validity, transportability across contexts, and consequential validity), and what boundaries or uncertainties are noted?
\end{itemize}
\item
  \textbf{Explain the problem framing, threat, and/or consequence model}

\begin{itemize}
  \item \textit{Specify capabilities, failure modes, or threat model} Which failure modes or threat scenarios are in scope, for which use cases, actors, or contexts, and with what assumptions about system boundaries and use conditions?
  \item \textit{Discuss different performance levels and outcomes} How is the measured performance linked to the real-world performance and consequences (including severity, affected groups, and decision relevance), and where are those links uncertain?
\end{itemize}
\item
  \textbf{Note any limits to validity and design trade-offs}

\begin{itemize}
  \item \textit{Explain external validity gaps and expected limits} What use contexts, populations, or deployment settings are not well represented, and what limits does that create for generalizing results?
  \item \textit{Explain representativeness and coverage, construct contamination, and item-format impacts} How representative is the item set of the intended task space, and to what extent might non-target factors (format effects, shortcuts, confounders) influence scores?
  \item \textit{Note any internal/external validity trade-offs} What trade-offs were made across internal, external, and construct validity (including feasibility/cost constraints), and how might those choices affect interpretation?
\end{itemize}
\item
  \textbf{Specify and explain metrics, aggregation, and interpretation}

\begin{itemize}
  \item \textit{Explain the metric(s) used by the evaluation} What metric(s) are used, why they are appropriate for the construct and decision context, and what alternatives were considered?
  \item \textit{Explain the score meaning, and the range, including floors or ceilings, and explain any baseline(s) or natural comparison} How should scores be interpreted (range, floor/ceiling, reference baselines, practical significance), and what interpretive limits are acknowledged?
  \item \textit{Explain aggregate and subgroup metrics, and asymmetry handling if relevant} How are aggregate and disaggregated results related? Include treatment of subgroup heterogeneity, error asymmetries, and any weighting choices, or lack of weighting.
\end{itemize}
\item
  \textbf{Quantify uncertainty, robustness, and interpretive limits}

\begin{itemize}
  \item \textit{Explain any sources of uncertainty/variation and trade-offs} Which uncertainty sources are considered (sampling, measurement, elicitation, grader/model variability), which are excluded, and what trade-offs motivated those choices?
  \item \textit{Perform and report input variation and robustness (prompting/stochasticity)} How sensitive are outcomes to prompting/scaffolding choices, stochasticity, language(s) used for questions, and implementation differences, how were these tests performed, and how is the robustness or uncertainty characterized?
  \item \textit{Design to allow uncertainty quantification} Which uncertainty estimators are used (e.g., confidence intervals/standard errors, paired comparisons, clustered or repeated-run summaries), and what assumptions do they rely on?
  \item \textit{Explain saturation, ceiling effects, and interpretability of above-human performance} Specify the point at which the measure should be considered saturated, and if scores approach ceilings or exceed human reference levels, explain how comparisons and practical conclusions should be interpreted.
\end{itemize}
\item
  \textbf{Perform and report statistical analyses}

\begin{itemize}
  \item \textit{Perform power analysis or other sample size justification, and report detectable effect sizes or equivalent} How were sample size, number of runs, and minimum detectable effects determined for the intended analyses and decisions? Was the alpha used justified?
  \item \textit{Address multiplicity control, preregistration, and researcher degrees of freedom} How are multiple comparisons, optional stopping, and flexible analytic choices handled, and which hypotheses/analysis plans were prespecified? How will this affect interpretation?
  \item \textit{Explain distributional assumptions, dependence structure, and model misspecification risks} What assumptions underlie inference from evaluation results (e.g., independence, clustering, distributional form, and assumptions needed for causal rather than conflated attribution of effects), how were these checked, and how would violations change interpretation?
\end{itemize}
\end{itemize}

\paragraph{Task Types \& Item Development}

\begin{itemize}
\item
  \textbf{Explain task and item construction, and perform design validation and report}

\begin{itemize}
  \item \textit{Explain the task/item development workflow, and the selection logic} How were tasks/items generated or selected, what inclusion/exclusion rules were applied, and how does this workflow align with the evaluation objective?
  \item \textit{Explain any relationship to prior evaluations/literature, and provide a rationale for any adaptation} How does this task/item design compare with prior benchmarks or methods, and what rationale supports adopting, adapting, or departing from prior work?
  \item \textit{Perform validation of task/item criteria, and mention any domain-expert consultation} Which domain experts informed task or rubric design, how was their input or other validation approaches incorporated, and what checks were used to validate topicality, difficulty, and criteria appropriateness?
\end{itemize}
\item
  \textbf{Note item sourcing, provenance, representativeness, and modality assumptions}

\begin{itemize}
  \item \textit{Explain legal and ethical sourcing and permissions} What legal rights, licenses, and ethical permissions apply to source materials and represented people/groups, and how were these verified?
  \item \textit{Explain item provenance, representativeness, and coverage of items} What are the sources and provenance of items, how representative is coverage of the intended task space, and what known coverage gaps or trade-offs remain?
  \item \textit{Explain formats, languages, and modalities (including multimodality) and validity and robustness implications} How do item format choices, chosen language(s), and modality (single- vs multi-modal) affect what capability is measured, and how results should be interpreted?
\end{itemize}
\end{itemize}

\paragraph{Human Subjects / Ethics}

\begin{itemize}
\item
  \textbf{Report human-subjects ethics, oversight, and safeguards where relevant}

\begin{itemize}
  \item \textit{Report any ethics/IRB approvals and oversight for human-involved components} Where people are involved in development, scoring, or baselines, what ethics approvals or equivalent oversight were obtained, if any, and what scope/limitations do they cover?
  \item \textit{Report participant safeguards and protocol transparency (when humans are involved)} What participant protections (e.g., consent, privacy, risk mitigation, compensation expectations) were documented, and which protocol details are documented and provided to participants?
\end{itemize}
\end{itemize}

\subsubsection*{Before Execution}

\paragraph{Protocol \& Pre-run}

\begin{itemize}
\item
  \textbf{Perform prospective analysis and report plan and externally verifiable pre-run readiness checks after evaluation development, before usage}

\begin{itemize}
  \item \textit{Before running the evaluation, but after development, publicly register or commit a protocol with hypotheses, methods, and decision criteria, and the relationship to any design preregistration which occurred} What analysis plans, hypotheses, thresholds, and decision rules were specified in advance, and what deviations are allowed or documented?
  \item \textit{Perform pre-run verification of items, code, and metrics} Before running the evaluation, what checks confirmed that items, scoring code, and metrics are ready and aligned with the stated plan?
  \item \textit{Register and report expected performance vs. baseline (pre-run)} Were prior expectations documented (including baseline/reference expectations), and how are these expectations intended to constrain interpretation after results are observed?
\end{itemize}
\end{itemize}

\paragraph{Scoring \& Validation}

\begin{itemize}
\item
  \textbf{Provide a correctness definition and explain ground-truth validity}

\begin{itemize}
  \item \textit{Provide correctness definition and ground-truth specification} How is correctness defined for the task, how is this checked, and how does it relate to any ground truth which exists?
  \item \textit{Report validation approach and treatment of malformed/refusal outputs} How was ground truth validated (including ambiguity, label quality, and inter-rater agreement where applicable) and how are refusals, malformed outputs, partial answers, or parser failures classified?
  \item \textit{Check and report human reference performance, and explain comparability (if relevant)} If human performance is used as a reference, which humans, under which conditions, and how was comparability ensured between human and model conditions, and what limitations remain?
\end{itemize}
\item
  \textbf{Create and provide rubrics and judge-system specification}

\begin{itemize}
  \item \textit{Create and provide rubric/scoring system, including construction and validation details} How was the rubric/scoring schema constructed, what evidence supports validity, and how were edge cases or multiple acceptable solutions handled?
  \item \textit{Report LLM judge(s) model/prompt details and validation} If LLM judges are used, what model/prompt/configuration is specified, and how was judge reliability or bias tested against independent checks?
  \item \textit{Report human judge(s) recruitment/demographics/compensation/consent details} If human judges are used, how were they recruited and characterized, what compensation/consent protocols applied, how was judge reliability and inter-rater reliability assessed, and how were representativeness limits documented?
\end{itemize}
\item
  \textbf{The evaluation performs and explains judge training and quality control processes, then reports them}

\begin{itemize}
  \item \textit{Explain judge training and quality-assurance procedures} What training/calibration processes were used for judges or graders, and what ongoing QA checks were planned to maintain scoring consistency?
  \item \textit{Perform and explain consistency checks and effort/method-effect controls} How were method effects (instructions, interface, prompt format, effort/time differences) identified or controlled to support fair comparisons?
  \item \textit{Explain blinding and independence safeguards (if applicable)} What blinding or independence safeguards were used (e.g., graders blinded to condition or identity), and how was blinding quality assessed?
\end{itemize}
\end{itemize}

\paragraph{Splits \& Holdouts}

\begin{itemize}
\item
  \textbf{Explain data partitioning, holdouts, and revision controls}

\begin{itemize}
  \item \textit{Detail pilot/tuning/evaluation/holdout separation and rationale} How were dataset partitions defined (including disjointness and intended role of each split), and why is this partitioning appropriate for the evaluation objective?
  \item \textit{Explain any splits and construction across relevant dimensions and/or different sub-constructs for reporting} How were splits constructed across key dimensions (e.g., task type, difficulty, domain, source) to manage uneven partitioning and/or to avoid distorted conclusions?
  \item \textit{Revision policy after split creation and pilot feedback (if relevant)} What rules govern post-split revisions to items/scoring, and how are leakage or overfitting risks managed when revisions based on pilot or tuning items occurs?
  \item \textit{Explain holdout governance for later and ongoing use} How are holdout items protected, refreshed, and accessed over time to preserve future validity?
\end{itemize}
\end{itemize}

\paragraph{Pilot \& Baselines}

\begin{itemize}
\item
  \textbf{Pilot the evaluation and calibrate baselines}

\begin{itemize}
  \item \textit{Pilot runs on pre-selected systems/configurations} Which systems/configurations were piloted, and what pilot goals were defined (feasibility, failure discovery, instrumentation/scoring checks)?
  \item \textit{Explain human/random/prior-model baselines for comparability} What baseline choices were made, and how was comparability ensured between baseline and target conditions?
  \item \textit{Pilot diagnostics for failure modes and scorer reliability} What pilot diagnostics were used to identify systematic failure modes, scoring artifacts, or grader reliability issues before main execution? How are these diagnostics or checks implemented for later use?
  \item \textit{Pilot saturation/ceiling checks} Were any near-ceiling performance patterns identified during pilot, how will they be assessed in the evaluation, and how did this inform whether the evaluation outcomes are meaningful?
\end{itemize}
\end{itemize}

\paragraph{Contamination / Gaming / Awareness}

\begin{itemize}
\item
  \textbf{Perform contamination, gameability, and evaluation-awareness controls}

\begin{itemize}
  \item \textit{Perform data contamination checks and exposure assessment} What checks estimate pre-exposure to benchmark content (or close variants), and how is contamination risk incorporated into interpretation?
  \item \textit{Gameability/sandbagging prevention and detection mechanisms} What mechanisms are used to detect or deter strategic behavior (including sandbagging or exploiting scoring shortcuts), and how robust are these mechanisms?
  \item \textit{Perform evaluation-awareness checks (if relevant)} How is model awareness of being evaluated assessed (if relevant), and how are awareness effects separated from underlying capability?
  \item \textit{Report access/secrecy controls for sensitive evaluation components} What access controls or secrecy measures apply to sensitive prompts/items/scorers, and what trade-offs exist between contamination resistance and reproducibility?
\end{itemize}
\end{itemize}

\paragraph{Pre-reporting}

\begin{itemize}
\item
  \textbf{Provide a task summary and reporting commitments before evaluation usage}

\begin{itemize}
  \item \textit{Provide a summary of task inputs/types/features} What pre-run summary was provided for task inputs/types/features so downstream audiences can understand the defined scope as understood before outcomes were produced?
  \item \textit{Create and preregister a reporting plan and comparison schema} Which outputs, subgroup slices, baseline comparisons, and decision-relevant summaries are committed in advance for reporting?
  \item \textit{Explain uncertainty and disaggregation disclosure plans in the advance reporting} Which uncertainty and disaggregation outputs will be disclosed? (This is a reporting commitment, possibly distinct from how they are statistically estimated in earlier sections.)
\end{itemize}
\end{itemize}

\subsubsection*{Execution}

\paragraph{Run Logging \& Repro Capture}

\begin{itemize}
\item
  \textbf{As-run logging and reproducibility provenance}

\begin{itemize}
  \item \textit{Capture as-run parameters/variables and configuration state} What parameters/configuration values were actually used at runtime (including seeds, decoding/scaffolding settings, tool configurations), and how were they recorded?
  \item \textit{Preserve full run logs, model/tool versions, and execution trace} Were complete run logs preserved (including model/version identifiers, tool use, failures/retries, and run-level metadata) in a form sufficient for third-party reconstruction?
  \item \textit{Record protocol deviations from pre-run plans} Which deviations from preregistered or pre-run plans occurred during execution, and how were they timestamped and justified as part of run provenance?
\end{itemize}
\end{itemize}

\paragraph{Mitigations / Adaptations}

\begin{itemize}
\item
  \textbf{Document and report execution-time adaptation of the evaluation and elicitation controls}

\begin{itemize}
  \item \textit{Document any execution-time changes to the evaluation and timing relative to planned protocol (if relevant)} What model or system mitigations/changes were applied during execution, if any, when were they introduced, and how might they affect comparability to planned conditions?
  \item \textit{Document the elicitation strategy actually used (e.g. hints, few-shot, multi-turn, scaffolding)} What prompting/elicitation strategy was actually used in runs, and how does this differ from planned settings (if at all)?
  \item \textit{Provide adaptation rationale and reproducibility implications (if relevant)} For execution-time adaptations, what rationale was documented, and what implications do these changes have for reproducibility and interpretation?
\end{itemize}
\end{itemize}

\paragraph{Analysis \& Run Differences}

\begin{itemize}
\item
  \textbf{Document failure patterns, run differences, and execution constraints}

\begin{itemize}
  \item \textit{Perform failure-pattern analysis (biases/confounders/scoring artifacts)} What post-run analyses examined common failures, confounders, and scoring artifacts, and how are these distinguished from prior design assumptions?
  \item \textit{Explain observed differences across runs/analyses and any impacts} What differences were observed across runs, environments, or analysis pipelines, and how materially might these differences affect conclusions?
  \item \textit{Explain how execution constraints (budget/data/time/access) may affect results and interpretation} Which constraints were actually encountered during execution, and how should they change interpretation relative to planned conditions?
  \item \textit{Note evaluated-model vs deployed-system differences} What differences exist between the evaluated setup and real deployed system, and how do these differences limit the conclusions or change how they should be interpreted?
\end{itemize}
\end{itemize}

\subsubsection*{Lifecycle}

\paragraph{Data Availability \& Access}

\begin{itemize}
\item
  \textbf{Provide evaluation outputs or document artifact availability and access pathways}

\begin{itemize}
  \item \textit{Licensing terms, open/modifiable availability, and model access requirements (black-box, weight access, or internal activations)} What licensing or access terms apply to datasets, prompts, code, and evaluation artifacts, and to what extent are these artifacts open/modifiable?
  \item \textit{Provide results and replication data publicly, or explain any access alternatives if non-public} Where possible, provide result files. If full public release is not possible, what access alternatives are provided (e.g., local/offline execution, representative samples, controlled access) and what limits remain?
  \item \textit{Provide reusable code or access conditions for third-party evaluations and future reuse} If code is available, what conditions govern third-party or future access? (This may include governance, approvals, and security/privacy constraints) If not, what are the access conditions for reuse or for replication and verification?
\end{itemize}
\end{itemize}

\paragraph{Later Use \& Maintenance}

\begin{itemize}
\item
  \textbf{Document governance, versioning, and re-evaluation criteria (if any exist) for future use}

\begin{itemize}
  \item \textit{Provide plans for updates, deprecation, and retirement, if relevant} What lifecycle plan exists for updating, deprecating, or retiring the evaluation as models, risks, and contexts change?
  \item \textit{Provide version/build status and change-tracking policy} How are versions/releases tracked, and what change-log policy supports comparability across versions?
  \item \textit{Explain any criteria for valid future use on new models/systems} What criteria determine whether reuse on new models/systems remains valid versus requiring redesign or recalibration?
  \item \textit{Monitoring triggers and re-evaluation protocol} What predefined triggers (performance shifts, saturation, contamination, model/system changes) require reconsideration of the evaluation? For auditors and developers, what process exists for when triggers are met?
\end{itemize}
\item
  \textbf{Provide operational documentation, portability, and maintenance support}

\begin{itemize}
  \item \textit{Include operational documents/APIs/requirements along with publication, and document any portability support} What implementation documentation (requirements, API/specs, setup/run guidance) supports reproducible use across environments and organizations?
  \item \textit{Provide guidelines for reuse, adaptation, and safe integration} What guidance is provided for reusing/adapting the evaluation (including acceptable and non-acceptable modifications) without undermining validity?
  \item \textit{The evaluation identifies maintenance ownership and feedback/contact channels, if any exist} Who owns ongoing maintenance, how can users report issues or request clarifications, and what service/update expectations are documented?
\end{itemize}
\end{itemize}

\subsubsection*{Reporting \& Publication}

\paragraph{Reporting \& Publication}

\begin{itemize}
\item
  \textbf{Provide results interpretation and evidence for claims}

\begin{itemize}
  \item \textit{Note any baselines and provide guidance for interpretation of results} How are results contextualized against relevant baselines, and what explicit guidance is given on valid and invalid interpretations?
  \item \textit{Perform and report sensitivity analyses, note differences vs prior reports, if any} What sensitivity analyses are reported, and if they exist, how are differences from prior reports explained?
  \item \textit{The report ties conclusions to results, relates them to decision criteria, and notes any limitations for comparisons} How are conclusions tied to reported evidence and pre-specified decision criteria, and what limitations or non-comparable conditions are disclosed for cross-model or cross-report comparisons?
  \item \textit{Note any disagreements or disputes over interpretation} Are notable disagreements about interpretation documented (including unresolved disputes), and what evidence is cited for competing views?
\end{itemize}
\item
  \textbf{Report multi-model comparison standards and trend claims, if relevant}

\begin{itemize}
  \item \textit{The report standardizes comparisons across reports, models/versions, and different sources of variation or uncertainty} What standardization choices (datasets, prompts, scoring, temperature, access conditions) support fair comparisons, and how is uncertainty incorporated in comparative claims?
  \item \textit{Include trend analysis and future-performance projections, if possible} If trends or forecasts are reported or hypothesized, what are the relevant assumptions, uncertainty bounds, or falsification conditions?
  \item \textit{Note model selection and setup uncertainty in comparative conclusions} How could uncertainty in model choice, elicitation setup, or evaluation scope affect scores, rankings, or comparative conclusions?
\end{itemize}
\item
  \textbf{Explain publication status and reporting artifacts}

\begin{itemize}
  \item \textit{The report notes submission/publication status} What is the publication/dissemination status of the evaluation report, and where can the (current) authoritative version be found? If not available, when and where will it be published?
  \item \textit{Complete an evaluation card/checklist and machine-readable and human-readable reporting template} Which structured reporting artifacts (eval card/checklist/template) are provided? Are all reporting fields populated, or if not, explained?
  \item \textit{Include content warnings and communication safeguards (if relevant)} What content warnings or communication safeguards are provided for potentially harmful or sensitive evaluation content?
  \item \textit{Provide a post-publication correction/update pathway} Is there a mechanism for errata, updates, or clarifications if issues are identified after release?
\end{itemize}
\end{itemize}

\paragraph{Process Reporting}

\begin{itemize}
\item
  \textbf{Include resource accounting and operational trade-offs}

\begin{itemize}
  \item \textit{Report compute/sample/resource requirements for running the evaluation} What resource envelope is required to run the evaluation (compute, sample counts, access needs, tooling assumptions), and how is this specified for prospective users?
  \item \textit{Include cost accounting methodology and reported cost metrics (tokens, dollars, time)} How are costs measured and reported (including what is included/excluded), and what comparability caveats apply across runs or organizations?
  \item \textit{The report describes operational trade-offs across run modes and deployment contexts} How do different operational modes (e.g., stronger elicitation, more repeats, safety constraints, limited access) trade off cost, speed, and evidentiary strength?
  \item \textit{Explain any difference between planned resource profile and realized execution or constraints encountered} How is the reported planned resource profile distinguished from the actual execution or constraints actually encountered (which are tracked elsewhere), and how are discrepancies explained?
\end{itemize}
\end{itemize}

\paragraph{Transparency}

\begin{itemize}
\item
  \textbf{Mention evaluation usage rules or conditions, release context, and access constraints}

\begin{itemize}
  \item \textit{Report any evaluation of Pre-release, pre-deployment, or checkpoint model versions and impacts on reporting validity} Was the evaluation run on any previous model versions and not reported? Were reported results generated on model versions different from the released model? What caveats are provided for using these results in external claims (e.g., model cards)?
  \item \textit{Provide a timeline of runs/results release and recipient audiences} Explain the timeline from execution to release, which audiences received results at each stage, and mention any implications this sequencing has for interpretation or governance.
  \item \textit{Access/publication restrictions and 'file-drawer' risks, where evaluations are not reported} What legal/procedural or strategic constraints limited publication/access, and how are risks of selective reporting or unreported evaluations addressed?
  \item \textit{The report notes any GUID/canary usage, or other verifiable cryptographic commitment mechanisms (e.g. hashes of non-public preregistration)} What mechanisms (e.g., canaries, cryptographic commitments, controlled disclosure proofs) support transparency, either to allow reuse, or to ensure validity when full artifact release is constrained?
  \item \textit{Include a disclosure policy or explanation for sensitive information hazards} How are transparency obligations balanced with security, privacy, and misuse risks when deciding what evaluation details to disclose?
\end{itemize}
\end{itemize}

\paragraph{Replication / Reproducibility}

\begin{itemize}
\item
  \textbf{Provide a replication package and reproducibility guide, or explain reproducibility constraints}

\begin{itemize}
  \item \textit{Make replication code and execution package available, or otherwise explain how reproducibility can be achieved} What executable artifacts are available (code, environment specs, scripts, seeds, dependencies), and under what access conditions?
  \item \textit{Ensure users can reproduce reported outputs or understand and audit irreproducible components} To what extent can published outputs be reproduced end-to-end, and if full reproduction is not possible, what auditable records or independent verification pathways are provided?
  \item \textit{Note data retention plans and auditability of reproducibility artifacts} If artifacts are not fully public, explain how they are securely retained, versioned, and made auditable over time.
\end{itemize}
\end{itemize}

\end{tcolorbox}

\section{PRISMA 2020 Checklist}\label{prisma}
\begingroup

\renewenvironment{landscape}{}{}
\footnotesize
\renewcommand*{\arraystretch}{1.3}
\setlength{\LTpre}{0pt}
\setlength{\LTpost}{0pt}
\begingroup
\captionsetup{labelformat=empty}
\begin{longtable}{|p{0.17\linewidth}|p{0.04\linewidth}|p{0.55\linewidth}|p{0.20\linewidth}|}

\hline
\rowcolor{lightgray}
\textbf{Section and Topic} & \textbf{Item} & \textbf{Checklist Item} & \textbf{Location where item is reported} \\
\hline
\endfirsthead

\hline
\rowcolor{lightgray}
\multicolumn{4}{|l|}{\textit{Table continued from previous page}} \\
\hline
\rowcolor{lightgray}
\textbf{Section and Topic} & \textbf{Item} & \textbf{Checklist Item} & \textbf{Location where item is reported} \\
\hline
\endhead

\hline
\multicolumn{4}{|r|}{\textit{Continued on next page}} \\
\hline
\endfoot

\hline
\endlastfoot

\multicolumn{4}{|l|}{\textbf{TITLE}} \\
\hline
Title & 1 & Identify the report as a systematic review. & Title \\
\hline
\multicolumn{4}{|l|}{\textbf{ABSTRACT}} \\
\hline
Abstract & 2 & See the PRISMA 2020 for Abstracts Checklist. & Abstract \\
\hline
\multicolumn{4}{|l|}{\textbf{INTRODUCTION}} \\
\hline
Rationale & 3 & Describe the rationale for the review in the context of existing knowledge. & Introduction; Background and Previous Work \\
\hline
Objectives & 4 & Provide an explicit statement of the objective(s) or question(s) the review addresses. & Introduction \\
\hline
\multicolumn{4}{|l|}{\textbf{METHODS}} \\
\hline
Eligibility criteria & 5 & Specify the inclusion and exclusion criteria for the review and how studies were grouped for the syntheses. & Methods -- Search, Screening, and Inclusion (including Search Strategy and Sources, Title/Abstract Screening, and Full Paper Screening); Methods -- Synthesis \& Groupings \\
\hline
Information sources & 6 & Specify all databases, registers, websites, organisations, reference lists and other sources searched or consulted to identify studies. Specify the date when each source was last searched or consulted. & Methods -- Search, Screening, and Inclusion; Search Strategy and Sources \\
\hline
Search strategy & 7 & Present the full search strategies for all databases, registers and websites, including any filters and limits used. & Methods -- Search, Screening, and Inclusion; Search Strategy and Sources \\
\hline
Selection process & 8 & Specify the methods used to decide whether a study met the inclusion criteria of the review, including how many reviewers screened each record and each report retrieved, whether they worked independently, and if applicable, details of automation tools used in the process. & Methods -- Search, Screening, and Inclusion; Title/Abstract Screening; Full Paper Screening \\
\hline
Data collection process & 9 & Specify the methods used to collect data from reports, including how many reviewers collected data from each report, whether they worked independently, any processes for obtaining or confirming data from study investigators, and if applicable, details of automation tools used in the process. & Methods -- Item Extraction and Characterization; Extracted Item-Level Characterization \\
\hline
Data items & 10a & List and define all outcomes for which data were sought. Specify whether all results that were compatible with each outcome domain in each study were sought (e.g.\ for all measures, time points, analyses), and if not, the methods used to decide which results to collect. & Methods -- Item Extraction and Characterization; Extracted Item-Level Characterization; Extracted Item Class; Extracted Item Type; Extracted Item Workflow Stage; Extracted Item Artifact Discussed; Extracted Item Objective; Extracted Item Supercategory; Section~\ref{codebook} \\
\cline{2-4}
 & 10b & List and define all other variables for which data were sought (e.g.\ participant and intervention characteristics, funding sources). Describe any assumptions made about any missing or unclear information. & Methods -- Study-Level Characterization; Methods -- Item Extraction and Characterization; Extracted Item-Level Characterization; Section~\ref{codebook} \\
\hline
Study risk of bias assessment & 11 & Specify the methods used to assess risk of bias in the included studies, including details of the tool(s) used, how many reviewers assessed each study and whether they worked independently, and if applicable, details of automation tools used in the process. & Methods -- Risk of Bias within studies, in synthesis and in reporting biases \\
\hline
Effect measures & 12 & Specify for each outcome the effect measure(s) (e.g.\ risk ratio, mean difference) used in the synthesis or presentation of results. & Not applicable (no effect measures or meta-analysis reported) \\
\hline
Synthesis methods & 13a & Describe the processes used to decide which studies were eligible for each synthesis (e.g.\ tabulating the study intervention characteristics and comparing against the planned groups for each synthesis (item~\#5)). & Methods -- Best Fit Framework; Methods -- Synthesis \& Groupings; Results -- Best Fit Framework; Results -- Synthesis \& Groupings; Tables~\ref{tab:framework_papers}--\ref{tab:framework-groupings} \\
\cline{2-4}
 & 13b & Describe any methods required to prepare the data for presentation or synthesis, such as handling of missing summary statistics, or data conversions. & Methods -- Item Extraction and Characterization; Extracted Item-Level Characterization; Section~\ref{codebook} (no summary-statistic conversions reported) \\
\cline{2-4}
 & 13c & Describe any methods used to tabulate or visually display results of individual studies and syntheses. & Results -- Selected Studies and Characteristics (Figures~\ref{fig:prisma}--\ref{fig:study-chars}); Results -- Extracted Item Characteristics (Figure~\ref{fig:item-chars}); Results -- Best Fit Framework (Tables~\ref{tab:framework_papers} and \ref{tab:initial-framework}); Results -- Synthesis \& Groupings (Table~\ref{tab:framework-groupings}) \\
\cline{2-4}
 & 13d & Describe any methods used to synthesize results and provide a rationale for the choice(s). If meta-analysis was performed, describe the model(s), method(s) to identify the presence and extent of statistical heterogeneity, and software package(s) used. & Methods -- Best Fit Framework; Methods -- Synthesis \& Groupings \\
\cline{2-4}
 & 13e & Describe any methods used to explore possible causes of heterogeneity among study results (e.g.\ subgroup analysis, meta-regression). & Not applicable (no heterogeneity analyses reported) \\
\cline{2-4}
 & 13f & Describe any sensitivity analyses conducted to assess robustness of the synthesized results. & Not reported / not conducted \\
\hline
Reporting bias assessment & 14 & Describe any methods used to assess risk of bias due to missing results in a synthesis (arising from reporting biases). & Methods -- Risk of Bias within studies, in synthesis and in reporting biases; Discussion -- Limitations \\
\hline
Certainty assessment & 15 & Describe any methods used to assess certainty (or confidence) in the body of evidence for an outcome. & Not reported / no formal certainty assessment \\
\hline
\multicolumn{4}{|l|}{\textbf{RESULTS}} \\
\hline
Study selection & 16a & Describe the results of the search and selection process, from the number of records identified in the search to the number of studies included in the review, ideally using a flow diagram. & Results -- Selected Studies and Characteristics; Figure~\ref{fig:prisma} \\
\cline{2-4}
 & 16b & Cite studies that might appear to meet the inclusion criteria, but which were excluded, and explain why they were excluded. & Aggregate exclusion reasons are reported in Results -- Selected Studies and Characteristics; Figure~\ref{fig:prisma}; individual excluded reports are not separately listed in the manuscript \\
\hline
Study characteristics & 17 & Cite each included study and present its characteristics. & Results -- Selected Studies and Characteristics; Figure~\ref{fig:study-chars}; Section~\ref{systematic-refs} \\
\hline
Risk of bias in studies & 18 & Present assessments of risk of bias for each included study. & Not applicable / not reported; Methods -- Risk of Bias within studies, in synthesis and in reporting biases \\
\hline
Results of individual studies & 19 & For all outcomes, present, for each study: (a)~summary statistics for each group (where appropriate) and (b)~an effect estimate and its precision (e.g.\ confidence/credible interval), ideally using structured tables or plots. & Not applicable for this review design; study-level characteristics are summarized in Results -- Selected Studies and Characteristics and included studies are listed in Section~\ref{systematic-refs} \\
\hline
Results of syntheses & 20a & For each synthesis, briefly summarise the characteristics and risk of bias among contributing studies. & Results -- Extracted Item Characteristics; Results -- Best Fit Framework; Results -- Synthesis \& Groupings; Discussion -- Limitations \\
\cline{2-4}
 & 20b & Present results of all statistical syntheses conducted. If meta-analysis was done, present for each the summary estimate and its precision (e.g.\ confidence/credible interval) and measures of statistical heterogeneity. If comparing groups, describe the direction of the effect. & Results -- Extracted Item Characteristics; Tables~\ref{tab:agreement-single} and \ref{tab:agreement-multi}; no meta-analysis was conducted \\
\cline{2-4}
 & 20c & Present results of all investigations of possible causes of heterogeneity among study results. & Not applicable (no heterogeneity investigations reported) \\
\cline{2-4}
 & 20d & Present results of all sensitivity analyses conducted to assess the robustness of the synthesized results. & Not applicable / not reported \\
\hline
Reporting biases & 21 & Present assessments of risk of bias due to missing results (arising from reporting biases) for each synthesis assessed. & Not reported as a separate results item; Methods -- Risk of Bias within studies, in synthesis and in reporting biases; Discussion -- Limitations \\
\hline
Certainty of evidence & 22 & Present assessments of certainty (or confidence) in the body of evidence for each outcome assessed. & Not reported / no formal certainty assessment \\
\hline
\multicolumn{4}{|l|}{\textbf{DISCUSSION}} \\
\hline
Discussion & 23a & Provide a general interpretation of the results in the context of other evidence. & Discussion \\
\cline{2-4}
 & 23b & Discuss any limitations of the evidence included in the review. & Discussion -- Limitations \\
\cline{2-4}
 & 23c & Discuss any limitations of the review processes used. & Methods -- Risk of Bias within studies, in synthesis and in reporting biases; Discussion -- Limitations \\
\cline{2-4}
 & 23d & Discuss implications of the results for practice, policy, and future research. & Discussion; Conclusion \\
\hline
\multicolumn{4}{|l|}{\textbf{OTHER INFORMATION}} \\
\hline
Registration and protocol & 24a & Provide registration information for the review, including register name and registration number, or state that the review was not registered. & Methods -- Search, Screening, and Inclusion (publicly available preregistered protocol on OSF; no registry number reported) \\
\cline{2-4}
 & 24b & Indicate where the review protocol can be accessed, or state that a protocol was not prepared. & Methods -- Search, Screening, and Inclusion \\
\cline{2-4}
 & 24c & Describe and explain any amendments to information provided at registration or in the protocol. & Not reported / no amendments described in the manuscript \\
\hline
Support & 25 & Describe sources of financial or non-financial support for the review, and the role of the funders or sponsors in the review. & Acknowledgments and Disclosure of Funding \\
\hline
Competing interests & 26 & Declare any competing interests of review authors. & Acknowledgments and Disclosure of Funding \\
\hline
Availability of data, code and other materials & 27 & Report which of the following are publicly available and where they can be found: template data collection forms; data extracted from included studies; data used for all analyses; analytic code; any other materials used in the review. & Data Availability; Methods -- Search, Screening, and Inclusion \\
\hline
\end{longtable}
\endgroup

\endgroup

\endgroup

\clearpage
\section{\EvalCards{} UI}\label{app:ui}
\subsection{Model view: GPT-5}
\begin{figure}[htbp]
    \centering
    \includegraphics[width=.95\linewidth]{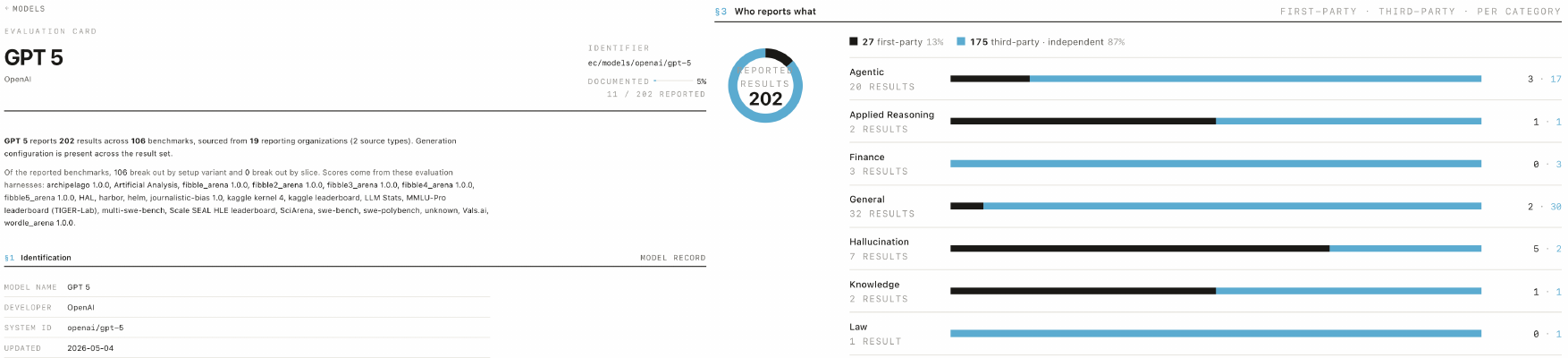}
    \caption{Identification (left) and Who reports what (right): The \textbf{Identification} section provides general model data, and the \textbf{Who reports what} section provides a breakdown of the distribution between first and third-party evaluation results per benchmark topic. The numbers to the right provide the counts for first and third-party, respectively.}
    \label{fig:gpt5-header}
\end{figure}

\begin{figure}[htbp]
    \centering
    \includegraphics[width=\linewidth]{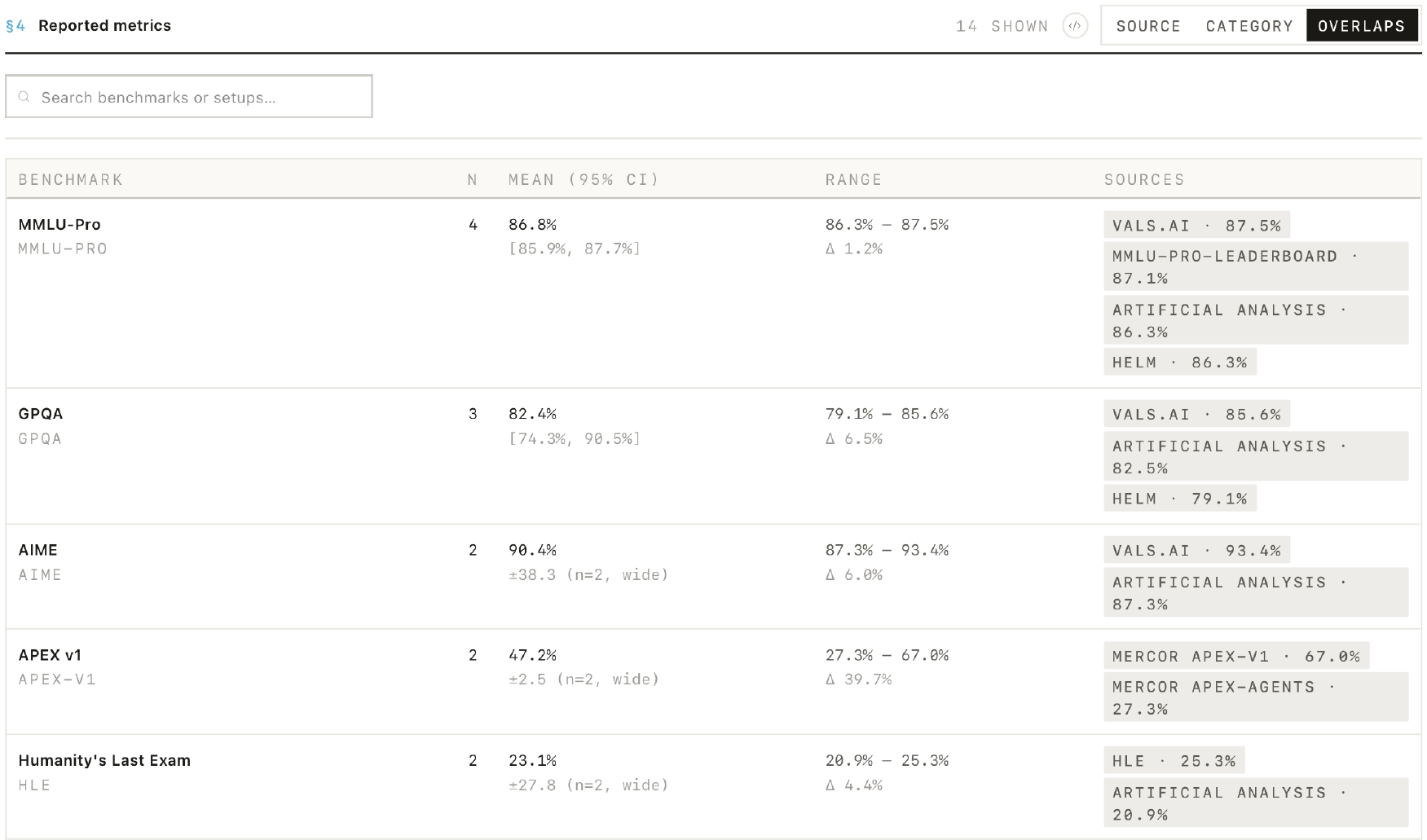}
    \caption{Reported metrics, Overlaps: Cross-source score comparison for GPT-5 benchmarks reported by multiple organizations, showing per-source scores, score range, and divergence.of each benchmark alongside the number of reported results (N), the mean score with 95\% confidence interval, and individual per-source scores, helping readers directly compare the results reported by different third-party evaluators.}
    \label{fig:overlap-gpt5}
\end{figure}

\begin{figure}[htbp]
    \centering
    \includegraphics[width=.95\linewidth]{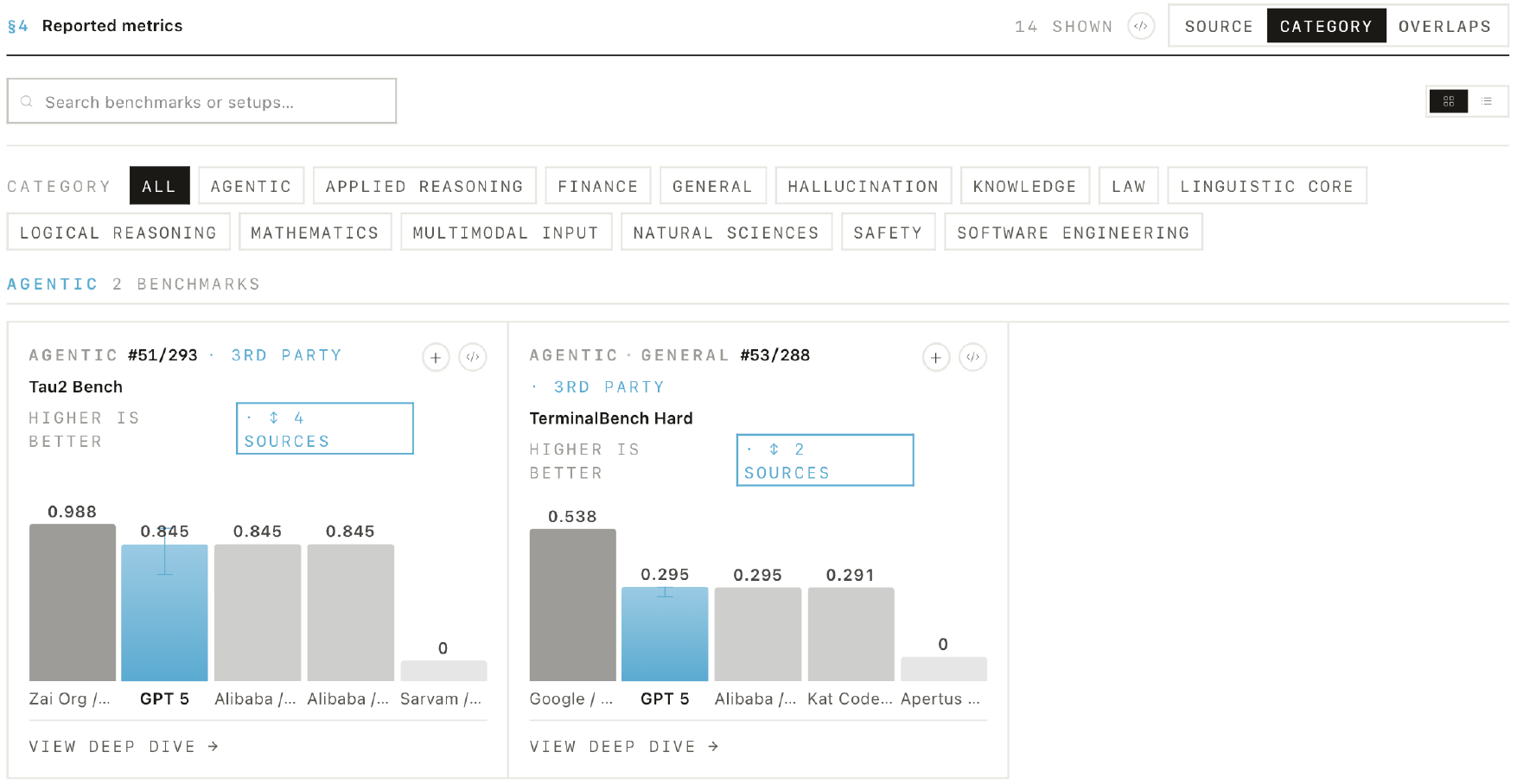}
    \caption{Reported metrics, Category: The Category view of the reported metrics compares model results on a particular benchmark across other models reported on that source.}
    \label{fig:category-gpt5}
\end{figure}

Model view provides context on the model evaluated and the reported evaluation results for that model through the \textbf{Identification} (\Cref{fig:gpt5-header}), \textbf{Who reports what} (\Cref{fig:gpt5-header}), and \textbf{Reported metrics} sections (\Cref{fig:overlap-gpt5,fig:category-gpt5}. 

\subsection{Benchmark view: MMLU-Pro}
To showcase the Benchmark view, we take a deep dive into MMLU-Pro \citep{wang2024mmlu}, one of many benchmarks with reported evaluation results. 
\begin{figure}[htbp]
    \centering
    \includegraphics[width=\linewidth]{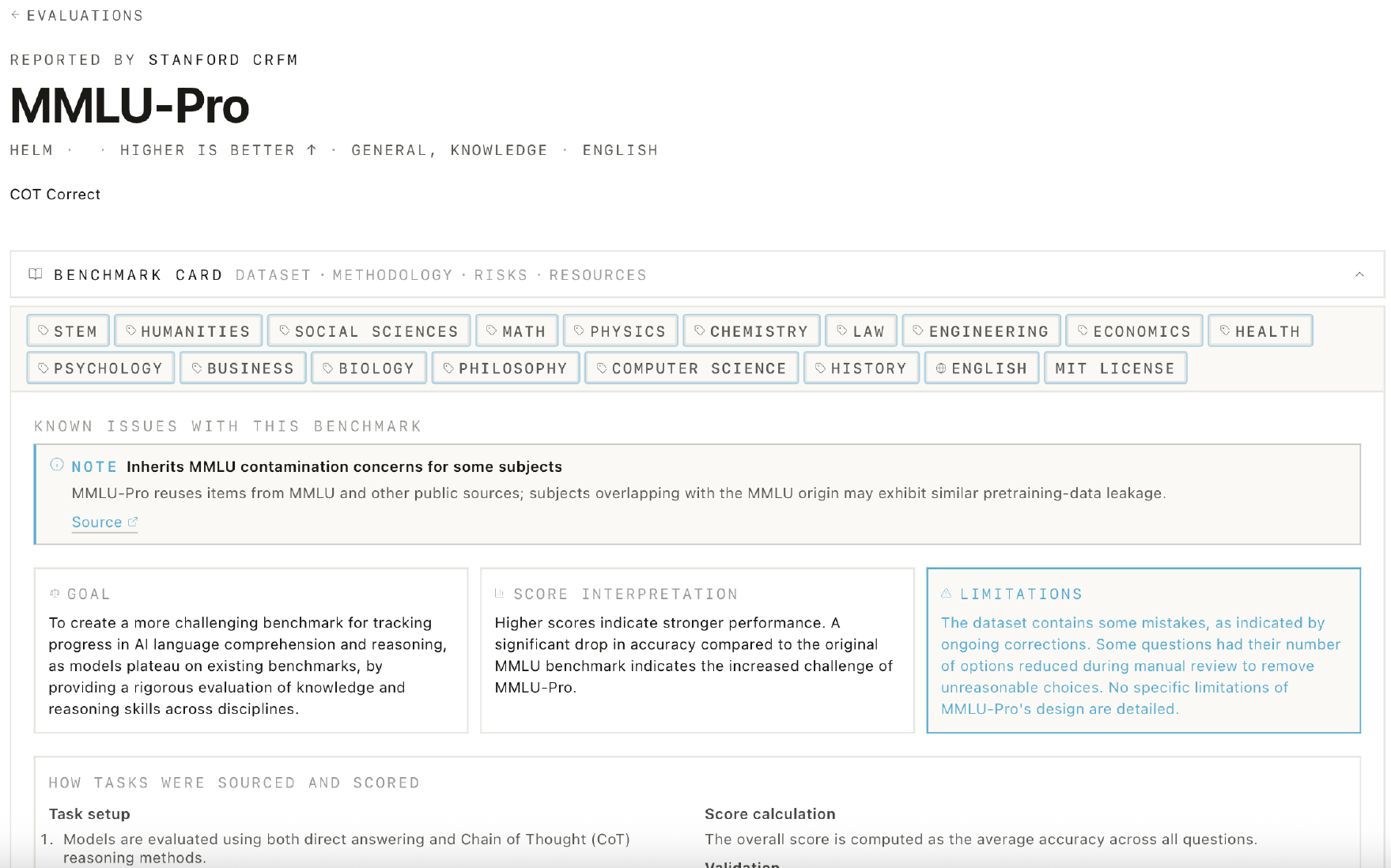}
    \caption{The benchmark card section shows a benchmark's coverage tags, licensing information, issues (e.g., contamination), goal, and an explanation of the score's meaning.}
    \label{fig:overview-mmlu}
\end{figure}
\begin{figure}[htbp]
    \centering
    \includegraphics[width=\linewidth]{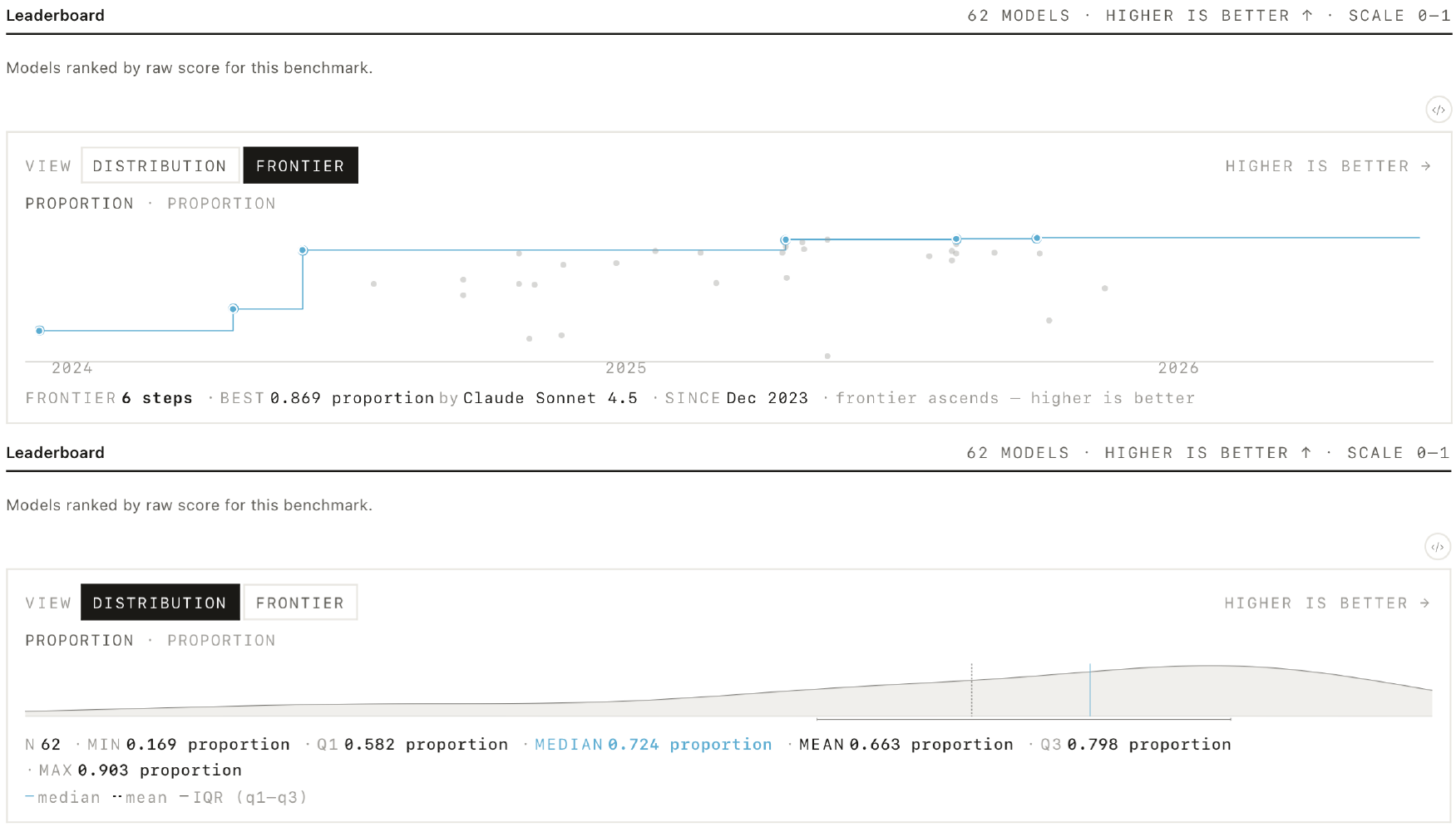}
    \caption{Leaderboard: (A) Frontier view showing the progression of top scores over time. (B) Distribution view showing the spread of scores across all reported models.}
    \label{fig:leaderboard-mmlu}
\end{figure}
\begin{figure}[htbp]
    \centering
    \includegraphics[width=\linewidth]{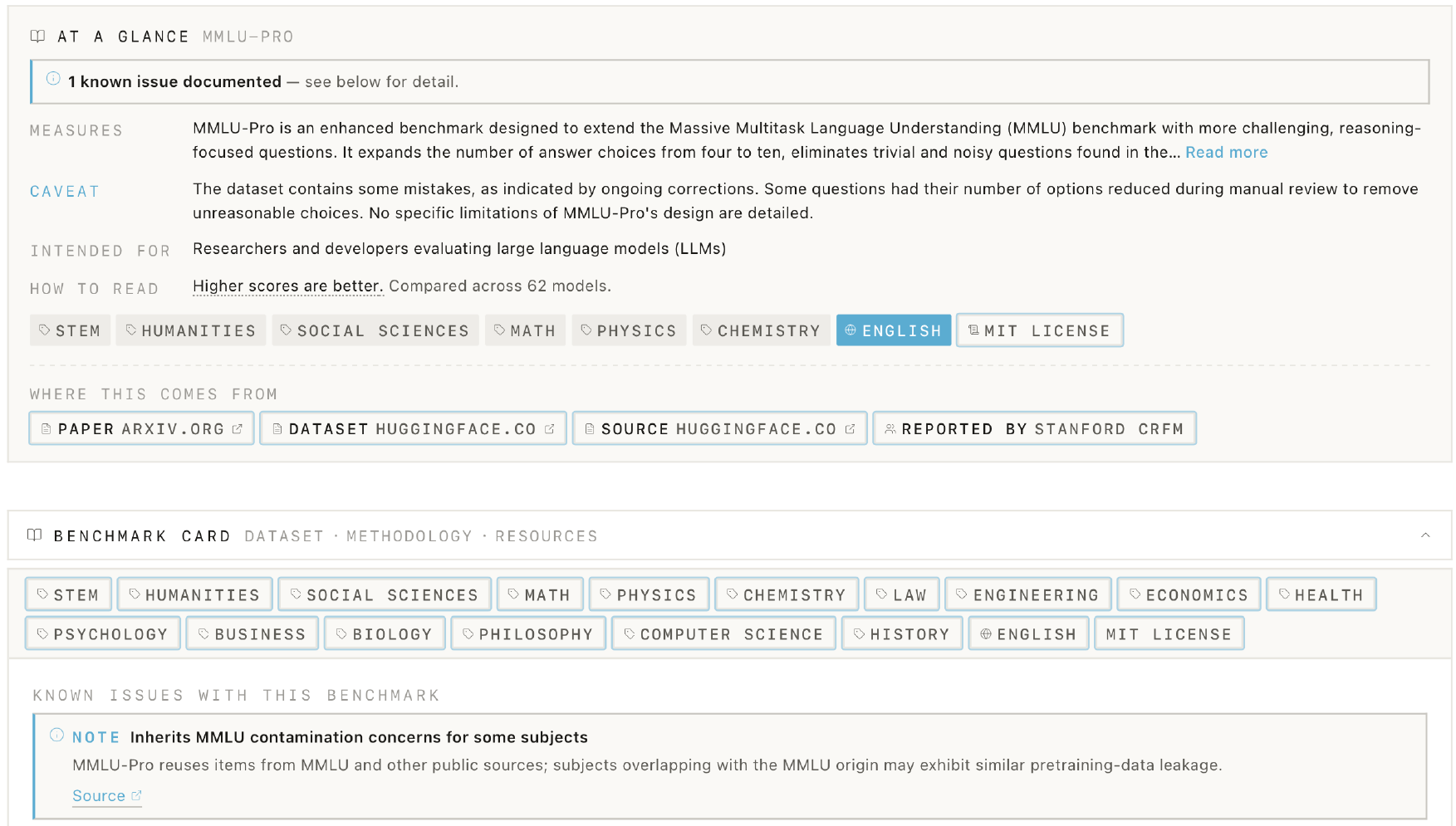}
    \caption{Summary mode displays benchmark data in plain language, summarizing the benchmark's purpose, known issues, and evaluation methodology for non-technical readers.}
    \label{fig:policy-mmlu}
\end{figure}
\begin{figure}[htbp]
    \centering
    \includegraphics[width=\linewidth]{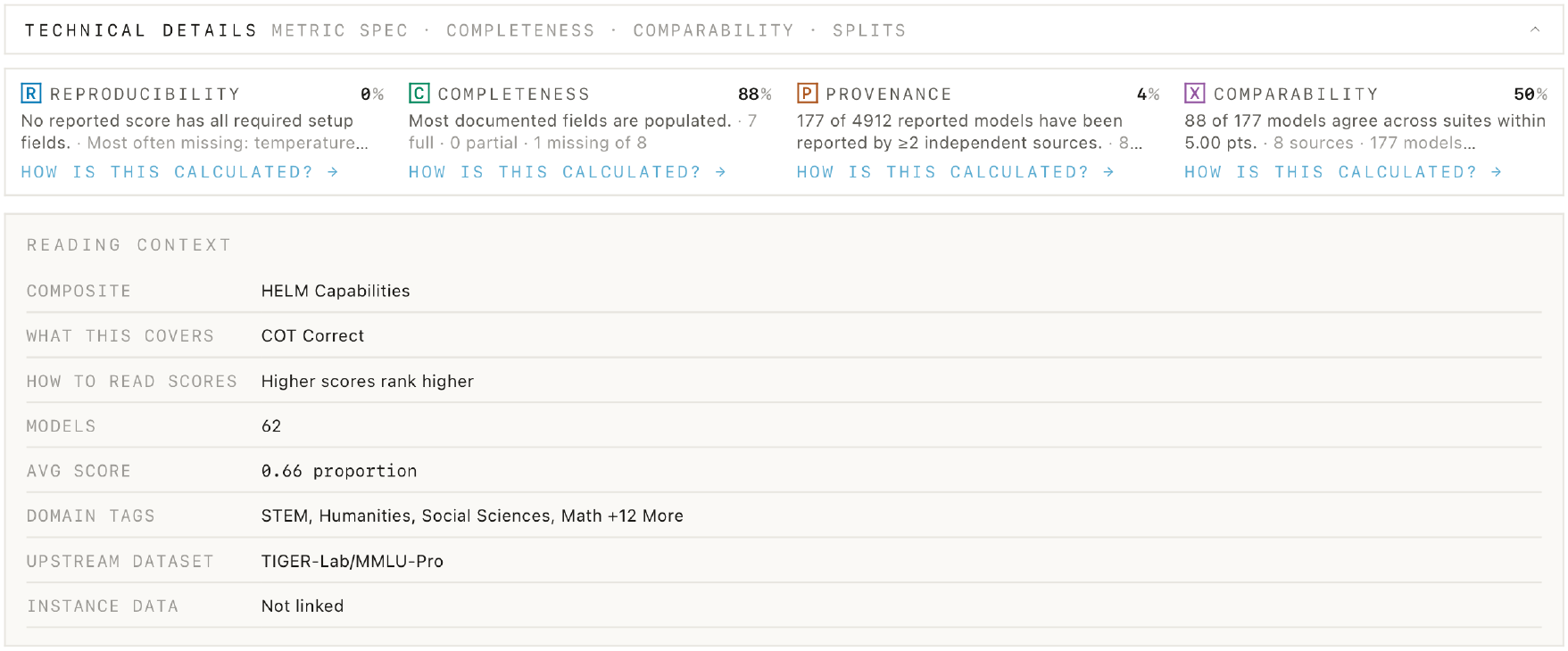}
    \caption{Interpretive signals panel displaying reproducibility, completeness, provenance, and comparability scores for an evaluation entry alongside metric specifications.}
    \label{fig:interpretive-mmlu}
\end{figure}
\begin{figure}[htbp]
    \centering
    \includegraphics[width=.75\linewidth]{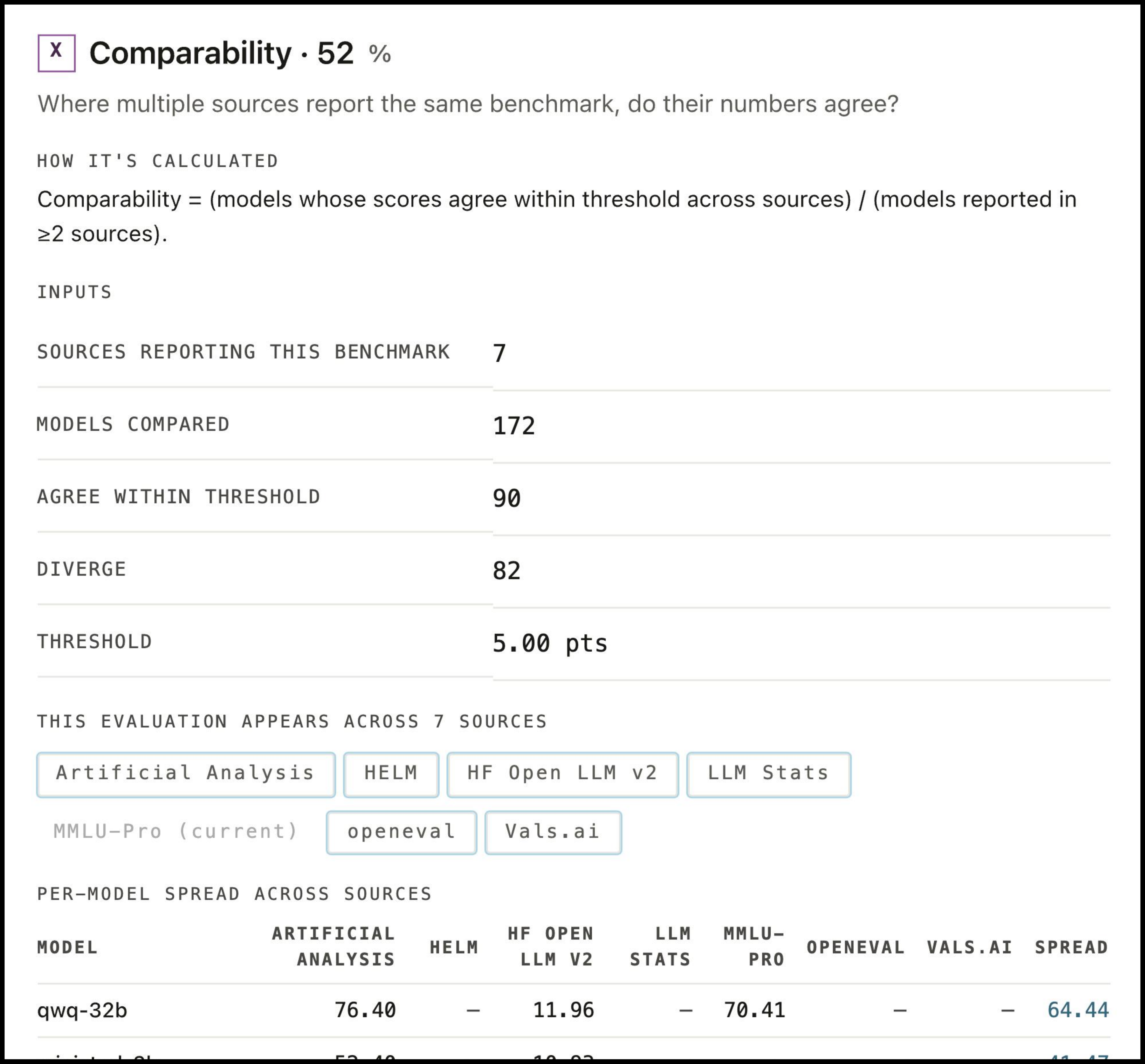}
    \caption{The comparability signal shows the threshold basis, number of models compared, and agreement rate across reporting sources.}
    \label{fig:interpretive-zoom-mmlu}
\end{figure}

\FloatBarrier
\subsection{Developer View}
\begin{figure}[htbp]
    \centering
    \includegraphics[width=\linewidth]{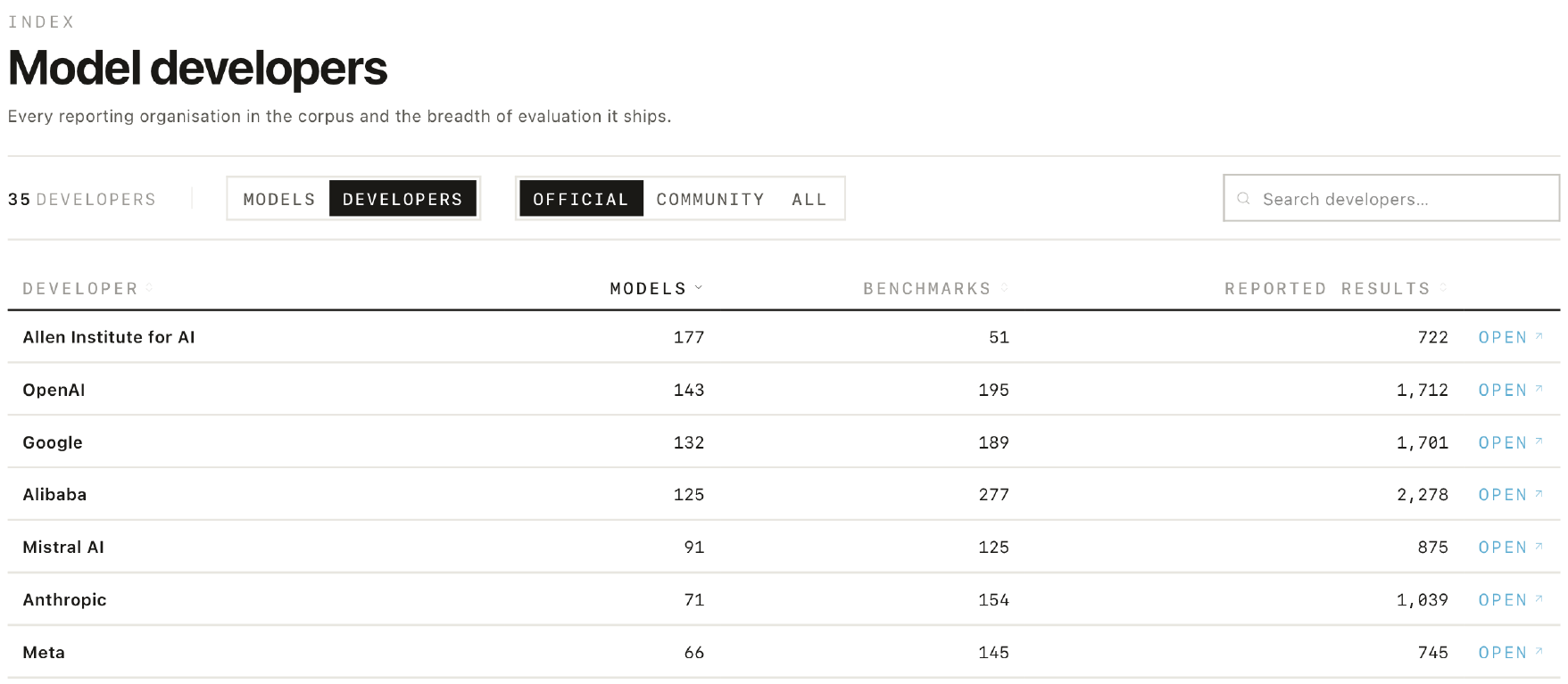}
    \caption{The Model Developers section displays the list of developers with reported models and the number of models and benchmarks submitted for each.}
    \label{fig:developer-mmlu}
\end{figure}



\clearpage

\section{Fields Ingested by \EvalCards{}}\label{app:evalcards-fields}
\begin{table}[htbp]
\centering
\small
\setlength{\tabcolsep}{4pt}
\caption{Auto-BenchmarkCard Schema Fields: Benchmark Details}
\label{tab:auto_benchmark_details}
\resizebox{\textwidth}{!}{%
\begin{tabular}{p{2.5cm}p{4.5cm}p{1.8cm}p{4.8cm}p{0.7cm}}
\toprule
\textbf{Section} & \textbf{Field} & \textbf{Type} & \textbf{Description} & \textbf{Req.} \\
\midrule
Benchmark Details & name & str & Official name of the benchmark as it appears in literature & yes \\
\addlinespace
Benchmark Details & overview & str & Comprehensive 2--3 sentence description of what the benchmark measures, its characteristics, and significance & yes \\
\addlinespace
Benchmark Details & data\_type & str & Primary data modality (e.g.\ text, image, audio, multimodal, tabular) & yes \\
\addlinespace
Benchmark Details & domains & List[str] & Application domains or subject areas (e.g.\ medical, legal, scientific) & yes \\
\addlinespace
Benchmark Details & languages & List[str] & Languages supported, using full names (e.g.\ English, Chinese, Multilingual) & yes \\
\addlinespace
Benchmark Details & similar benchmarks & List[str] & Names of closely related or comparable benchmarks & yes \\
\addlinespace
Benchmark Details & resources & List[str] & URLs to papers, datasets, leaderboards, and documentation & yes \\
\addlinespace
Benchmark Details & provenance & dict & Maps each field to \{source, evidence\}. Present in all content sections. & no \\
\bottomrule
\end{tabular}%
}
\end{table}

\begin{table}[htbp]
\centering
\small
\setlength{\tabcolsep}{4pt}
\caption{Auto-BenchmarkCard Schema Fields: Purpose \& Intended Users}
\label{tab:auto_benchmark_purpose}
\resizebox{\textwidth}{!}{%
\begin{tabular}{p{2.5cm}p{4.5cm}p{1.8cm}p{4.8cm}p{0.7cm}}
\toprule
\textbf{Section} & \textbf{Field} & \textbf{Type} & \textbf{Description} & \textbf{Req.} \\
\midrule
Purpose \& Intended Users & goal & str & Primary objective and research question the benchmark addresses & yes \\
\addlinespace
Purpose \& Intended Users & audience & List[str] & Target user groups (e.g.\ AI researchers, model developers, safety evaluators) & yes \\
\addlinespace
Purpose \& Intended Users & tasks & List[str] & Specific evaluation tasks covered (e.g.\ question answering, code generation) & yes \\
\addlinespace
Purpose \& Intended Users & limitations & str & Known limitations, biases, or constraints users should be aware of & yes \\
\addlinespace
Purpose \& Intended Users & out of scope uses & List[str] & Explicit examples of inappropriate or unsupported use cases & yes \\
\addlinespace
Purpose \& Intended Users & provenance & dict & Maps each field to \{source, evidence\}. Present in all content sections. & no \\
\bottomrule
\end{tabular}%
}
\end{table}

\begin{table}[htbp]
\centering
\small
\setlength{\tabcolsep}{4pt}
\caption{Auto-BenchmarkCard Schema Fields: Data}
\label{tab:auto_benchmark_data}
\resizebox{\textwidth}{!}{%
\begin{tabular}{p{2.5cm}p{4.5cm}p{1.8cm}p{4.8cm}p{0.7cm}}
\toprule
\textbf{Section} & \textbf{Field} & \textbf{Type} & \textbf{Description} & \textbf{Req.} \\
\midrule
Data & source & str & Data origins, collection methods, and any preprocessing steps applied & yes \\
\addlinespace
Data & size & str & Dataset size with specific numbers (e.g.\ 10,000 examples, 50K questions across 3 splits) & yes \\
\addlinespace
Data & format & str & Data structure and file formats (e.g.\ JSON with question-answer pairs) & yes \\
\addlinespace
Data & annotation & str & Annotation methodology, quality control, inter-annotator agreement, and human involvement & yes \\
\addlinespace
Data & provenance & dict & Maps each field to \{source, evidence\}. Present in all content sections. & no \\
\bottomrule
\end{tabular}%
}
\end{table}

\begin{table}[htbp]
\centering
\small
\setlength{\tabcolsep}{4pt}
\caption{Auto-BenchmarkCard Schema Fields: Methodology}
\label{tab:auto_benchmark_methodology}
\resizebox{\textwidth}{!}{%
\begin{tabular}{p{2.5cm}p{4.5cm}p{1.8cm}p{4.8cm}p{0.7cm}}
\toprule
\textbf{Section} & \textbf{Field} & \textbf{Type} & \textbf{Description} & \textbf{Req.} \\
\midrule
Methodology & methods & List[str] & Evaluation approaches applied (e.g.\ zero-shot, few-shot prompting, fine-tuning) & yes \\
\addlinespace
Methodology & metrics & List[str] & Quantitative metrics used (e.g.\ accuracy, F1-score, BLEU, exact match) & yes \\
\addlinespace
Methodology & calculation & str & How metrics are computed, including normalization or aggregation methods & yes \\
\addlinespace
Methodology & interpretation & str & Guidelines for interpreting scores, score ranges, and caveats & yes \\
\addlinespace
Methodology & baseline results & str & Performance of established models or baselines with specific numbers & yes \\
\addlinespace
Methodology & validation & str & Quality assurance measures and steps taken to ensure reproducible evaluations & yes \\
\addlinespace
Methodology & provenance & dict & Maps each field to \{source, evidence\}. Present in all content sections. & no \\
\bottomrule
\end{tabular}%
}
\end{table}

\begin{table}[htbp]
\centering
\small
\setlength{\tabcolsep}{4pt}
\caption{Auto-BenchmarkCard Schema Fields: Ethical \& Legal}
\label{tab:auto_benchmark_ethical}
\resizebox{\textwidth}{!}{%
\begin{tabular}{p{2.5cm}p{4.5cm}p{1.8cm}p{4.8cm}p{0.7cm}}
\toprule
\textbf{Section} & \textbf{Field} & \textbf{Type} & \textbf{Description} & \textbf{Req.} \\
\midrule
Ethical \& Legal & privacy \& anonymity & str & Data protection measures, anonymization techniques, and handling of PII & yes \\
\addlinespace
Ethical \& Legal & data licensing & str & License terms, usage restrictions, and redistribution permissions & yes \\
\addlinespace
Ethical \& Legal & consent procedures & str & Informed consent processes, participant rights, and withdrawal procedures & yes \\
\addlinespace
Ethical \& Legal & compliance & str & Adherence to GDPR, IRB approval, and other ethical review processes & yes \\
\addlinespace
Ethical \& Legal & provenance & dict & Maps each field to \{source, evidence\}. Present in all content sections. & no \\
\bottomrule
\end{tabular}%
}
\end{table}

\begin{table}[htbp]
\centering
\small
\setlength{\tabcolsep}{4pt}
\caption{Auto-BenchmarkCard Schema Fields: Possible Risks}
\label{tab:auto_benchmark_risk}
\resizebox{\textwidth}{!}{%
\begin{tabular}{p{2.5cm}p{4.5cm}p{1.8cm}p{4.8cm}p{0.7cm}}
\toprule
\textbf{Section} & \textbf{Field} & \textbf{Type} & \textbf{Description} & \textbf{Req.} \\
\midrule
Possible Risks & category & str & Risk category name & no \\
\addlinespace
Possible Risks & description & str & Description of the risk & no \\
\addlinespace
Possible Risks & type & str & Risk type classification & no \\
\addlinespace
Possible Risks & concern & str & Specific concern this risk raises & no \\
\addlinespace
Possible Risks & url & str\textbar{}null & Link to the risk entry in IBM Risk Atlas & no \\
\addlinespace
Possible Risks & taxonomy & str & Taxonomy this risk belongs to & no \\
\bottomrule
\end{tabular}%
}
\end{table}

\begin{table}[htbp]
\centering
\small
\setlength{\tabcolsep}{4pt}
\caption{Auto-BenchmarkCard Schema Fields: Flagged Fields}
\label{tab:auto_benchmark_flagged_fields}
\resizebox{\textwidth}{!}{%
\begin{tabular}{p{2.5cm}p{4.5cm}p{1.8cm}p{4.8cm}p{0.7cm}}
\toprule
\textbf{Section} & \textbf{Field} & \textbf{Type} & \textbf{Description} & \textbf{Req.} \\
\midrule
Flagged Fields & \textless{}section\textgreater{}.\textless{}field\textgreater{} & str & Added by FactReasoner. Flags potential hallucinations or low factual alignment with source material. & no \\
\bottomrule
\end{tabular}%
}
\end{table}

\begin{table}[htbp]
\centering
\small
\setlength{\tabcolsep}{4pt}
\caption{EEE Evaluation-Level Schema Fields: Top-level}
\label{tab:eee_top-level}
\resizebox{\textwidth}{!}{%
\begin{tabular}{p{2.5cm}p{4.5cm}p{1.8cm}p{4.8cm}p{0.7cm}}
\toprule
\textbf{Section} & \textbf{Field} & \textbf{Type} & \textbf{Description} & \textbf{Req.} \\
\midrule
Top-level & schema version & string & Version of the schema used for this evaluation data & yes \\
\addlinespace
Top-level & evaluation id & string & Unique identifier for this specific evaluation run. Use eval\_name... & yes \\
\addlinespace
Top-level & evaluation timestamp & string & Timestamp for when the evaluation was run & no \\
\addlinespace
Top-level & retrieved timestamp & string & Timestamp for when this record was created - using Unix Epoch tim... & yes \\
\bottomrule
\end{tabular}%
}
\end{table}

\begin{table}[htbp]
\centering
\small
\setlength{\tabcolsep}{4pt}
\caption{EEE Evaluation-Level Schema Fields: Source Metadata}
\label{tab:eee_source_metadata}
\resizebox{\textwidth}{!}{%
\begin{tabular}{p{2.5cm}p{4.5cm}p{1.8cm}p{4.8cm}p{0.7cm}}
\toprule
\textbf{Section} & \textbf{Field} & \textbf{Type} & \textbf{Description} & \textbf{Req.} \\
\midrule
Source Metadata & source metadata & object & Metadata about the source of the leaderboard data & yes \\
\addlinespace
Source Metadata & source\_name & string & Name of the source (e.g. title of the source leaderboard or name ... & no \\
\addlinespace
Source Metadata & source\_type & enum: "documentation" \textbar{} "evaluation\_run" & Whether the data comes from a direct evaluation run or from docum... & yes \\
\addlinespace
Source Metadata & source\_organization\_name & string & Name of the organization that provides the data & yes \\
\addlinespace
Source Metadata & source\_organization\_url & string & URL for the organization that provides the data & no \\
\addlinespace
Source Metadata & source\_organization\_logo\_url & string & URL for the Logo for the organization that provides the data & no \\
\addlinespace
Source Metadata & evaluator\_relationship & enum: "first\_party" \textbar{} "third\_party" \textbar{} "collaborative" \textbar{} "other" & Relationship between the evaluator and the model & yes \\
\addlinespace
Source Metadata & additional\_details & dict[str, string] & Additional parameters (key-value pairs, all values must be string... & no \\
\bottomrule
\end{tabular}%
}
\end{table}

\begin{table}[htbp]
\centering
\small
\setlength{\tabcolsep}{4pt}
\caption{EEE Evaluation-Level Schema Fields: Eval Library}
\label{tab:eee_eval_library}
\resizebox{\textwidth}{!}{%
\begin{tabular}{p{2.5cm}p{4.5cm}p{1.8cm}p{4.8cm}p{0.7cm}}
\toprule
\textbf{Section} & \textbf{Field} & \textbf{Type} & \textbf{Description} & \textbf{Req.} \\
\midrule
Eval Library & eval library & object & Evaluation library/framework used to run the evaluation & yes \\
\addlinespace
Eval Library & name & string & Name of the evaluation library (e.g. inspect\_ai, lm\_eval, helm) & yes \\
\addlinespace
Eval Library & version & string & Version of the evaluation library. Use 'unknown' if the version i... & yes \\
\addlinespace
Eval Library & additional\_details & dict[str, string] & Additional parameters (key-value pairs, all values must be string... & no \\
\bottomrule
\end{tabular}%
}
\end{table}

\begin{table}[htbp]
\centering
\small
\setlength{\tabcolsep}{4pt}
\caption{EEE Evaluation-Level Schema Fields: Model Info}
\label{tab:eee_model_info}
\resizebox{\textwidth}{!}{%
\begin{tabular}{p{2.5cm}p{4.5cm}p{1.8cm}p{4.8cm}p{0.7cm}}
\toprule
\textbf{Section} & \textbf{Field} & \textbf{Type} & \textbf{Description} & \textbf{Req.} \\
\midrule
Model Info & model info & object & Complete model specification including basic information, technic... & yes \\
\addlinespace
Model Info & name & string & Model name provided by evaluation source & yes \\
\addlinespace
Model Info & id & string & Model name in HuggingFace format (e.g. meta-llama/Llama-3.1-8B-In... & yes \\
\addlinespace
Model Info & developer & string & Name of organization that provides the model (e.g. 'OpenAI') & no \\
\addlinespace
Model Info & inference\_platform & string & Name of inference platform which provides an access to models by ... & no \\
\addlinespace
Model Info & inference\_engine & object & Name of inference engine which provides an access to optimized mo... & no \\
\addlinespace
Model Info & inference\_engine.name & string & Name of the inference engine & no \\
\addlinespace
Model Info & inference\_engine.version & string & Version of the inference engine & no \\
\addlinespace
Model Info & additional\_details & dict[str, string] & Additional parameters (key-value pairs, all values must be string... & no \\
\bottomrule
\end{tabular}%
}
\end{table}


\begin{table}[htbp]
\centering
\small
\setlength{\tabcolsep}{4pt}
\caption{EEE Evaluation-Level Schema Fields: Evaluation Results (Core)}
\label{tab:eee_evaluation_results_core}
\resizebox{\textwidth}{!}{%
\begin{tabular}{p{2.5cm}p{4.5cm}p{1.8cm}p{4.8cm}p{0.7cm}}
\toprule
\textbf{Section} & \textbf{Field} & \textbf{Type} & \textbf{Description} & \textbf{Req.} \\
\midrule
Evaluation Results & evaluation results & array[object] & Array of evaluation results & yes \\
\addlinespace
Evaluation Results & evaluation\_result\_id & string & Stable identifier for this metric result inside an evaluation run & no \\
\addlinespace
Evaluation Results & evaluation\_name & string & Name of the evaluation & yes \\
\addlinespace
Evaluation Results & source\_data & oneOf: url \textbar{} hf\_dataset \textbar{} other & Source of dataset for this evaluation: URL, HuggingFace dataset, or other & yes \\
\addlinespace
Evaluation Results & source\_data.dataset\_name & string & Name of the source dataset & yes \\
\addlinespace
Evaluation Results & source\_data.url & array[string] & URL(s) for the source of the evaluation data (only when source\_type=url) & yes \\
\addlinespace
Evaluation Results & source\_data.additional\_details & dict[str, string] & Additional parameters (key-value pairs, all values must be strings) & no \\
\addlinespace
Evaluation Results & source\_data.hf\_repo & string & HuggingFace repository identifier (only when source\_type=hf\_dataset) & no \\
\addlinespace
Evaluation Results & source\_data.hf\_split & string & One of train, val or test (only when source\_type=hf\_dataset) & no \\
\addlinespace
Evaluation Results & source\_data.samples\_number & integer & Number of samples in the dataset (only when source\_type=hf\_dataset) & no \\
\addlinespace
Evaluation Results & source\_data.sample\_ids & array[string] & Array of sample IDs used for evaluation (only when source\_type=hf\_dataset) & no \\
\addlinespace
Evaluation Results & evaluation\_timestamp & string & Timestamp for when the evaluations were run & no \\
\bottomrule
\end{tabular}%
}
\end{table}

\begin{table}[htbp]
\centering
\small
\setlength{\tabcolsep}{4pt}
\caption{EEE Evaluation-Level Schema Fields: Evaluation Results -- Metric Config}
\label{tab:eee_evaluation_results_metric_config}
\resizebox{\textwidth}{!}{%
\begin{tabular}{p{2.5cm}p{4.5cm}p{1.8cm}p{4.8cm}p{0.7cm}}
\toprule
\textbf{Section} & \textbf{Field} & \textbf{Type} & \textbf{Description} & \textbf{Req.} \\
\midrule
Evaluation Results & metric\_config & object & Details about the metric & yes \\
\addlinespace
Evaluation Results & metric\_config.evaluation\_description & string & Description of the evaluation & no \\
\addlinespace
Evaluation Results & metric\_config.metric\_id & string & Stable metric identifier for joining, deduping, and querying & no \\
\addlinespace
Evaluation Results & metric\_config.metric\_name & string & Display name for the metric (e.g., Accuracy, F1-macro, pass@1) & no \\
\addlinespace
Evaluation Results & metric\_config.metric\_kind & string & Normalized metric family/type used for safe aggregation & no \\
\addlinespace
Evaluation Results & metric\_config.metric\_unit & string & Unit of the metric values (e.g., proportion, percent, points, ms) & no \\
\addlinespace
Evaluation Results & metric\_config.metric\_parameters & dict[str, string \textbar{} number \textbar{} boolean \textbar{} null] & Metric-specific parameters (e.g., \{"k": 1\} for pass@k) & no \\
\addlinespace
Evaluation Results & metric\_config.lower\_is\_better & boolean & Whether a lower score is better & yes \\
\addlinespace
Evaluation Results & metric\_config.score\_type & enum & Type of score (binary, continuous, levels) & no \\
\addlinespace
Evaluation Results & metric\_config.level\_names & array[string] & Names of the score levels & no \\
\addlinespace
Evaluation Results & metric\_config.level\_metadata & array[string] & Additional description for each score level & no \\
\addlinespace
Evaluation Results & metric\_config.has\_unknown\_level & boolean & Indicates whether there is an unknown level & no \\
\addlinespace
Evaluation Results & metric\_config.min\_score & number & Minimum possible score for continuous metric & no \\
\addlinespace
Evaluation Results & metric\_config.max\_score & number & Maximum possible score for continuous metric & no \\
\bottomrule
\end{tabular}%
}
\end{table}

\begin{table}[htbp]
\centering
\small
\setlength{\tabcolsep}{4pt}
\caption{EEE Evaluation-Level Schema Fields: Evaluation Results -- Score Details}
\label{tab:eee_evaluation_results_score_details}
\resizebox{\textwidth}{!}{%
\begin{tabular}{p{2.5cm}p{4.5cm}p{1.8cm}p{4.8cm}p{0.7cm}}
\toprule
\textbf{Section} & \textbf{Field} & \textbf{Type} & \textbf{Description} & \textbf{Req.} \\
\midrule
Evaluation Results & score\_details & object & The score for the evaluation and related details & yes \\
\addlinespace
Evaluation Results & score\_details.score & number & The score for the evaluation & yes \\
\addlinespace
Evaluation Results & score\_details.details & dict[str, string] & Additional parameters (key-value pairs) & no \\
\addlinespace
Evaluation Results & score\_details.uncertainty & object & Quantification of uncertainty around the reported score & no \\
\addlinespace
Evaluation Results & \makecell[l]{score\_details.uncertainty.\\standard\_error.value} & number & The standard error value & yes \\
\addlinespace
Evaluation Results & \makecell[l]{score\_details.uncertainty.\\confidence\_interval.lower} & number & Lower bound of the confidence interval & yes \\
\addlinespace
Evaluation Results & \makecell[l]{score\_details.uncertainty.\\confidence\_interval.upper} & number & Upper bound of the confidence interval & yes \\
\addlinespace
Evaluation Results & \makecell[l]{score\_details.uncertainty.\\confidence\_interval.\\confidence\_level} & number & Confidence level (e.g., 0.95) & no \\
\bottomrule
\end{tabular}%
}
\end{table}

\begin{table}[htbp]
\centering
\small
\setlength{\tabcolsep}{4pt}
\caption{EEE Evaluation-Level Schema Fields: Evaluation Results -- Generation Config}
\label{tab:eee_evaluation_results_generation_config}
\resizebox{\textwidth}{!}{%
\begin{tabular}{p{2.5cm}p{4.5cm}p{1.8cm}p{4.8cm}p{0.7cm}}
\toprule
\textbf{Section} & \textbf{Field} & \textbf{Type} & \textbf{Description} & \textbf{Req.} \\
\midrule
Evaluation Results & generation\_config & object & Generation configuration & no \\
\addlinespace
Evaluation Results & \makecell[l]{generation\_config.\\generation\_args.temperature} & number & Sampling temperature & no \\
\addlinespace
Evaluation Results & \makecell[l]{generation\_config.\\generation\_args.top\_p} & number & Nucleus sampling parameter & no \\
\addlinespace
Evaluation Results & \makecell[l]{generation\_config.\\generation\_args.top\_k} & number & Top-k sampling parameter & no \\
\addlinespace
Evaluation Results & \makecell[l]{generation\_config.\\generation\_args.max\_tokens} & integer & Maximum number of tokens to generate & no \\
\bottomrule
\end{tabular}%
}
\end{table}

\begin{table}[htbp]
\centering
\small
\setlength{\tabcolsep}{4pt}
\caption{EEE Evaluation-Level Schema Fields: Detailed Results}
\label{tab:eee_detailed_results}
\resizebox{\textwidth}{!}{%
\begin{tabular}{p{2.5cm}p{4.5cm}p{1.8cm}p{4.8cm}p{0.7cm}}
\toprule
\textbf{Section} & \textbf{Field} & \textbf{Type} & \textbf{Description} & \textbf{Req.} \\
\midrule
Detailed Results & detailed evaluation results & object & Reference to the evaluation results for all individual samples in... & no \\
\addlinespace
Detailed Results & format & enum: "jsonl" \textbar{} "json" & Format of the detailed evaluation results & no \\
\addlinespace
Detailed Results & file\_path & string & Path to the detailed evaluation results file & no \\
\addlinespace
Detailed Results & hash\_algorithm & enum: "sha256" \textbar{} "md5" & Hash algorithm used for checksum and sample\_hash in instance-leve... & no \\
\addlinespace
Detailed Results & checksum & string & Checksum value of the file & no \\
\addlinespace
Detailed Results & total\_rows & integer & Total number of rows in the detailed evaluation results file & no \\
\addlinespace
Detailed Results & additional\_details & dict[str, string] & Additional parameters (key-value pairs, all values must be string... & no \\
\bottomrule
\end{tabular}%
}
\end{table}

\clearpage
\section{Mapping from Systematic Literature Review to \EvalCards{} Schema and Interpretive Signals}
\label{app:schema-mapping}

\begin{figure}[ht]
  \centering
  \includegraphics[
    page=1,
    width=1\linewidth,
    clip
  ]{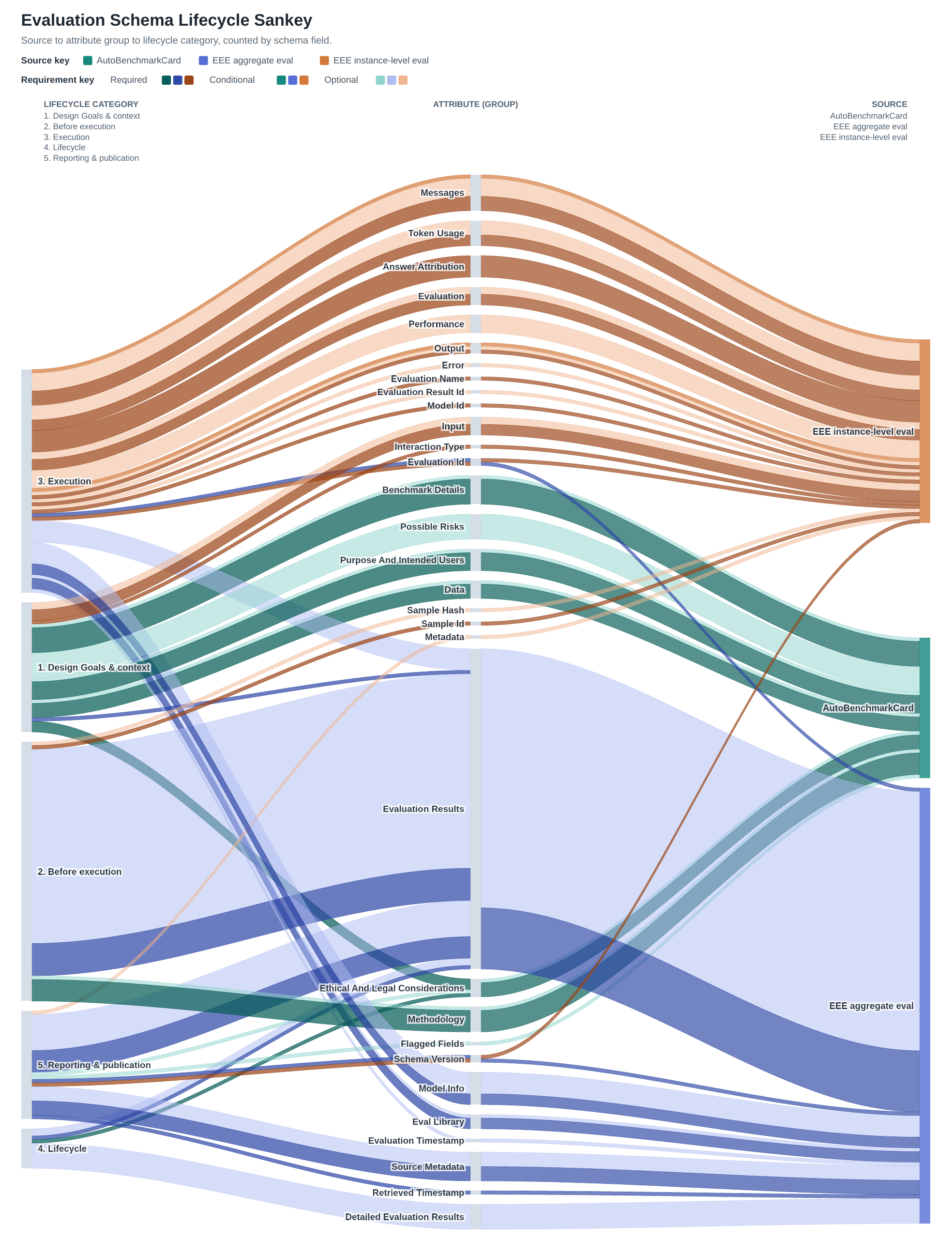}
  \caption{Sankey diagram mapping from input groups, to individual items, to sources ingested by \EvalCards{}.}
  \label{fig:item_mapping_sankey}
\end{figure}

\begin{figure*}[t]
\centering
\begin{tikzpicture}[
  >=Latex,
  font=\scriptsize,
  every node/.style={align=center, rounded corners=2pt, inner sep=2.5pt},
  stage/.style={draw, thick, minimum width=26mm, minimum height=8mm},
  field/.style={draw, thick, fill=ecLight, minimum width=38mm, minimum height=8mm},
  outbox/.style={draw, thick, minimum width=30mm, minimum height=9mm},
  a/.style={stage, draw=ecBlue, fill=ecBlue!12},
  b/.style={stage, draw=ecViolet, fill=ecViolet!12},
  c/.style={stage, draw=ecTeal, fill=ecTeal!12},
  d/.style={stage, draw=ecOrange, fill=ecOrange!12},
  e/.style={stage, draw=ecGreen, fill=ecGreen!12},
  oR/.style={outbox, draw=ecRed, fill=ecRed!10},
  oT/.style={outbox, draw=ecTeal, fill=ecTeal!10},
  oO/.style={outbox, draw=ecOrange, fill=ecOrange!12},
  oB/.style={outbox, draw=ecBlue, fill=ecBlue!10},
  arrow/.style={->, thick, draw=ecGray}
]
\node[a] (g1) at (0,0) {1. Design};
\node[b] (g2) at (0,-1.15) {2. Before execution};
\node[c] (g3) at (0,-2.30) {3. Execution};
\node[d] (g4) at (0,-3.45) {4. Lifecycle};
\node[e] (g5) at (0,-4.60) {5. Reporting \& publication};
\node[field] (f1) at (4.15,0) {Goals, construct\\validity, task types};
\node[field] (f2) at (4.15,-1.15) {Protocol, splits,\\baselines, contamination};
\node[field] (f3) at (4.15,-2.30) {Run logs, mitigations,\\generation config};
\node[field] (f4) at (4.15,-3.45) {Data access, later use,\\maintenance};
\node[field] (f5) at (4.15,-4.60) {Transparency, replication,\\publication details};
\node[oB] (out4) at (9.00,-1.15) {Comparability};
\node[oR] (out1) at (9.00,-2.30) {Completeness};
\node[oT] (out2) at (9.00,-3.45) {Reproducibility};
\node[oO] (out3) at (9.00,-4.60) {Provenance};
\draw[arrow] (g1) -- (f1);
\draw[arrow] (g2) -- (f2);
\draw[arrow] (g3) -- (f3);
\draw[arrow] (g4) -- (f4);
\draw[arrow] (g5) -- (f5);
\draw[arrow, draw=ecBlue] (f2.east) -- (out4.west);
\draw[arrow, draw=ecRed] (f3.east) -- (out1.west);
\draw[arrow, draw=ecTeal] (f3.east) to[out=0,in=180] (out2.west);
\draw[arrow, draw=ecOrange] (f5.east) -- (out3.west);
\end{tikzpicture}
\caption{Traceability from the literature-derived framework to \EvalCards{} fields and interpretive outputs.}
\label{fig:framework-map}
\end{figure*}
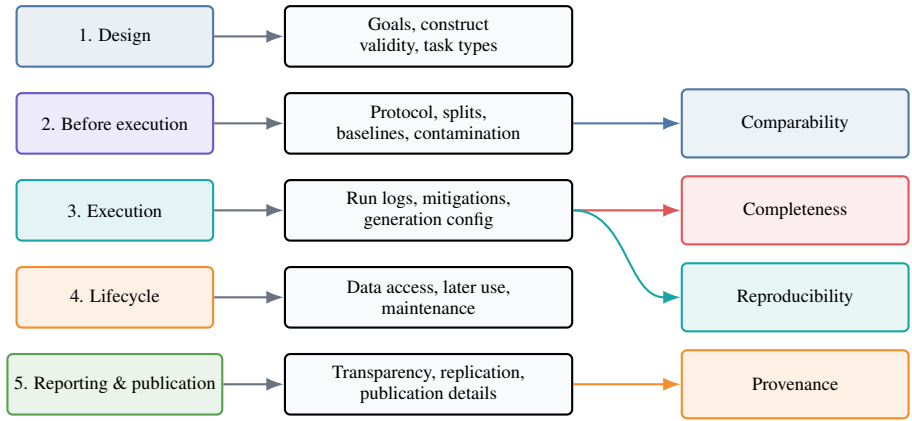

We show how the systematic literature review items map to fields ingested by \EvalCards{} from different sources in in \Cref{fig:item_mapping_sankey}, and how they map to interpretive signals in \Cref{fig:framework-map}.

\section{Code and Demo}
\label{code-and-demo}

Our code is available at \url{https://huggingface.co/spaces/evaleval/general-eval-card/tree/main}, and the live demo at \url{https://evalcards.evalevalai.com}.

\end{document}